\documentclass{article} %
\usepackage{preprint,times}
\usepackage{xcolor}

\usepackage{amsmath,amsfonts,bm}

\def\eqref#1{equation~\ref{#1}}

\def\1{\bm{1}}

\def\vd{{\bm{d}}}

\DeclareMathAlphabet{\mathsfit}{\encodingdefault}{\sfdefault}{m}{sl}
\SetMathAlphabet{\mathsfit}{bold}{\encodingdefault}{\sfdefault}{bx}{n}

\usepackage{graphicx}
\usepackage{url}
\usepackage{inconsolata}
\usepackage{fontawesome5}
\usepackage{booktabs}
\usepackage{placeins}
\usepackage{float}
\usepackage{hyperref}%
\usepackage{etoolbox}
\AtBeginEnvironment{table}{\setlength{\belowcaptionskip}{6pt}}

\definecolor{citecolor}{HTML}{2779af}
\definecolor{linkcolor}{HTML}{c0392b}
\hypersetup{breaklinks=true,colorlinks,citecolor=citecolor,linkcolor=linkcolor}

\title{Dynamic Token Merging for Efficient\\Vision-Language Models}

\usepackage{xspace}
\definecolor{vdtmcolor}{HTML}{606cdb}
\DeclareRobustCommand{\dtm}{\textcolor{vdtmcolor}{\textbf{DTM}}\xspace}
\DeclareRobustCommand{\vdtm}{\dtm}

\definecolor{fastv-yellow}{HTML}{BE8D12}
\DeclareRobustCommand{\fastv}{\textcolor{fastv-yellow}{\textbf{FastV}}\xspace}

\definecolor{phir-magenta}{HTML}{c404a4} %
\DeclareRobustCommand{\phir}{\textcolor{phir-magenta}{\textbf{$\varphi$\mbox{-}R}}\xspace}

\definecolor{random}{HTML}{00a816} %
\DeclareRobustCommand{\random}{\textcolor{random}{\textbf{random}}\xspace}
\DeclareRobustCommand{\Random}{\textcolor{random}{\textbf{Random}}\xspace}

\DeclareRobustCommand{\vmrtfivegemmatwo}{V-MrT5Gemma2\xspace}
\DeclareRobustCommand{\vmrgemmathree}{V-MrGemma3\xspace}
\DeclareRobustCommand{\mrtfive}{MrT5\xspace}

\definecolor{pidP}{HTML}{648DD3}
\definecolor{pidI}{HTML}{9A7BCD}
\definecolor{pidD}{HTML}{369C89}

\author{Chris Gu \quad Marco Andono Sie\thanks{Equal contribution.} \quad Nathan Zhou\footnotemark[1] \quad Christopher Potts \quad Julie Kallini\\
Stanford University\\
{\normalfont\small \href{mailto:chrisgu@cs.stanford.edu}{\texttt{chrisgu@cs.stanford.edu}}}}
\begin{document}
\raggedbottom

\maketitle

\begin{abstract}
Vision-language models (VLMs) represent each image with hundreds of visual tokens, which can make inference costly even when a prompt depends on only part of the image. \textbf{Dynamic Token Merging (\dtm)}, introduced in MrT5 for byte-level text encoders, learns which tokens to remove: a lightweight delete gate scores contextualized hidden states, acts as a differentiable attention bias during training, and physically removes tokens at inference. We argue that \dtm is a general token deletion framework rather than a byte-level technique, and extend it to the vision modality and to decoder-only models. Applying \dtm to T5Gemma 2 and Gemma 3 yields \vmrtfivegemmatwo, which deletes visual tokens inside a bidirectional multimodal encoder, and \vmrgemmathree, which compacts the prompt and its key--value caches during decoder-only prefill. Across ten benchmarks, \dtm's advantage over attention-based and random deletion grows with the deletion rate and is largest on document, chart, and scene-text reading. 
At 80\% visual token deletion ($\approx$60\% fewer FLOPs), \dtm raises accuracy over FastV from 51.8 to 63.0 on ChartQA and from 63.3 to 67.8 on TextVQA with \mbox{\vmrtfivegemmatwo}, as well as from 62.2 to 68.8 and from 71.2 to 75.1 with \mbox{\vmrgemmathree}, retaining 90--96\% of unpruned performance on both tasks with significant efficiency gains. These results support \dtm as a general framework for efficient transformers.

\end{abstract}
\vspace{-0.8em}

{\centering
\href{https://github.com/cgu2022/dtm_for_vlms}{\textcolor{black}{\faGithub}\hspace{0.5em}\textcolor{blue!60!black}{\texttt{cgu2022/dtm\_for\_vlms}}}\par}
\vspace{0.5em}

\section{Introduction}
\label{sec:introduction}

Long token sequences are a major source of inference cost in transformer models, and this problem arises across many input modalities. Vision-language models (VLMs), for example, may represent a single image with hundreds of visual tokens, even when a prompt depends on only part of the image: reading one value from a bar chart may depend on only a few labels and bar ends. Byte-level language models encounter a similar problem, since they process much longer sequences than conventional subword models, even though a word rarely needs one representation per byte.

Despite these differences, both settings produce long token sequences that must be processed repeatedly across transformer layers. One way to improve efficiency is to shorten the sequence after the first few layers, so that later layers operate on fewer tokens. The challenge is determining which representations can be removed while preserving the information needed for the end task. In VLMs, this choice is often made by a fixed rule computed from the model's own attention, such as \fastv \citep{chen2024imageworth12tokens}; the rule itself does not adapt when the model is fine-tuned for a downstream task, and simple selectors, including random selection, can be competitive with it \citep{wen-etal-2025-token}.

MrT5 introduced \textbf{Dynamic Token Merging} (\dtm; \citealt{kallini2025mrt5dynamictokenmerging}) as a learned mechanism for reducing sequence lengths in byte-level text encoders. \dtm applies a scalar \emph{delete gate} to contextual hidden states to identify tokens that can be removed, with the gate trained jointly with the model: during training, its scores act as soft attention biases, and at inference, low-scoring tokens are physically removed. Because token selection operates on contextualized representations rather than the original input tokens, information from tokens selected for deletion is already encoded in the representations that remain, which is the sense in which \dtm \emph{merges} tokens. In MrT5, \dtm shortened byte-level sequences by up to 75\% with little loss in downstream accuracy. Although \dtm was originally developed for byte-level text encoders, its mechanism is not inherently tied to text: the gate reads only hidden states, and soft and hard deletion act only through attention.

\begin{figure}[t]
    \centering
    \includegraphics[width=0.98\linewidth]{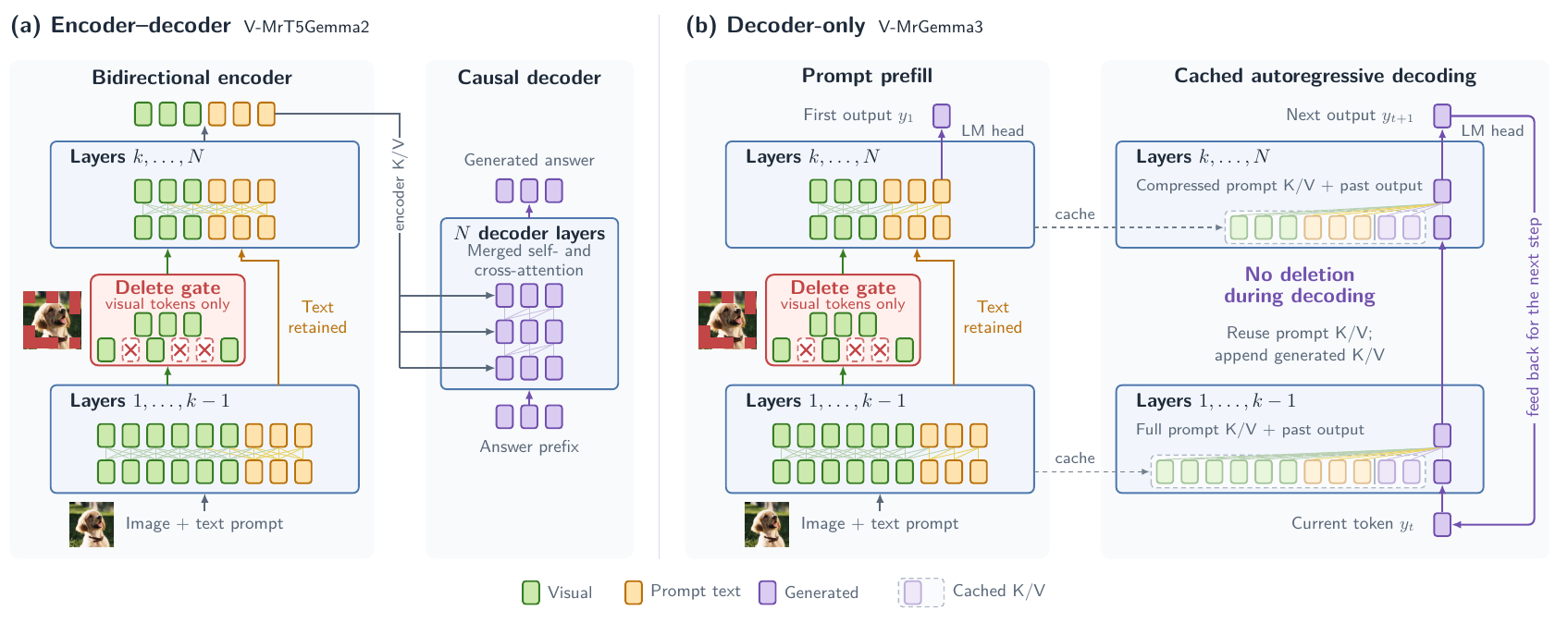}
    \setlength{\abovecaptionskip}{-1pt}
    \caption{\textbf{Dynamic Token Merging (DTM) in encoder--decoder and decoder-only VLMs.}
    \textbf{(a)} \vmrtfivegemmatwo deletes selected visual tokens before encoder layer $k$, retaining all text tokens. The shortened encoder output feeds the decoder's merged self- and cross-attention.
    \textbf{(b)} \vmrgemmathree's prefill uses bidirectional attention within images and causal attention for text, caching the full prompt before the gate and the compacted prompt after it; decoding reuses these caches and appends generated K/V without further deletion. Both diagrams show hard deletion at inference. The gate and deletion rule are the same in both architectures; only where the gate takes effect differs.}
    \label{fig:vmrt5gemma2}
\end{figure}

\textbf{In this paper, we broaden \dtm beyond its original setting and position it as a general framework for learned token deletion in transformers.} We characterize \dtm by three components (a learned delete gate, a soft-to-hard deletion paradigm, and a deletion-rate controller) and extend the framework along two dimensions: from text to visual-token sequences, and from bidirectional encoder architectures to decoder-only models. We instantiate these extensions in two general settings. First, \vmrtfivegemmatwo applies \dtm to visual tokens within the bidirectional multimodal encoder of the encoder--decoder T5Gemma 2 \citep{zhang2025t5gemma2seeingreading}, the setting closest to MrT5. Second, \vmrgemmathree extends the framework to visual-token compression in the decoder-only Gemma 3 \citep{gemmateam2025gemma3technicalreport}, where deletion occurs during prompt prefill and compacts the key--value caches reused during generation (Figure~\ref{fig:vmrt5gemma2}). Together with MrT5, these models form a broader \dtm family that applies the same compression method across distinct modalities and architectures.

We fine-tune both model architectures with LoRA \citep{hu2021loralowrankadaptationlarge} on ten benchmarks, from document and chart reading to counting, hallucination detection, and grounding, and compare \dtm with random deletion and two attention-based selectors that receive the same fine-tuning. \textbf{\dtm shows particularly strong gains under aggressive compression on tasks requiring fine-grained visual reading, including document, chart, and scene-text understanding.} At matched 80\% visual token deletion, which removes 58--63\% of language-model FLOPs on document, chart, and scene-text tasks, the encoder--decoder \vmrtfivegemmatwo achieves 57.5 ANLS on DocVQA and 63.0 relaxed accuracy on ChartQA, compared with 49.5 and 51.8 for our adapted \fastv, and the decoder-only \vmrgemmathree extends these gains even though its gate cannot see the question.

Collectively, our contributions are:
\textbf{(1) \dtm for the vision modality.} We apply \dtm to visual tokens while protecting text, and show on ten benchmarks and three backbones that its advantage over attention-based and random deletion grows with the deletion rate, both when each deployment budget has its own fine-tuned checkpoint and when one checkpoint serves budgets set at inference.
\textbf{(2) \dtm for decoder-only models.} We extend the delete gate from encoders to decoder-only models by compacting the prompt and its per-layer caches during prefill, and show that the gains carry over to Gemma 3.
\textbf{(3) \dtm as a general framework.} We identify the components that define \dtm independently of its byte-level origin and show that they transfer across modality and architecture, with analyses that characterize where the framework is most effective.

\section{Related Work}
\label{sec:related_work}

\paragraph{Non-Learned Visual Compression.} 
Non-learned approaches specify the compression rule without training a separate selector.
These rules can operate before the multimodal language backbone or within it.
VisionZip \citep{yang2026visionziplongerbetternecessary} selects and merges visual features before language-model processing using attention and feature similarity.
Within the backbone, \fastv \citep{chen2024imageworth12tokens} ranks visual tokens using received attention, while SparseVLM \citep{zhang2025sparsevlmvisualtokensparsification} uses selected text tokens to guide pruning and recycles information from discarded tokens.
PyramidDrop \citep{xing2025pyramiddropacceleratinglargevisionlanguage} follows a similar attention-based pruning principle, distributing token removal across multiple depths with a progressively shrinking visual token budget during training and inference.
FEATHER \citep{endo2025featherthrottlerevisitingvisual} addresses positional effects through RoPE-free scoring, spatial coverage, and two pruning stages.
LLaVolta \citep{chen2024efficientlargemultimodalmodels} instead compresses visual context through nonparametric average pooling inside the language model and studies schedules that change compression over training, whereas explicit merging methods such as Token Merging (ToMe; \citealt{bolya2023tokenmergingvitfaster}) combine token representations directly. These methods use fixed compression rules, although their surrounding models may be fine-tuned.

\paragraph{Learned Visual Compression.}
Learned approaches optimize the selection function through training, with precedents in both vision and language. DynamicViT \citep{rao2021dynamicvitefficientvisiontransformers} trains importance predictors with differentiable attention masking and progressively sparsifies vision-transformer tokens. VisionSelector \citep{zhu2026visionselectorendtoendlearnablevisual} trains a scorer before the language model using differentiable top-$k$ selection. LearnPruner \citep{takezoe2026learnprunerrethinkingattentionbasedtoken} combines a learned visual scorer after the vision encoder with text-guided selection at an intermediate language-model layer. In language, MrT5 \citep{kallini2025mrt5dynamictokenmerging} introduced \dtm for byte-level models (Section~\ref{sec:background-mrt5}), which we extend to visual tokens and decoder-only models. Unlike selectors placed before the language model, \dtm deletes tokens inside the backbone once, after attention has allowed retained states to absorb information from tokens that are later removed.

\section{Background}
\label{sec:background}

\subsection{T5Gemma 2 and Gemma 3}
\label{sec:background-models}

Gemma 3 is a decoder-only transformer \citep{gemmateam2025gemma3technicalreport}. To create T5Gemma 2, \citet{zhang2025t5gemma2seeingreading} initialize both the encoder and decoder stacks from pretrained Gemma 3 checkpoints and train the resulting encoder--decoder model with UL2 \citep{tay2023ul2unifyinglanguagelearning}. Both architectures process images with a SigLIP vision encoder \citep{zhai2023sigmoidlosslanguageimage}, then pool and project its representations into 256 visual tokens in the language model's input embedding space. In T5Gemma 2, these tokens enter a bidirectional encoder together with the text prompt, and each decoder layer attends jointly to its preceding decoder states and to all final encoder states under a single normalization, which \citet{zhang2025t5gemma2seeingreading} call \emph{merged attention}. In Gemma 3, the visual tokens are part of the prompt of a single causal stack; attention is bidirectional within an image but causal elsewhere, so image tokens cannot attend to a question that follows the image. Both models interleave five sliding-window (local) attention layers with one global-attention layer; indexing layers from 0, with layer 0 being the input states, layers $6,12,18,\ldots$ are global. Appendix~\ref{app:model-architectures} provides configurations, attention windows, and architecture diagrams for all three models.

\subsection{MrT5 and Dynamic Token Merging}
\label{sec:background-mrt5}

MrT5 \citep{kallini2025mrt5dynamictokenmerging} introduced \dtm to shorten the long input sequences of the byte-level encoder--decoder model ByT5 \citep{xue2022byt5tokenfreefuturepretrained}. A delete gate is placed before a fixed encoder layer $k$: layers $1,\ldots,k-1$ contextualize the full sequence, and layers $k$ onward process only the tokens that the gate retains. Let $\mathbf H\in\mathbb R^{n\times d}$ contain the $n$ hidden states entering layer $k$. The gate maps them to scores
\begin{equation}
\mathbf g=c\,\sigma\!\left[-\bigl(\operatorname{RMSNorm}(\mathbf H)\,\mathbf w+b\mathbf 1_n\bigr)\right]\in(c,0)^n,\qquad c<0,
\label{eq:delete-gate}
\end{equation}
where $\mathbf w\in\mathbb R^{d}$ and $b\in\mathbb R$ are learned, $\operatorname{RMSNorm}$ \citep{zhang2019rootmeansquarelayer} has its own learned scale, $\sigma$ is the sigmoid function, and $c$ is a fixed constant. We negate the argument of $\sigma$ so that a larger projection favors retention; this is equivalent to MrT5's parameterization up to the signs of $\mathbf w$ and $b$. The gate adds only $2d+1$ parameters and negligible FLOPs (Appendix~\ref{app:vdtm_flops_appendix}). A score $g_j$ near zero preserves token $j$, while a score near $c$ suppresses it.

During training, \dtm deletes tokens \emph{softly}: each score is added to the attention logits of the corresponding key in every attention layer after the gate, including the decoder's attention to encoder states. For query $i$ and key $j$ in one attention head,
\begin{equation}
\widetilde s_{ij}=\frac{\mathbf q_i^\top\mathbf k_j}{\sqrt{d_h}}+M_{ij}+g_j,
\qquad
A_{ij}=\frac{\exp(\widetilde s_{ij})}{1+\sum_u\exp(\widetilde s_{iu})},
\label{eq:gated-attention}
\end{equation}
where $\mathbf q_i,\mathbf k_j\in\mathbb R^{d_h}$ are the query and key after normalization and positional transforms, and $M_{ij}\in\{0,-\infty\}$ is the host model's attention mask (causal or bidirectional, local or global). Soft deletion emulates removal while keeping the sequence intact, so the task loss trains the gate jointly with the model. Following MrT5, attention uses $\operatorname{softmax}_1$ \citep{miller_attention_2023}, whose additional 1 in the denominator lets attention mass dissipate when every key is suppressed; a standard softmax would be invariant to a uniform bias. At inference, \dtm deletes tokens \emph{hard}: states with $g_j\leq\tau$, for a threshold $c<\tau<0$, are physically removed, so all later layers process a shorter sequence (Figure~\ref{fig:soft-hard-deletion-figure} in Appendix~\ref{app:soft-and-hard-deletion} illustrates both). 

The gate is encouraged to delete by the regularizer $\mathcal L_G=\frac1n\sum_j g_j$, giving the objective $\mathcal L=\mathcal L_{\mathrm{CE}}+\alpha_t\mathcal L_G$. To reach a target deletion rate $\delta$, MrT5 adjusts $\alpha_t$ with a controller resembling a proportional--integral (PI) controller from classical control theory, driven by the error $e_t=\delta-r_t$ between the target and the observed rate $r_t=\frac1n\sum_j\mathbf 1_{[g_j\leq\tau]}$ (Equation~\ref{eq:pid} without its D term). MrT5's diagnostic tasks show that the gate learns to merge relevant context into retained tokens, not only to discard irrelevant ones, which gives the method its name.

\section{Extending the DTM Framework}
\label{sec:extending-dtm}

MrT5 is one instance of a more general recipe. We define a \textbf{DTM model} as a transformer with three components: \textbf{(i)} a learned delete gate that scores contextualized hidden states at a fixed layer boundary (Equation~\ref{eq:delete-gate}); \textbf{(ii)} soft deletion through attention biases during training and hard deletion at inference (Equation~\ref{eq:gated-attention}); and \textbf{(iii)} a mechanism for targeting a specific deletion rate, typically through a regularizer (Equation~\ref{eq:pid}). None of these components refers to bytes or requires an encoder. What differs across instances is which tokens may be deleted and the layer at which sequences are shortened (Table~\ref{tab:dtm-instances}); we instantiate \dtm for visual tokens and decoder-only models below.

\begin{table}[t]
\centering
\caption{\textbf{Three instances of the DTM framework.} All share the delete gate, soft-to-hard deletion, and controlled deletion objective (Section~\ref{sec:background-mrt5}). MrT5 \citep{kallini2025mrt5dynamictokenmerging} uses PI control with full fine-tuning; our instances use PID control with LoRA. Only the tokens eligible for deletion and where the shortened sequence takes effect differ across instances.}
\footnotesize
\setlength{\tabcolsep}{3.1pt}
\begin{tabular}{@{}llllll@{}}
\toprule
Instance & Backbone & Deletes & Gate after & Sees question & Hard deletion shortens \\
\midrule
\mrtfive & ByT5 (enc.--dec.) & Bytes & Encoder layer 3 & --- & Encoder, cross-attn. \\
\vmrtfivegemmatwo & T5Gemma 2 (enc.--dec.) & Visual & Encoder layer 5 & Yes & Encoder, merged attn. \\
\vmrgemmathree & Gemma 3 (dec.-only) & Visual & Layer 5 (prefill) & No & Prompt, KV cache \\
\bottomrule
\end{tabular}
\label{tab:dtm-instances}
\end{table}

\subsection{Application to the Vision Modality}
\label{sec:dtm-vision}

To apply \dtm to images, we restrict deletion to visual tokens. The gate scores every visual hidden state entering layer $k$; text tokens and image-boundary markers receive $g_j=0$ and are never deleted, so the prompt reaches all later layers intact. We compute the regularizer and the observed deletion rate specifically over the \emph{visual tokens} $\mathcal V_t$ of each batch, $\mathcal L_G=|\mathcal V_t|^{-1}\sum_{j\in\mathcal V_t}g_j$, so the target $\delta$ represents the fraction of visual tokens removed, rather than text and vision tokens together. In \vmrtfivegemmatwo, the gate sits in the bidirectional encoder, where visual states can attend to the question before they are scored. As in MrT5, hard deletion shortens both the remaining encoder layers and the encoder keys and values that every decoder layer reads through merged attention (Figure~\ref{fig:vmrt5gemma2}a). Retained tokens keep their original position indices.

\subsection{Application to Decoder-Only Models}
\label{sec:dtm-decoder}

MrT5's gate operates in an encoder, whose output is computed once and stays fixed during generation. A decoder-only model instead processes the prompt and the answer in one causal stack, reusing cached keys and values at every generation step. We therefore apply \dtm during \emph{prompt prefill} (Figure~\ref{fig:vmrt5gemma2}b). Layers $1,\ldots,k-1$ process and cache the full prompt; the gate then scores the visual prompt states, and hard deletion removes the selected states so that layers $k,\ldots,L$ compute and cache only the compacted prompt. Generated tokens are never deleted. Deletion thus reduces prefill computation in every later layer, along with the cache memory and attention cost of each decoding step. Retained tokens keep their original positions, and masks are evaluated in the original position space, so each unpruned pair's causal and local-window visibility is unchanged. During training, soft deletion biases later layers' attention for both prompt and answer queries.

The more fundamental difference lies in what the gate can see. Gemma 3's image tokens attend bidirectionally to one another but cannot attend to a question that follows the image, so the gate in \vmrgemmathree scores visual states contextualized by the image alone, whereas the gate in \vmrtfivegemmatwo scores question-conditioned states. This setting therefore tests whether \dtm's gate depends on the prompt or on task-trained properties of the visual states themselves (Section~\ref{sec:matched-results}).

\subsection{Other Improvements to DTM}
\label{sec:dtm-improvements}

\paragraph{Derivative control.}
Task fine-tuning favors retaining visual information, pushing the deletion rate away from its target. We add a \textcolor{pidD}{derivative (D)} term, which responds to recent changes in the error, to MrT5's \textcolor{pidP}{proportional (P)} and \textcolor{pidI}{integral (I)} terms:
\begin{equation}
\alpha_{t+1}=\max\Bigl(0,\,
\textcolor{pidP}{\underbrace{k_p\bar e_t}_{\mathrm{P}}}
+\textcolor{pidI}{\underbrace{k_i\textstyle\sum_{u=1}^{t}e_u}_{\mathrm{I}}}
+\textcolor{pidD}{\underbrace{k_d(e_t-e_{t-\omega})}_{\mathrm{D}}}
\Bigr),
\qquad \bar e_t=\rho\,\bar e_{t-1}+(1-\rho)\,e_t,
\label{eq:pid}
\end{equation}
where $t$ counts controller updates and $\omega$ is a lookback.
Thus, when the model is deleting too few tokens and the error is increasing, the derivative term increases $\alpha_t$ more quickly.
As in MrT5, $\rho=0.9$ and the non-negative clamp prevents the regularizer from adding pressure toward retention. We use \textcolor{pidP}{$k_p=1$}, \textcolor{pidI}{$k_i=10^{-5}$}, \textcolor{pidD}{$k_d=0.5$}, $\omega=10$, and $\alpha_0=0.1$ for all backbones. In an ablation across five tasks and four deletion targets, PID kept the observed deletion rate closer to its target than the PI controller in 19 of 20 task--target pairs. This tighter control improved task performance most at the 80\% target, where PID outperformed PI on all five tasks by 0.81 points on average; at 20--60\%, task-score differences were small and did not consistently favor PID (Appendix~\ref{app:pid}).

\paragraph{Threshold-shifted biases for answer queries.}
After hard deletion, the gate continues to reweight the visual tokens that remain.
During input processing (encoder attention in \vmrtfivegemmatwo and prompt attention in \vmrgemmathree), we retain the original gate bias $g_j$ on visual keys.
During answer generation, we instead use $\min(g_j-\tau,0)$ for visual keys, while text keys receive no gate bias.
Thus, visual keys with $g_j \ge \tau$ incur no gate penalty when the model generates its answer, whereas any retained key with $g_j < \tau$, which can occur under fixed-budget deletion, remains downweighted.

\paragraph{Initial deletion-rate calibration.}
The original \dtm implementation initializes the gate bias to $b=1$
\citep{kallini2025mrt5dynamictokenmerging}. We instead apply a one-time calibration of $b$ on the first batch so that a chosen fraction $\delta_{\mathrm{init}}$ of visual tokens starts below the threshold; we found that this calibration improves accuracy at 60\% and 80\% targets (Appendix~\ref{app:gate-initialization}).

\paragraph{Other settings.}
We adapt the backbone with LoRA rather than MrT5's full fine-tuning, and omit MrT5's attention-score regularizer, which did not improve accuracy in our setting (Appendix~\ref{app:attn-reg}). We use $c=-22$ and $\tau=-7$ throughout.

\section{Experiments}
\label{sec:experiments}

\paragraph{Models and tasks.}
We use the encoder--decoder T5Gemma 2 1B--1B and the decoder-only Gemma 3 4B as our primary backbones and T5Gemma 2 270M--270M as a smaller encoder--decoder model. We evaluate on ten benchmarks, covering document and infographic reading (DocVQA; \citealt{mathew2021docvqadatasetvqadocument}; InfoVQA; \citealt{mathew2021infographicvqa}), chart understanding (ChartQA; \citealt{masry2022chartqabenchmarkquestionanswering}), scene-text reading (TextVQA; \citealt{singh2019vqamodelsread}), general and assistive visual question answering (VQAv2; \citealt{goyal2017makingvvqamatter}; GQA; \citealt{hudson2019gqanewdatasetrealworld}; VizWiz; \citealt{gurari2018vizwizgrandchallengeanswering}), counting (TallyQA; \citealt{acharya2019tallyqa}), object hallucination detection (POPE; \citealt{li2023evaluatingobjecthallucinationlarge}), and referring-expression grounding (RefCOCO+; \citealt{yu2016modelingcontextreferringexpressions}). We report each task's own metric on a 0--100 scale. POPE evaluates checkpoints trained on VQAv2; the other benchmarks use task-specific fine-tuning for each pruning method. Every method, including the unpruned references, is fine-tuned with LoRA on the attention and MLP projections at learning rate $10^{-4}$, with the vision encoder frozen; the \dtm gate is trained jointly with the adapters. Training steps vary by task, so the results are not equalized by training compute; Appendix~\ref{app:experimental-details} lists the evaluation splits, task metrics, training budgets, hyperparameters, and evidence region sources.

\paragraph{Baselines.}
We compare \dtm with \random deletion and two attention-based selectors, \fastv \citep{chen2024imageworth12tokens} and \phir from FEATHER \citep{endo2025featherthrottlerevisitingvisual}. All methods delete visual tokens after five layers (L5, i.e., $k=6$) and receive the same task fine-tuning. The adapted \fastv and \phir learn backbone parameters while their ranking rules remain fixed, whereas \dtm also learns the ranking function. \Random deletion controls for adaptation to shorter visual sequences without informed selection, and unpruned fine-tuned backbones serve as \emph{Base} references. Our \fastv ranks visual keys by received attention averaged over heads and all valid input queries (prompt queries for Gemma 3). Our \phir applies FEATHER's RoPE-free scoring, which counters the positional bias that RoPE's distance-dependent decay induces in \fastv, with the same query aggregation but without FEATHER's uniform sampling or two-stage procedure \citep{endo2025featherthrottlerevisitingvisual}. Appendix~\ref{app:query-aggregation} compares all-query scoring with text-query variants on all three backbones, including the criterion of LearnPruner's \citep{takezoe2026learnprunerrethinkingattentionbasedtoken} second stage; all-query scoring wins in most tested comparisons.

\paragraph{Hyperparameter tuning across all models.}
We fine-tune every method on DocVQA with deletion after each of layers 3--7 (Appendix~\ref{sec:depth-analysis}). L5 is the best layer for \fastv and \phir on T5Gemma 2 1B--1B, and both drop sharply at L6, the first global-attention layer; \dtm, which reads hidden states rather than attention weights, leads at every layer on all three backbones at 40--80\% deletion. We evaluate all models at L5 on the basis of our T5Gemma 2 layer sweep results.

\paragraph{Two evaluation settings.}
The two settings correspond to two ways a fine-tuned model can be deployed. In the \emph{matched-rate} setting, a model provider serves several deployment targets, such as device classes or latency tiers, whose compute budgets are fixed in advance, and fine-tunes a separate checkpoint for each; we therefore train separate models toward 20\%, 40\%, 60\%, and 80\% visual token deletion and evaluate each at its own training target, measuring the best accuracy each method reaches at a given budget. In the \emph{fixed-checkpoint} setting, a single fine-tuned model serves budgets that are set at inference, for example per request or as load changes; we evaluate each checkpoint at deletion rates from 10\% to 90\% in 10-point increments, focusing on models trained at 60\%. At inference, every method enforces the requested budget by ranking visual tokens and deleting $\lfloor \delta \cdot n_v\rfloor$ of the $n_v$ visual tokens, which for \dtm separates the budget from the gate's natural threshold-crossing rate (Appendix~\ref{app:adaptive-budgets}).

\paragraph{Computational cost.}
We compute analytical forward-pass FLOPs of the language backbone and vocabulary projection from the recorded sequence lengths before and after deletion, excluding the vision encoder and image projector (Appendix~\ref{app:flops}). Because each method deletes the same number of tokens after the same layer, all four methods share the same backbone cost at a given deletion rate and differ only in how they select tokens, which costs little: the \dtm gate adds one $d$-dimensional projection per visual token, \fastv reuses the preceding layer's attention weights, and \phir adds one RoPE-free score computation in that layer (Appendix~\ref{app:flops_pruning}). Averaged over the ten benchmarks, deleting 40\% and 80\% of visual tokens saves 29.4\% and 58.4\% of FLOPs on T5Gemma 2 1B--1B, 30.9\% and 61.4\% on Gemma 3 4B, and 27.4\% and 53.1\% on T5Gemma 2 270M--270M (Table~\ref{tab:matched}; Appendix~\ref{app:flops-details}); savings are smaller on RefCOCO+, whose text prompts are longer.

\section{Results and Analysis}
\label{sec:analysis}

\subsection{Matched-rate Evaluation Setting}
\label{sec:matched-results}

\begin{table}[t]
\centering
\caption{\textbf{Matched-rate evaluation at 40\% and 80\% visual token deletion for T5Gemma 2 (1B--1B) and Gemma 3 (4B), deletion after layer 5.} Each score comes from a model trained and evaluated at the given deletion rate $\delta$; Base is the unpruned model fine-tuned in the same way. Bold marks the best method per column within each model; each task uses its own metric (Section~\ref{sec:experiments}). The gray row gives the percentage of language-model FLOPs saved, which is the same for every method. Figures~\ref{fig:t5-train60} and~\ref{fig:gemma-train60} include 60\%, and Figures~\ref{fig:t5-matched}--\ref{fig:270m-matched} (appendix) show 20--80\%. \dtm's lead grows with the deletion rate on both architectures and is largest on DocVQA, ChartQA, and TextVQA.}
\label{tab:matched}
\setlength{\tabcolsep}{2.6pt}
\setlength{\aboverulesep}{0pt}\setlength{\belowrulesep}{0pt}%
\def\vd{\hspace{\tabcolsep}{\color{black!22}\vrule width 0.4pt}\hspace{\tabcolsep}}%
\def\fl#1{{\color{black!55}\textit{#1}}}%
\renewcommand{\arraystretch}{1.12}
\resizebox{\linewidth}{!}{%
\begin{tabular}{@{}lcc@{\vd}cc@{\vd}cc@{\vd}cc@{\vd}cc@{\vd}cc@{\vd}cc@{\vd}cc@{\vd}cc@{\vd}cc@{}}
\toprule
\textit{T5Gemma 2} & \multicolumn{2}{c@{\vd}}{DocVQA} & \multicolumn{2}{c@{\vd}}{ChartQA} & \multicolumn{2}{c@{\vd}}{InfoVQA} & \multicolumn{2}{c@{\vd}}{TextVQA} & \multicolumn{2}{c@{\vd}}{VQAv2} & \multicolumn{2}{c@{\vd}}{VizWiz} & \multicolumn{2}{c@{\vd}}{GQA} & \multicolumn{2}{c@{\vd}}{TallyQA} & \multicolumn{2}{c@{\vd}}{POPE} & \multicolumn{2}{c}{RefCOCO+} \\[3pt]
$\delta$ & 40\% & 80\% & 40\% & 80\% & 40\% & 80\% & 40\% & 80\% & 40\% & 80\% & 40\% & 80\% & 40\% & 80\% & 40\% & 80\% & 40\% & 80\% & 40\% & 80\% \\
\midrule
Base & \multicolumn{2}{c@{\vd}}{82.0} & \multicolumn{2}{c@{\vd}}{69.7} & \multicolumn{2}{c@{\vd}}{47.2} & \multicolumn{2}{c@{\vd}}{72.0} & \multicolumn{2}{c@{\vd}}{77.5} & \multicolumn{2}{c@{\vd}}{71.4} & \multicolumn{2}{c@{\vd}}{71.6} & \multicolumn{2}{c@{\vd}}{74.9} & \multicolumn{2}{c@{\vd}}{84.8} & \multicolumn{2}{c}{79.9} \\
\cmidrule(l){1-21}
\dtm & \textbf{78.4} & \textbf{57.5} & 66.2 & \textbf{63.0} & 43.0 & \textbf{34.1} & 70.8 & \textbf{67.8} & \textbf{77.3} & \textbf{75.6} & 69.8 & \textbf{69.6} & \textbf{71.2} & \textbf{70.0} & \textbf{73.9} & \textbf{72.0} & 84.7 & \textbf{84.0} & \textbf{79.6} & \textbf{75.3} \\
\fastv & 75.9 & 49.5 & \textbf{67.3} & 51.8 & \textbf{43.2} & 31.4 & \textbf{71.3} & 63.3 & 76.6 & 72.4 & \textbf{70.4} & 66.9 & 71.0 & 66.5 & 73.2 & 69.2 & \textbf{85.2} & 82.3 & 77.9 & 73.3 \\
\phir & 74.6 & 47.5 & 66.5 & 55.5 & 42.9 & 30.7 & 70.1 & 58.4 & 76.3 & 71.6 & 70.3 & 64.0 & 70.9 & 67.4 & 72.4 & 69.2 & 84.8 & 81.9 & 78.2 & 71.5 \\
\Random & 66.8 & 37.2 & 65.9 & 51.0 & 37.1 & 27.7 & 68.9 & 55.1 & 76.0 & 72.4 & 70.1 & 67.1 & 69.9 & 66.9 & 71.6 & 68.7 & 84.6 & 82.3 & 78.7 & 70.0 \\
\fl{FLOPs saved} & \fl{29.6} & \fl{58.9} & \fl{29.4} & \fl{58.3} & \fl{29.5} & \fl{58.7} & \fl{30.2} & \fl{60.0} & \fl{30.5} & \fl{60.6} & \fl{30.2} & \fl{59.9} & \fl{30.3} & \fl{60.1} & \fl{30.5} & \fl{60.6} & \fl{30.5} & \fl{60.5} & \fl{23.5} & \fl{46.6} \\
\midrule
\textit{Gemma 3} & & & & & & & & & & & & & & & & & & & & \\
Base & \multicolumn{2}{c@{\vd}}{84.5} & \multicolumn{2}{c@{\vd}}{76.5} & \multicolumn{2}{c@{\vd}}{56.7} & \multicolumn{2}{c@{\vd}}{78.7} & \multicolumn{2}{c@{\vd}}{78.5} & \multicolumn{2}{c@{\vd}}{73.7} & \multicolumn{2}{c@{\vd}}{69.2} & \multicolumn{2}{c@{\vd}}{73.5} & \multicolumn{2}{c@{\vd}}{84.5} & \multicolumn{2}{c}{78.2} \\
\cmidrule(l){1-21}
\dtm & \textbf{80.6} & \textbf{58.8} & \textbf{77.2} & \textbf{68.8} & \textbf{51.7} & \textbf{40.0} & \textbf{77.5} & \textbf{75.1} & 77.7 & \textbf{76.7} & 72.7 & 70.8 & \textbf{69.3} & \textbf{68.2} & 72.9 & \textbf{71.6} & \textbf{85.4} & \textbf{84.0} & \textbf{78.3} & \textbf{72.7} \\
\fastv & 78.1 & 48.9 & 74.8 & 62.2 & 51.4 & 36.4 & 76.6 & 71.2 & \textbf{77.8} & 73.8 & \textbf{73.6} & \textbf{72.0} & 68.9 & 64.6 & 73.6 & 69.9 & 84.8 & 80.4 & 78.2 & 69.5 \\
\phir & 77.6 & 47.7 & 75.2 & 61.9 & 49.1 & 34.2 & 75.8 & 69.3 & 77.6 & 74.1 & 73.4 & 70.3 & 68.8 & 64.6 & \textbf{73.7} & 70.6 & 84.2 & 81.4 & 77.7 & 69.4 \\
\Random & 68.9 & 42.1 & 70.8 & 58.4 & 47.6 & 34.3 & 73.3 & 65.3 & 77.0 & 73.6 & 73.0 & 69.7 & 68.4 & 64.4 & 72.7 & 69.6 & 83.8 & 81.7 & 76.1 & 69.6 \\
\fl{FLOPs saved} & \fl{31.1} & \fl{62.0} & \fl{30.8} & \fl{61.3} & \fl{31.0} & \fl{61.7} & \fl{31.7} & \fl{63.0} & \fl{32.0} & \fl{63.7} & \fl{31.6} & \fl{63.0} & \fl{31.7} & \fl{63.1} & \fl{32.0} & \fl{63.6} & \fl{31.9} & \fl{63.6} & \fl{24.8} & \fl{49.5} \\
\bottomrule
\end{tabular}%
}
\end{table}

Table~\ref{tab:matched} compares the methods at 40\% and 80\% deletion, which save about 30\% and 60\% of FLOPs. At 40\%, \dtm achieves small gains: \vmrtfivegemmatwo leads on five of ten benchmarks and is within 1.1 points of the strongest baseline on every benchmark. On DocVQA, \dtm leads by 2.5. At 80\%, \dtm leads on all ten benchmarks by 3.76 points on average. \textbf{The largest gains under aggressive compression occur on tasks that require reading fine visual detail:} \vmrtfivegemmatwo{} scores 57.5 on DocVQA, 63.0 on ChartQA, and 67.8 on TextVQA, 8.0, 7.5, and 4.5 points above the strongest baseline. From 40\% to 80\%, \dtm loses only 3.3 points on ChartQA and 3.0 on TextVQA, whereas \fastv loses 15.5 and 7.9. On VQAv2, VizWiz, GQA, TallyQA, and POPE, the attention-based selectors are at most 0.6 points above \random at 80\%, consistent with \citet{wen-etal-2025-token}, while \dtm leads by 1.7--3.2.

The same pattern holds for the decoder-only \vmrgemmathree, whose gate cannot see the question (Table~\ref{tab:matched} bottom). It leads on seven of ten benchmarks at 40\%, trailing by at most 0.9 points, and on nine at 80\%, where it averages 3.55 points above the strongest baseline and exceeds \fastv by 9.9, 6.6, and 3.9 points on DocVQA, ChartQA, and TextVQA; \fastv remains stronger on VizWiz (72.0 vs.\ 70.8). The two instances differ in where deletion acts, in a bidirectional encoder whose visual states have read the question or in a causal prefill whose visual states have not, yet \dtm's advantage has the same shape in both, consistent with the gate not requiring question-conditioned states. At 60\%, \dtm leads on all ten T5Gemma 2 benchmarks and six Gemma 3 benchmarks (Figures~\ref{fig:t5-train60} and~\ref{fig:gemma-train60}); at 20\%, it is within 1.9 points of the strongest baseline on both backbones (Appendix~\ref{app:matched-all}). On T5Gemma 2 270M--270M, the advantage emerges later: \dtm leads on three, seven, and eight benchmarks at 40\%, 60\%, and 80\% (Appendix~\ref{app:small-model}). Nonetheless, these results demonstrate that \textbf{\dtm is the best choice among the evaluated methods for most tasks when a targeted computation budget is known and fixed.}

\subsection{Fixed-checkpoint Evaluation Setting}
\label{sec:fixed-results}

\begin{figure}[t]
\centering
\includegraphics[width=\linewidth]{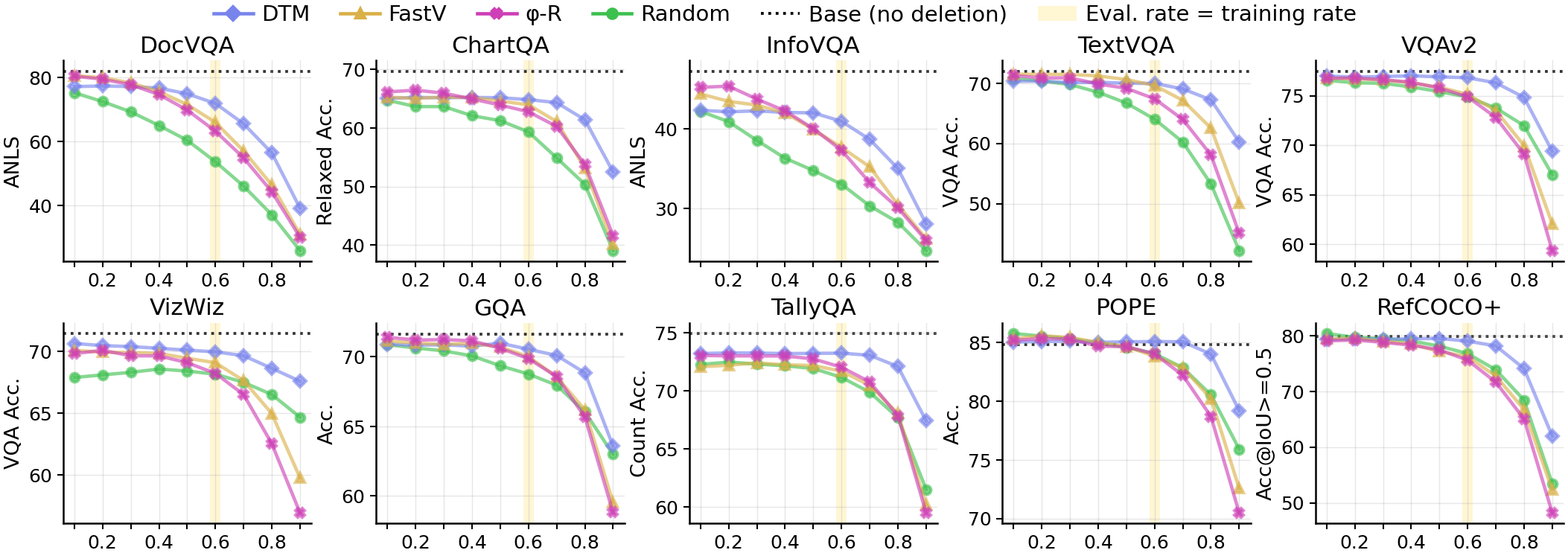}
\setlength{\abovecaptionskip}{-3.5pt}
\caption{\textbf{One 60\%-trained T5Gemma 2 1B--1B checkpoint (\vmrtfivegemmatwo) across inference budgets, deletion after layer 5.} Each curve holds one task- and method-specific checkpoint fixed and evaluates it at deletion rates from 10\% to 90\%; the training rate, 60\%, is highlighted. \dtm's advantage widens as the inference budget tightens beyond the training target.}
\label{fig:t5-train60}

\medskip

\includegraphics[width=\linewidth]{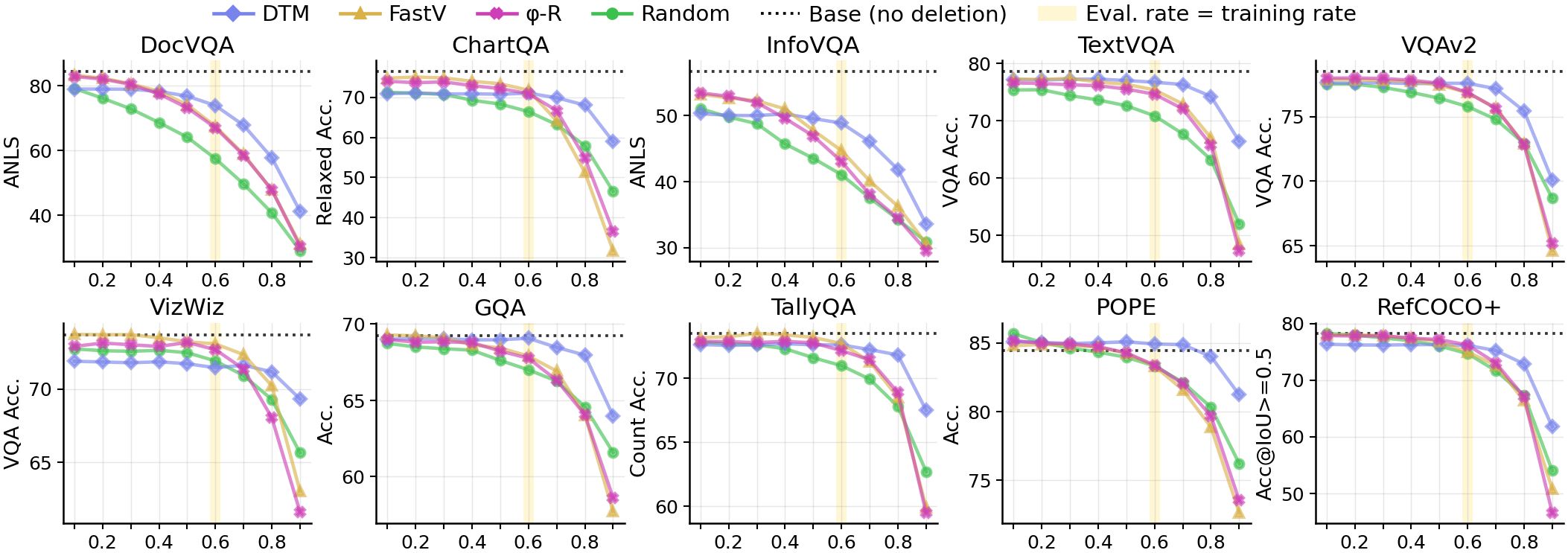}
\setlength{\abovecaptionskip}{-3.5pt}
\caption{\textbf{One 60\%-trained Gemma 3 4B checkpoint (\vmrgemmathree) decoder-only across inference budgets, deletion after five decoder layers during prefill.} Each curve holds one checkpoint fixed across deletion rates from 10\% to 90\%, with 60\% highlighted. \vmrgemmathree shows the same widening advantage at 80--90\% deletion.}
\label{fig:gemma-train60}

\end{figure}

Figure~\ref{fig:t5-train60} shows that \dtm retains its comparative advantage when the inference budget becomes tighter than the training target. For T5Gemma 2 1B--1B, \dtm leads all baselines on every benchmark at each tested deletion rate from 60\% through 90\%. At 80\% inference deletion, the 60\%-trained models score 56.5 on DocVQA, 61.4 on ChartQA, and 67.2 on TextVQA, compared with \fastv's 46.5, 53.1, and 62.6. This advantage extends to Gemma 3, where \dtm leads on every benchmark at 80\% and 90\% deletion (Figure~\ref{fig:gemma-train60}). On ChartQA at 90\%, it scores 59.0, compared with 46.6 for \random, the strongest baseline, and 31.8 for \fastv. The 270M--270M model also leads on all ten benchmarks at 80\% (Appendix~\ref{app:small-model}). Across training targets (Appendix~\ref{app:fixed-all}), the decoder-only model transfers best: a Gemma 3 \dtm checkpoint trained at only 20\% loses 2.5 points on average at 80\% deletion, compared with the checkpoint trained at 80\%, against 5.4--7.1 for the baselines, whereas on T5Gemma 2 1B--1B, \dtm's loss (7.4) is similar to the attention-based baselines (5.6--7.4).

Increasing the token budget does not necessarily recover the performance of a model adapted to that budget: on DocVQA, the 60\%-trained T5Gemma 2 \dtm model scores 77.4 at 20\% inference deletion, compared with 81.0 for the model trained at 20\%. Conversely, using a 60\%-trained checkpoint at 80\% deletion costs only 0.6, 0.2, and 0.2 points on average relative to training at 80\% for T5Gemma 2 1B--1B, Gemma 3 4B, and T5Gemma 2 270M--270M, respectively (Figure~\ref{fig:rate-transfer}, appendix). On Gemma 3 ChartQA, raising the training target from 20\% to 80\% improves performance at 90\% inference deletion from 49.3 to 63.0, but lowers it at 20\% deletion from 77.0 to 69.2. A fixed checkpoint is thus useful across budgets, but its training target still shapes the accuracy--cost trade-off; over the full 10--90\% range, mean normalized AUC is highest for 60\% training in both T5Gemma 2 sizes and 40\% in Gemma 3 (Appendix~\ref{app:budget-analysis}).

To interpret the fixed-checkpoint curves, we identify the largest tested deletion rate that preserves at least 90\% of each task's unpruned Base score (Table~\ref{tab:retention-budgets}). For both larger backbones trained at 60\%, \dtm reaches 50\% deletion on DocVQA and 70\% on ChartQA, compared with \fastv's 40\% and 60\%. On TextVQA, it reaches 80\% on both backbones, ten percentage points beyond \fastv, and on POPE it meets this criterion even at 90\%. However, neither \dtm model reaches 90\% of Base at any tested rate on InfoVQA, even though both become the strongest methods at high deletion; \fastv and \phir meet that threshold through 30\%. \textbf{The results show that \dtm is the best choice on most tasks when a single fine-tuned checkpoint must generalize to variable inference budgets}, although it can be a worse choice on certain benchmarks when preserving unpruned performance is the priority.

\subsection{Spatial Evidence Retention}
\label{sec:qualitative-analysis}

\begin{figure}[t]
\centering
\includegraphics[width=1.0\linewidth]{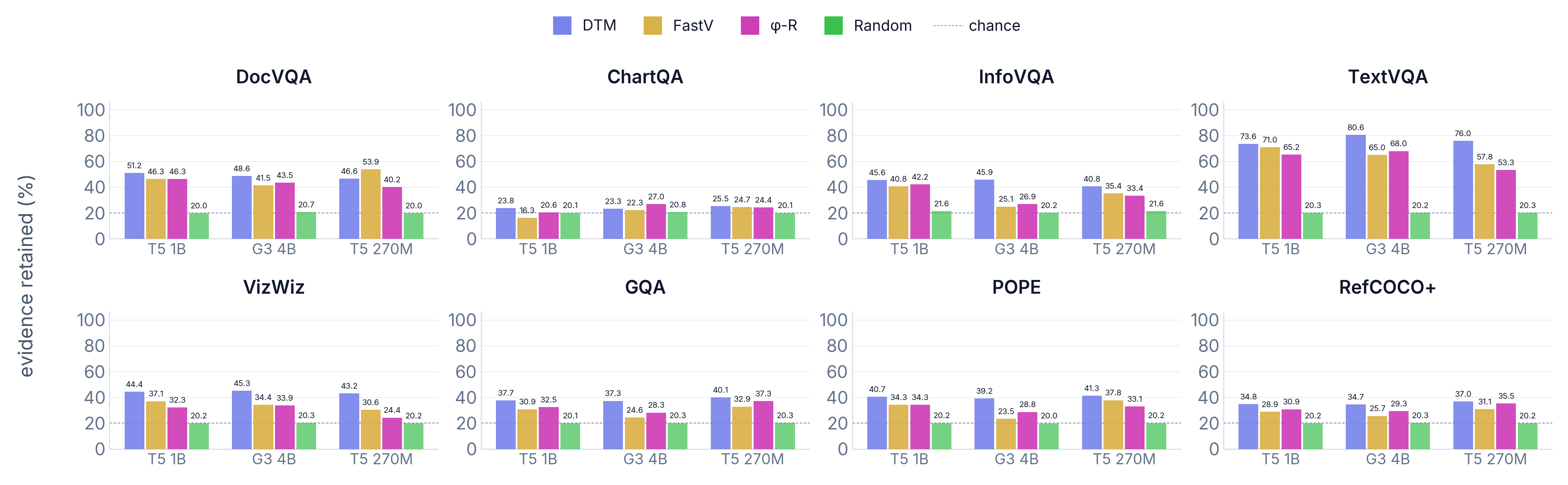}
\setlength{\abovecaptionskip}{-5pt}
\caption{\textbf{Spatial evidence retention at matched 80\% visual token deletion after layer 5.} Bars show the mean percentage of evidence patches retained on the same eligible questions for each model--dataset pair; dashed lines mark the expected retention of random deletion. T5 1B, T5 270M, and G3 4B denote T5Gemma 2 1B--1B, T5Gemma 2 270M--270M, and Gemma 3 4B. \dtm retains the most evidence in 22 of 24 comparisons.}
\label{fig:evidence-retained}
\end{figure}

We measure how much spatial evidence each method retains---the fraction of image patches overlapping task-relevant regions, identified from spatial annotations or grounding proxies (Appendix~\ref{app:evidence-rates})---and report results on the subset of questions for which such evidence regions are available. Figure~\ref{fig:evidence-retained} shows that \dtm achieves the highest mean evidence retention in 22 of the 24 model--dataset comparisons at 80\% deletion. Across 20--80\% deletion, absolute retention decreases as deletion increases while \dtm's advantage grows: weighting model--dataset pairs equally, its lead over \fastv increases from 2.8 to 4.6, 6.7, and 7.7 percentage points, and its lead over \phir from 2.2 to 7.7 points. However, spatial retention does not fully predict task performance. At 80\% deletion on Gemma 3 4B ChartQA, \phir retains more evidence than \dtm (27.0\% vs.\ 23.3\%) yet achieves lower relaxed accuracy on the question subset (70.6\% vs.\ 80.5\%); on T5Gemma 2 270M--270M DocVQA, \fastv retains more evidence (53.9\% vs.\ 46.6\%) but obtains lower ANLS (41.2 vs.\ 41.9). Consistent with \dtm's premise, retained tokens may already carry context from deleted ones, so spatial retention diagnoses token selection without establishing which information the model actually uses.

Across 15 randomly sampled ChartQA and 15 TextVQA questions at 80\% deletion (Appendix~\ref{app:qualitative}), \dtm concentrates its budget on chart titles, tick labels, and bar ends, and on the full ChartQA sets it keeps 57--67\% of text-label patches, against 39--61\% for the attention-based selectors (chance: 20\%). This is not a general preference for text, since the attention-based selectors cover more OCR text on DocVQA, but for answer-bearing content (Figure~\ref{fig:evidence-retained}), which is associated with correctness: where \dtm keeps more evidence than \fastv, it alone answers correctly on 13.5\% of questions, against 6.2\% for \fastv alone, although this comparison is correlational (Appendix~\ref{app:interp}).

\section{Conclusion}
\label{sec:conclusion}

MrT5 introduced Dynamic Token Merging to make byte-level language models more efficient. We have argued that \dtm is more general: a learned delete gate, soft-to-hard deletion, and a controlled deletion objective define a framework for token deletion in transformers, of which MrT5 is one instance. Extended with small modifications to visual tokens in both an encoder--decoder and a decoder-only VLM, these components give \dtm its strongest gains over attention-based and random deletion when visual token budgets are tight, particularly on document, chart, and scene-text tasks, and a model trained at one deletion rate remains effective across inference budgets. The decoder-only gate succeeds without seeing the question, and \dtm, whose gate reads hidden states rather than attention weights, keeps more task-relevant evidence and is less sensitive than attention-based selectors to whether the preceding layer uses local or global attention. Its gains are smaller under mild deletion and on some general VQA tasks. We hope that treating \dtm as a general framework encourages its application to further modalities and architectures, such as video, audio, and multi-image inputs, and its combination with other compression techniques.

\subsection*{AI use statement}

In this work, we used generative AI tools to implement methods: coding assistants wrote much of the code for our model implementations, baselines, training and evaluation pipelines, and analysis scripts, and helped execute experiments. We have not used generative AI tools to generate synthetic data sets, develop theoretical models or conceptual frameworks, formulate mathematical claims or proofs, or propose hypotheses; the research ideas and analyses were conceived by the authors. Additionally, we used generative AI tools to help rewrite and polish prose, draft and edit parts of the paper, create and edit code for figures and tables, check the analyses reported in the paper and write them up in full, locate research literature, and format references. We have reviewed all AI-assisted work: submitted code was manually inspected by multiple authors, all sources identified by AI were manually reviewed, and all AI-created or edited text was checked by the authors. We take responsibility for the final content of this work, including text, claims, or artifacts produced with the aid of generative AI.

\subsection*{Reproducibility statement}

Training hyperparameters and evaluation settings are described in Section~\ref{sec:experiments} and Appendix~\ref{app:experimental-details}, model configurations in Appendix~\ref{app:model-architectures}, and FLOPs calculations in Appendix~\ref{app:flops}. In the supplementary materials, we have included annotated code for \dtm, including \vmrtfivegemmatwo and \vmrgemmathree, and for our \fastv, \phir, and \random baselines, as well as dataset preprocessing, training and evaluation setups, task metrics, and FLOPs calculations; comments in the code refer to the corresponding equations and sections of the paper. Upon publication, we will also provide model checkpoints.

\clearpage

\appendix
\makeatletter
\setlength{\@fptop}{0pt}
\setlength{\@fpsep}{12pt plus 4pt}
\setlength{\@fpbot}{0pt plus 1fil}
\makeatother
\renewcommand{\floatpagefraction}{0.8}
\renewcommand{\bottomfraction}{0.7}
\setcounter{topnumber}{3}
\setcounter{bottomnumber}{2}
\setcounter{totalnumber}{5}

\section{DTM Mechanisms}
\label{app:vdtm_mechanisms}

\subsection{Soft and Hard Deletion}
\label{app:soft-and-hard-deletion}

\begin{figure}[!htbp]
    \centering
    \includegraphics[width=\linewidth]{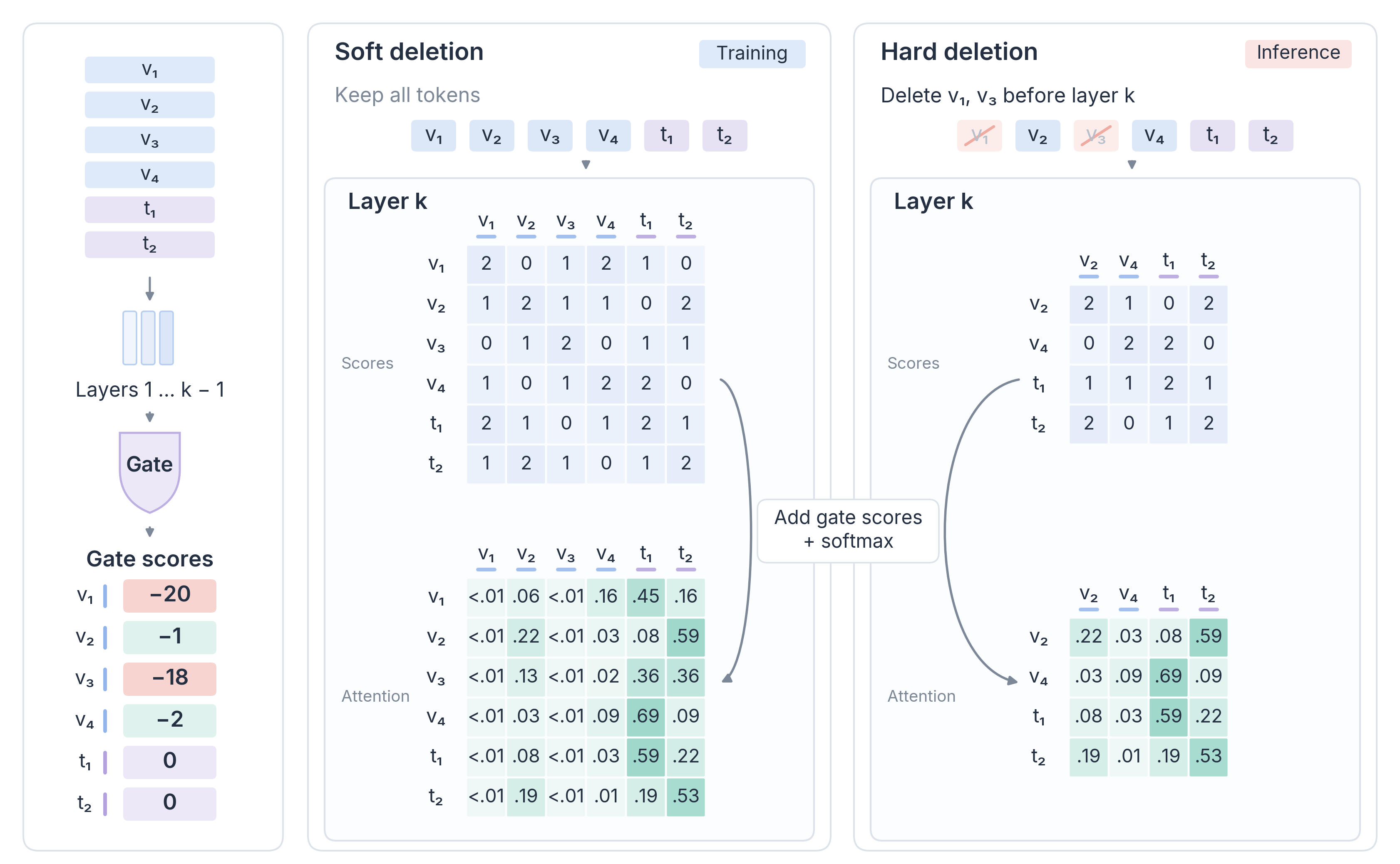}
    \caption{\textbf{Soft and hard visual token deletion in DTM.}
    Vision and text tokens, denoted as $s_{v_i}$ and $s_{t_i}$, are
    processed normally from layers $1$ through $k-1$. The hidden states
    before layer $k$ are entered into the deletion gate, producing a score
    for each token. During hard deletion, hidden states with gate scores
    less than $\tau$ are pruned; the gate scores are applied as biases to
    only the remaining pre-softmax attention scores. Soft deletion adds
    gate scores to key columns; hard deletion removes $s_{v_1}$ and $s_{v_3}$
    and retains the text tokens.}
    \label{fig:soft-hard-deletion-figure}
\end{figure}

\FloatBarrier %

\subsection{Initial Deletion-Rate Calibration}
\label{app:gate-initialization}

Our goal is to initialize the gate so that its measured deletion
rate is close to a chosen value $\delta_{\mathrm{init}}\in(0,1)$.
We achieve this by adjusting the scalar bias $b$ while holding
the initialized gate weights fixed.

Let $\mathcal V_0$ index the visual tokens used for calibration
during the first forward pass. For each token $j\in\mathcal V_0$,
let $\mathbf x_j\in\mathbb R^d$ be its normalized hidden state and
define its scalar projection score as
$z_j=\mathbf w^\top\mathbf x_j$, where $\mathbf w$ is the
initialized gate weight vector. From Equation~\ref{eq:delete-gate},
the gate score is
\begin{equation}
g_j=c\,\sigma[-(z_j+b)],
\qquad c<\tau<0.
\end{equation}
The deletion criterion compares this scalar gate score with the
threshold: $g_j\leq\tau$.

To determine how the bias affects this criterion, we rearrange it:
\begin{equation}
\begin{aligned}
g_j\leq\tau
&\iff \sigma[-(z_j+b)]\geq\tau/c \\
&\iff -(z_j+b)\geq\operatorname{logit}(\tau/c) \\
&\iff z_j\leq-\operatorname{logit}(\tau/c)-b,
\end{aligned}
\end{equation}
where $\operatorname{logit}(p)=\log[p/(1-p)]$ is the inverse
sigmoid. The first inequality reverses because $c<0$.
Thus, changing $b$ shifts the cutoff applied to the projection
scores $z_j$.

Let $q_0$ be the empirical $\delta_{\mathrm{init}}$-quantile
of the projection scores $\{z_j:j\in\mathcal V_0\}$, so $\delta_{\mathrm{init}}$ of these scores are at or below $q_0$.
For example, $\delta_{\mathrm{init}}=0.6$ selects the
60th percentile. We initialize the bias as
\begin{equation}
b_0=-\operatorname{logit}(\tau/c)-q_0.
\label{eq:gate-init}
\end{equation}
With this initialization, $g_j\leq\tau$ exactly when $z_j\leq q_0$,
so the deletion criterion is satisfied by the
desired fraction of calibration tokens. Calibration is performed only once and sets the starting value
of the trainable bias. 

\paragraph{Effect of calibration.}
Table~\ref{tab:gate-init} compares calibration with the default constant intercept on DocVQA and VizWiz, at all four training targets and two thresholds ($\tau=-7$ and $-11$). The constant intercept starts training at 84.4--99.6\% deletion regardless of the target, whereas calibration starts within 1.5 points of it in all 16 configurations. The effect on accuracy depends on the target: calibration improves the best validation score by 0.45 and 0.75 points on average at 60\% and 80\%, but has no consistent effect at 20\% and 40\% ($-0.13$ and $-0.05$). Its clearest benefit is therefore a predictable starting point, with accuracy gains only when the budget is tight.

\begin{table}[htbp]
\centering
\caption{\textbf{Calibrated initial deletion rate versus the default constant gate intercept on T5Gemma 2 1B--1B.} Step-0 deletion is the validation deletion rate measured before the first optimizer step; best scores are best in-training validation scores at the trained rate (L5, $c=-22$). The default intercept starts training at 84--99.6\% deletion regardless of the target, whereas calibration starts within 1.5 points of the target in all 16 configurations. Calibration improves the best score by 0.45 and 0.75 points on average at 60\% and 80\% targets but has no consistent effect at 20\% and 40\% ($-0.13$ and $-0.05$).}
\footnotesize
\setlength{\tabcolsep}{4pt}
\begin{tabular}{llrrrrrc}
\toprule
& & & \multicolumn{2}{c}{Step-0 deletion (\%)} & \multicolumn{2}{c}{Best score} & \\
\cmidrule(lr){4-5}\cmidrule(lr){6-7}
Task & Target & $\tau$ & Calibrated & Default & Calibrated & Default & $\Delta$ \\
\midrule
DocVQA & 20\% & $-7$  & 21.1 & 98.5 & 81.06 & 81.46 & $-0.40$ \\
DocVQA & 20\% & $-11$ & 21.1 & 84.4 & 81.36 & 81.07 & $+0.29$ \\
VizWiz & 20\% & $-7$  & 20.3 & 99.6 & 70.24 & 70.48 & $-0.24$ \\
VizWiz & 20\% & $-11$ & 20.3 & 93.4 & 69.86 & 70.02 & $-0.16$ \\
\midrule
DocVQA & 40\% & $-7$  & 41.5 & 98.5 & 78.25 & 78.15 & $+0.10$ \\
DocVQA & 40\% & $-11$ & 41.5 & 84.4 & 77.78 & 77.50 & $+0.28$ \\
VizWiz & 40\% & $-7$  & 40.4 & 99.6 & 69.94 & 70.79 & $-0.85$ \\
VizWiz & 40\% & $-11$ & 40.4 & 93.4 & 69.80 & 69.54 & $+0.26$ \\
\midrule
DocVQA & 60\% & $-7$  & 61.4 & 98.5 & 71.74 & 72.47 & $-0.72$ \\
DocVQA & 60\% & $-11$ & 61.3 & 84.4 & 70.00 & 68.78 & $+1.22$ \\
VizWiz & 60\% & $-7$  & 60.8 & 99.6 & 69.74 & 69.51 & $+0.23$ \\
VizWiz & 60\% & $-11$ & 61.0 & 93.4 & 70.10 & 69.01 & $+1.09$ \\
\midrule
DocVQA & 80\% & $-7$  & 81.3 & 98.5 & 57.90 & 56.18 & $+1.73$ \\
DocVQA & 80\% & $-11$ & 81.2 & 84.4 & 52.51 & 51.62 & $+0.89$ \\
VizWiz & 80\% & $-7$  & 81.0 & 99.6 & 69.63 & 69.17 & $+0.46$ \\
VizWiz & 80\% & $-11$ & 81.1 & 93.4 & 68.90 & 68.99 & $-0.09$ \\
\bottomrule
\end{tabular}

\vspace{4pt}
\begin{tabular}{lrr}
\toprule
Target & Mean $\Delta$ (calibrated$-$default) & Calibration wins \\
\midrule
20\%   & $-0.13$ & 1/4 \\
40\%   & $-0.05$ & 3/4 \\
60\%   & $+0.45$ & 3/4 \\
80\%   & $+0.75$ & 3/4 \\
\midrule
All 16 & $+0.25$ & 10/16 \\
\bottomrule
\end{tabular}

\label{tab:gate-init}
\end{table}

\subsection{PI versus PID Control}
\label{app:pid}

We isolate the derivative term by retraining one \dtm arm per task and training target with $k_d=0$ and the derivative disabled, changing nothing else (T5Gemma 2 1B--1B, L5, $c=-22$, $\tau=-7$), for five tasks at all four targets. Table~\ref{tab:pid} reports the best in-training validation score at the trained deletion rate, along with the controller's tracking error.

\begin{table}[htbp]
\centering
\caption{\textbf{PI versus PID deletion-rate control on T5Gemma 2 1B--1B at all four training targets.} Each \dtm arm is retrained with the derivative term disabled ($k_d=0$), changing nothing else (L5, $c=-22$, $\tau=-7$). Scores are best in-training validation scores at the trained deletion rate; tracking error $|e_t|$ and overshoot above the target are in percentage points of training-time deletion. The derivative term tightens tracking at every target (lower mean $|e_t|$ in 19 of 20 pairs, one tie) but improves accuracy only at the 80\% target, where PID wins on all five tasks by 0.81 points on average.}
\footnotesize
\setlength{\tabcolsep}{5pt}
\begin{tabular}{llrrrcc}
\toprule
Task & Target & PID & PI & $\Delta$ (PI$-$PID) & Mean $|e_t|$, PID $\to$ PI & Max overshoot, PID $\to$ PI \\
\midrule
DocVQA          & 20\% & 81.06 & 81.54 & $+0.48$ & 2.41 $\to$ 4.29  & \phantom{0}8.0 $\to$ 23.2 \\
ChartQA            & 20\% & 69.38 & 68.54 & $-0.83$ & 4.15 $\to$ 10.83 & 19.8 $\to$ 32.4 \\
VizWiz          & 20\% & 70.24 & 69.91 & $-0.33$ & 4.40 $\to$ 18.19 & 17.0 $\to$ 34.2 \\
TextVQA            & 20\% & 71.37 & 71.69 & $+0.32$ & 3.88 $\to$ \phantom{0}4.57 & 10.4 $\to$ 26.8 \\
GQA                & 20\% & 71.53 & 71.18 & $-0.35$ & 2.60 $\to$ \phantom{0}4.56 & 36.3 $\to$ 47.7 \\
\midrule
DocVQA          & 40\% & 78.25 & 78.59 & $+0.34$ & 3.91 $\to$ \phantom{0}7.48 & \phantom{0}9.0 $\to$ 20.2 \\
ChartQA            & 40\% & 65.89 & 67.92 & $+2.03$ & 2.76 $\to$ \phantom{0}4.07 & 12.9 $\to$ 21.8 \\
VizWiz          & 40\% & 69.94 & 69.65 & $-0.29$ & 6.79 $\to$ 13.96 & 26.6 $\to$ 37.1 \\
TextVQA            & 40\% & 70.96 & 71.15 & $+0.19$ & 3.73 $\to$ \phantom{0}3.70 & 25.5 $\to$ 36.4 \\
GQA                & 40\% & 71.05 & 71.31 & $+0.26$ & 2.48 $\to$ \phantom{0}4.44 & 33.5 $\to$ 39.7 \\
\midrule
DocVQA          & 60\% & 71.74 & 71.77 & $+0.02$ & 5.32 $\to$ \phantom{0}9.24 & 12.1 $\to$ 22.2 \\
ChartQA            & 60\% & 64.58 & 65.21 & $+0.62$ & 1.94 $\to$ \phantom{0}3.11 & 16.9 $\to$ 20.8 \\
VizWiz          & 60\% & 69.74 & 70.21 & $+0.47$ & 5.23 $\to$ \phantom{0}8.11 & 18.2 $\to$ 24.3 \\
TextVQA            & 60\% & 69.94 & 69.94 & $+0.00$ & 3.00 $\to$ \phantom{0}3.08 & 14.4 $\to$ 25.1 \\
GQA                & 60\% & 70.32 & 70.72 & $+0.40$ & 2.51 $\to$ \phantom{0}3.63 & 39.5 $\to$ 39.3 \\
\midrule
DocVQA          & 80\% & 57.90 & 55.51 & $-2.39$ & 5.78 $\to$ \phantom{0}9.17 & \phantom{0}5.7 $\to$ 14.1 \\
ChartQA            & 80\% & 63.07 & 62.86 & $-0.21$ & 1.07 $\to$ \phantom{0}2.25 & \phantom{0}4.4 $\to$ 15.7 \\
VizWiz          & 80\% & 69.63 & 68.88 & $-0.75$ & 4.43 $\to$ \phantom{0}6.51 & \phantom{0}8.1 $\to$ 11.9 \\
TextVQA            & 80\% & 67.89 & 67.72 & $-0.17$ & 1.57 $\to$ \phantom{0}2.70 & \phantom{0}5.1 $\to$ \phantom{0}9.9 \\
GQA                & 80\% & 69.98 & 69.46 & $-0.52$ & 1.57 $\to$ \phantom{0}2.86 & 19.5 $\to$ 19.9 \\
\bottomrule
\end{tabular}

\vspace{4pt}
\begin{tabular}{lrrrr}
\toprule
Target & Mean $\Delta$ (PI$-$PID) & PI wins & PID tighter, mean $|e_t|$ & PID tighter, overshoot \\
\midrule
20\%    & $-0.14$ & 2/5  & 5/5  & 5/5 \\
40\%    & $+0.51$ & 4/5  & 4/5  & 5/5 \\
60\%    & $+0.30$ & 5/5  & 5/5  & 4/5 \\
80\%    & $-0.81$ & 0/5  & 5/5  & 5/5 \\
\midrule
All 20  & $-0.03$ & 11/20 & 19/20 & 19/20 \\
\bottomrule
\end{tabular}

\label{tab:pid}
\end{table}

PID tracks its target more tightly at every target: it lowers the mean tracking error in 19 of 20 pairs (on the 20th, TextVQA at 40\%, the two differ by 0.03 points) and the maximum overshoot in 19; on GQA at 60\%, both controllers overshoot early by about 39 points, but PID recovers faster. The accuracy effect depends on the target. At 20--60\%, the two controllers are within noise (mean $\Delta$ of $-0.14$, $+0.51$, and $+0.30$), whereas at 80\%, PID wins on all five tasks (mean $-0.81$), with the largest gain on DocVQA ($2.39$ ANLS). Early overshoots of 20--48 points under PI aren't significant at 20--60\% targets, but at an 80\% target even an overshoot of 10--20 points trains the model at 90--100\% deletion (about 94\% on DocVQA), discarding information that it cannot recover within its training budget.

\subsection{Attention-Score Regularization}
\label{app:attn-reg}

Because soft deletion suppresses a key only by adding $g_j\geq c$ to its logits, the backbone could in principle learn to raise $\mathbf q_i^\top\mathbf k_j$ enough to cancel the bias. MrT5 counters this with an attention-score regularizer $\mathcal L_S$ \citep[App.~D]{kallini2025mrt5dynamictokenmerging}. On T5Gemma 2 1B--1B at 80\% deletion (L5), we measured this failure directly and tested two defenses: $\mathcal L_S$ in MrT5's form ($\beta=5$), and a logit soft-cap $v\tanh(s/v)$ with $v=20<|c|$, which makes cancellation arithmetically impossible. MrT5's knee $m=5$ lies below the mean attention score of this backbone (6.1), so we set $m=|c|=22$. Every arm is compared at an exactly enforced 79.7\% deletion rate (Table~\ref{tab:attn-reg}).

\begin{table}[htbp]
\centering
\caption{\textbf{Attention-score regularization for the delete gate.} The first column is the \dtm score without regularization (best checkpoint), and the next two are changes relative to it. ``Cap 20 without gate'' is the cap's effect on a fine-tuned model with no deletion. The soft$-$hard gap is the score difference between soft and hard deletion without regularization. Arms within a row share a training recipe. Neither defense improves accuracy on average, so we use neither.}
\label{tab:attn-reg}
\footnotesize
\setlength{\tabcolsep}{6pt}
\begin{tabular}{lrrrrr}
\toprule
Task & No regularizer & $+\mathcal L_S$ & $+$cap 20 & Cap 20 without gate & Soft$-$hard gap \\
\midrule
DocVQA & 51.69 & $+0.24$ & $+1.41$ & $-1.09$ & 8.03 \\
ChartQA & 64.27 & $-0.89$ & $-2.03$ & $-1.72$ & 0.26 \\
TextVQA & 67.24 & $+0.68$ & $-0.45$ & $-0.63$ & $-0.02$ \\
VQAv2   & 73.17 & $+0.04$ & $+0.54$ & $-0.37$ & 0.28 \\
\midrule
Mean $\Delta$ & --- & $+0.02$ & $-0.13$ & $-0.95$ & --- \\
\bottomrule
\end{tabular}

\end{table}

The failure mode exists: on every task, scores on suppressed keys reach 38--44 logits against a bias of $-22$, and on DocVQA 3\% of suppressed keys end up with a positive post-bias score. However, it matters only where soft deletion already overstates hard deletion, which is substantial only on DocVQA (8.0 points). Both defenses work mechanically: the cap bounds suppressed-key scores at 19.3 and reduces leaked attention 3--9$\times$, and a cancellation-targeted variant of $\mathcal L_S$ lowers the largest post-bias score on suppressed keys monotonically in $\beta$ (from 25.3 to 8.1). Neither improves accuracy. MrT5's form of $\mathcal L_S$ changes the mean score by only $+0.02$, and the cancellation variant lowers DocVQA ANLS by 1.3, 2.1, and 4.9 points at $\beta=1,5,25$. The cap helps only on DocVQA and costs accuracy on the ungated model on all four tasks. We therefore use neither defense.

\subsection{Natural Thresholds and Per-Example Budgets}
\label{app:adaptive-budgets}

\begin{table}[htbp]
\centering
\caption{\textbf{Natural thresholding versus a fixed per-example budget within the same checkpoint (T5Gemma 2 1B--1B, L5).} Each row uses the main-experiment \dtm checkpoint for that task and training target. Natural deletion removes every visual token with $g_j\leq\tau$, so the per-example rate varies (mean and 5th--95th percentiles shown); forced deletion removes the same mean fraction from every example. The two passes' FLOPs differ by at most 0.61\%. Metrics follow the main results, except that RefCOCO+ uses mean IoU. Natural thresholding scores higher in 16 of 18 pairs, including all nine at the 60\% target, by 0.29 points on average.}
\footnotesize
\setlength{\tabcolsep}{4pt}
\begin{tabular}{llrrrrc}
\toprule
Task & Target & Nat.\ del.\ (\%) & Nat.\ $p_5$--$p_{95}$ (\%) & Natural & Forced & $\Delta$ (Nat.$-$Forced) \\
\midrule
DocVQA   & 60\% & 59.98 & 40.2--82.4 & 73.22 & 71.95 & $+1.26$ \\
DocVQA   & 80\% & 80.43 & 68.0--91.8 & 56.87 & 57.04 & $-0.17$ \\
ChartQA  & 60\% & 61.69 & 43.8--79.7 & 65.10 & 65.05 & $+0.05$ \\
ChartQA  & 80\% & 80.47 & 67.6--90.6 & 63.18 & 62.40 & $+0.78$ \\
TextVQA  & 60\% & 63.93 & 31.2--88.3 & 70.10 & 69.79 & $+0.31$ \\
TextVQA  & 80\% & 77.64 & 53.5--94.5 & 68.60 & 67.92 & $+0.68$ \\
VizWiz   & 60\% & 59.84 & 27.7--91.8 & 70.11 & 69.96 & $+0.15$ \\
VizWiz   & 80\% & 80.32 & 58.2--97.7 & 69.64 & 69.73 & $-0.09$ \\
GQA      & 60\% & 60.79 & 43.8--78.1 & 70.69 & 70.58 & $+0.11$ \\
GQA      & 80\% & 79.91 & 71.5--87.5 & 70.09 & 69.85 & $+0.23$ \\
TallyQA  & 60\% & 67.10 & 48.8--82.8 & 73.38 & 73.17 & $+0.21$ \\
TallyQA  & 80\% & 83.31 & 70.7--93.4 & 72.27 & 71.82 & $+0.45$ \\
RefCOCO+ & 60\% & 60.47 & 46.9--75.0 & 70.39 & 70.06 & $+0.33$ \\
RefCOCO+ & 80\% & 79.46 & 71.5--87.1 & 65.55 & 65.48 & $+0.08$ \\
VQAv2    & 60\% & 59.93 & 44.9--75.0 & 77.26 & 77.00 & $+0.26$ \\
VQAv2    & 80\% & 79.40 & 70.3--87.5 & 76.31 & 76.07 & $+0.24$ \\
POPE     & 60\% & 57.50 & 42.6--72.7 & 85.19 & 85.11 & $+0.08$ \\
POPE     & 80\% & 79.80 & 71.5--87.1 & 84.56 & 84.31 & $+0.24$ \\
\bottomrule
\end{tabular}

\label{tab:natural-threshold}
\end{table}

Our evaluations enforce an exact budget for every example, which separates the budget from the gate's natural threshold-crossing rate (Section~\ref{sec:experiments}). Table~\ref{tab:natural-threshold} compares two inference policies within the same main-experiment checkpoint in 18 T5Gemma 2 1B--1B task/target pairs: natural deletion of every visual token with $g_j\leq\tau$, and exact-count deletion of that policy's measured mean rate from every example. Natural deletion scores higher in 16 of 18 comparisons, including all nine at the 60\% target, by 0.29 points on average, with the largest gain on DocVQA at 60\% (+1.26 ANLS); the two exceptions are DocVQA and VizWiz at 80\%. The natural policy allocates budgets unevenly: on VizWiz at the 60\% target, per-example deletion ranges from 27.7\% to 91.8\% between the 5th and 95th percentiles. Natural thresholding therefore has a small measured advantage over a fixed per-example budget, but the comparison also changes the remaining tokens' weighting (naturally kept tokens receive no answer-query bias; Section~\ref{sec:dtm-improvements}), so it does not isolate the effect of token allocation. We report fixed budgets in the main results so that every method removes the same number of tokens from every example.

\FloatBarrier
\section{Model Architectures}
\label{app:model-architectures}

\subsection{Model Sizes and Configurations}

We use T5Gemma 2 1B--1B, T5Gemma 2 270M--270M, and Gemma 3 4B. Table~\ref{tab:model-configurations} lists the configurations of the backbones used in our implementation. The T5Gemma 2 encoder and decoder are initialized from the pretrained Gemma 3 1B and 270M models, and subsequently adapted with UL2 \citep{zhang2025t5gemma2seeingreading}. T5Gemma 2 shares its encoder input, decoder input, and vocabulary-projection weights. Gemma 3 also ties its input embeddings and vocabulary projection.

\begin{table}[htbp]
\centering
\caption{\textbf{Configurations of the three model backbones.}
T5Gemma 2 dimensions and head counts apply to both language stacks.
Language parameter counts include the shared text embeddings once; totals additionally include the vision encoder and image projector. All three backbones receive the same 256 visual tokens per image, so a deletion rate removes the same number of visual tokens in each.}
\label{tab:model-configurations}
\footnotesize
\setlength{\tabcolsep}{5pt}
\begin{tabular}{lrrr}
\toprule
 & \multicolumn{2}{c}{T5Gemma 2} & Gemma 3 \\
\cmidrule(lr){2-3}
Configuration & 1B--1B & 270M--270M & 4B \\
\midrule
Encoder / decoder blocks & 26 / 26 & 18 / 18 & --- / 34 \\
Hidden width $d$ & 1,152 & 640 & 2,560 \\
Feed-forward width $d_{\mathrm{ff}}$ & 6,912 & 2,048 & 10,240 \\
Query heads & 4 & 4 & 8 \\
Key/value heads & 1 & 1 & 4 \\
Head width $d_h$ & 256 & 256 & 256 \\
Vocabulary size & 262,144 & 262,144 & 262,208 \\
Local-attention window & 512 & 512 & 1,024 \\
Visual tokens per image & 256 & 256 & 256 \\
\midrule
Language parameters (billions) & 1.698 & 0.368 & 3.880 \\
Total parameters (billions) & 2.116 & 0.786 & 4.300 \\
\bottomrule
\end{tabular}

\end{table}

\begin{figure}[tbp]
\centering
\includegraphics[width=0.8\linewidth]{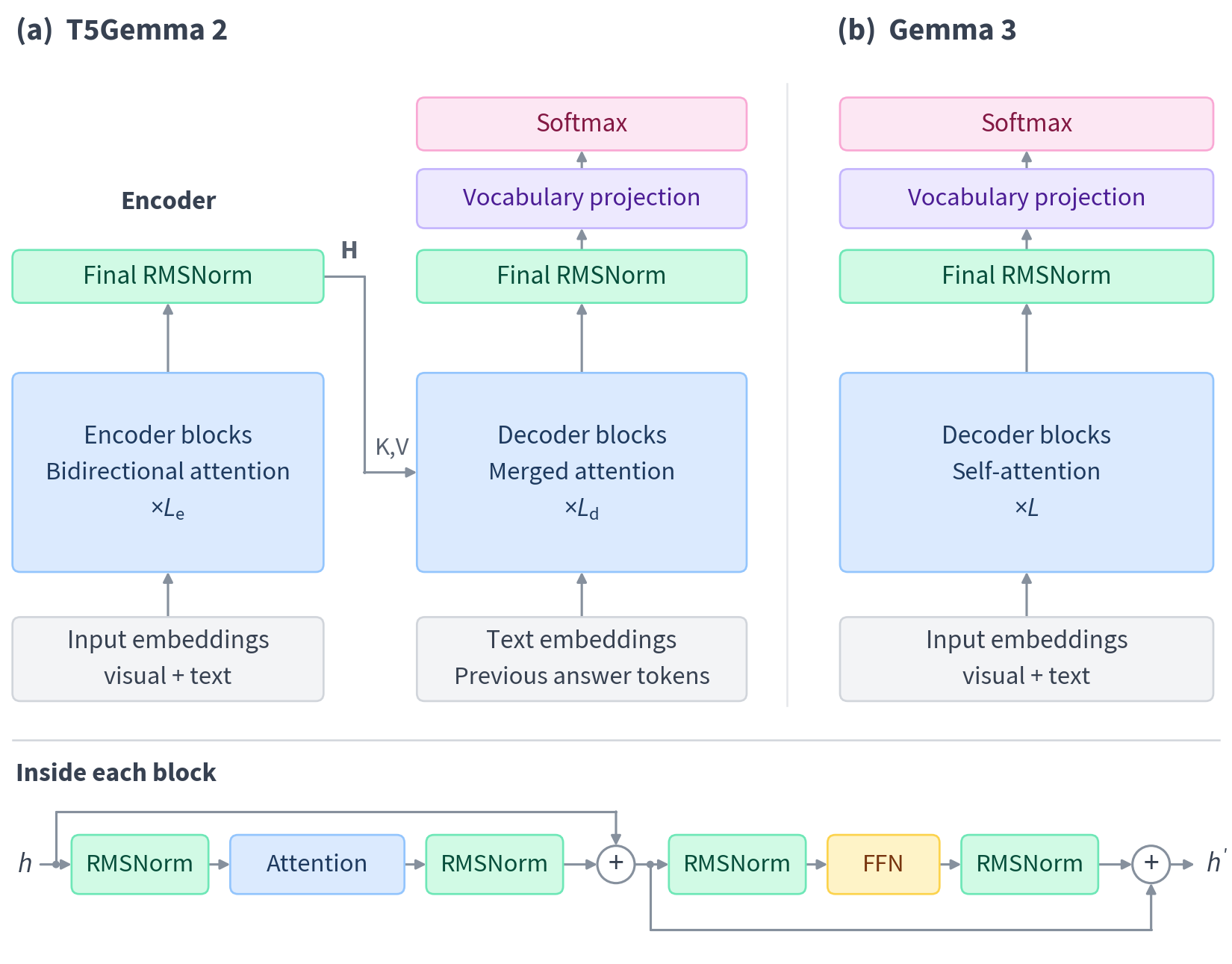}
\caption{\textbf{T5Gemma 2 and Gemma 3 language-backbone architectures.}
Visual tokens enter after SigLIP encoding, pooling, and image projection.
(a) T5Gemma 2 uses a bidirectional encoder and a decoder with merged
self- and cross-attention. Each decoder block projects the final normalized
encoder states $\mathbf H$ into its own keys and values.
(b) Gemma 3 uses a decoder-only backbone.
Here $L_{\mathrm{e}}$ and $L_{\mathrm{d}}$ denote encoder and decoder depths:
both are 26 for T5Gemma 2 1B--1B and 18 for 270M--270M; Gemma 3 4B has
$L=34$ decoder blocks. }
\label{fig:backbone-architectures}
\end{figure}

All three models use grouped-query attention, query/key normalization, rotary positional embeddings, and gated GELU feed-forward sublayers. Each attention and feed-forward sublayer applies RMSNorm \citep{zhang2019rootmeansquarelayer} before the operation and again before residual addition. For a sublayer function $f$, this order is
\begin{equation}
\mathbf y=\mathbf x+
\operatorname{RMSNorm}_{\mathrm{post}}\!\left(
f\!\left(\operatorname{RMSNorm}_{\mathrm{pre}}(\mathbf x)\right)
\right).
\end{equation}
The pre- and post-normalizations have separate parameters, and a final RMSNorm follows each complete language stack. The single key/value head in each T5Gemma 2 model is the multi-query special case of grouped-query attention. Figure~\ref{fig:backbone-architectures} illustrates both backbone architectures and the normalization and residual connections within each block.

\subsection{Vision Processing}

All three models use the same approximately 417-million-parameter SigLIP \citep{zhai2023sigmoidlosslanguageimage} vision encoder. Each $896\times896$ image is divided into $14\times14$ patches, producing a $64\times64$ grid of 4,096 patch representations.  Spatial average pooling reduces the grid to $16\times16$, yielding 256 visual tokens per image. These representations are passed through a linear projection into the language model's hidden width. Visual tokens are inserted in the encoder input for T5Gemma 2 and in the decoder prompt for Gemma 3.

\subsection{Attention Visibility and Layer Indexing}

As in Section~\ref{sec:background-models}, layer 0 denotes the input states and transformer layers are indexed from one. The global-attention layers are $6,12,18,24$ in each T5Gemma 2 1B--1B stack, $6,12,18$ in each 270M--270M stack, and $6,12,18,24,30$ in Gemma 3 4B. All remaining layers use local attention, including the final layers when the stack ends before the next global layer.

T5Gemma 2's encoder uses a 512-token bidirectional local window. In our implementation, this includes up to 255 preceding positions, the current position, and 256 following positions. Its decoder uses a causal 512-token window over decoder keys in local layers; all valid encoder keys remain visible through merged attention in every decoder layer. Gemma 3 uses a causal 1,024-token local window for text queries, including the current position, and permits bidirectional attention within each image block. Thus, image-prefix states cannot attend to a later text question, although image patches can attend to one another.

As in Figure~\ref{fig:vmrt5gemma2}, $k$ denotes the first layer that processes the shortened sequence, so layers $1,\ldots,k-1$ process the full sequence. We name each configuration by the number of full-sequence layers: our main setting, L5, deletes after five layers and before the sixth ($k=6$). Attention-based selectors use scores from layer $k-1$; consequently, L5 reads the local-attention layer 5, whereas L6 reads the first global-attention layer, layer 6.

\paragraph{Merged attention.}
Let $\mathbf H\in\mathbb R^{n\times d}$ and $\mathbf X\in\mathbb R^{m\times d}$ contain the final encoder states and the decoder states of T5Gemma 2, where $n,m$ are their sequence lengths. For one query head and its associated key/value head, the learned projections $\mathbf W_q,\mathbf W_k,\mathbf W_v\in\mathbb R^{d\times d_h}$ and $\mathbf W_o\in\mathbb R^{d_h\times d}$ give
\begin{equation}
\mathbf Q=\mathbf X\mathbf W_q,\quad 
\mathbf K=[\mathbf X;\mathbf H]\mathbf W_k,\quad
\mathbf V=[\mathbf X;\mathbf H]\mathbf W_v,\quad
\mathbf O=\mathbf A\mathbf V\mathbf W_o,
\label{eq:merged-projections}
\end{equation}
so $\mathbf Q\in\mathbb R^{m\times d_h}$, $\mathbf K,\mathbf V\in\mathbb R^{(m+n)\times d_h}$, and the attention weights $\mathbf A$ have shape $m\times(m+n)$. We omit head indices, query/key normalization, and positional transforms. Decoder queries jointly attend to visible decoder keys and all valid encoder keys, with causality applied to the decoder-key block. Under \dtm, hard deletion removes rows of $\mathbf H$, shortening the encoder block of $\mathbf K$ and $\mathbf V$ in every decoder layer.

\FloatBarrier

\section{Experimental Details}
\label{app:experimental-details}

Table~\ref{tab:task-details} lists each benchmark's metric, the number of evaluated examples, and the training steps; Table~\ref{tab:hyperparameters} lists the shared adaptation and gate recipe. All evaluations use zero-shot prompts. To keep splits consistent, we evaluate on validation splits wherever one exists: DocVQA, InfoVQA, TextVQA, and VQAv2 withhold their test answers for server-side scoring, and the evidence annotations for TextVQA, VizWiz, and GQA are released only for training and validation images. TallyQA and POPE have no validation split, so we use their test splits. DocVQA and InfoVQA use average normalized Levenshtein similarity (ANLS), ChartQA uses relaxed accuracy, TextVQA, VQAv2, and VizWiz use their VQA answer-scoring rules, and TallyQA uses exact-count accuracy; RefCOCO+ counts a predicted box as correct when its intersection-over-union with the reference is at least 0.5.

\begin{table}[htbp]
\centering
\caption{\textbf{Evaluation populations and training steps (T5Gemma 2 1B--1B).} Every method and deletion rate is evaluated on the same examples. VQAv2 and GQA use fixed, uniformly random subsets of their validation splits (25,600 of 214,354 and 12,800 of 132,062 questions); every other benchmark uses its full split. POPE evaluates the VQAv2-trained checkpoints. Budgets differ across tasks, so we compare methods within each task rather than at equal training compute. The last column gives the source of the evidence regions in Figures~\ref{fig:evidence-retained} and~\ref{fig:evidence-other-rates}, which are computed on the evaluated examples that have at least one evidence region (``With evidence''); OCR-word regions are the words that spell a gold answer.}
\label{tab:task-details}
\footnotesize
\setlength{\tabcolsep}{3pt}
\resizebox{\linewidth}{!}{%
\begin{tabular}{@{}llrrrl@{}}
\toprule
Task & Split / metric & \begin{tabular}[b]{@{}r@{}}Evaluated\\examples\end{tabular} & \begin{tabular}[b]{@{}r@{}}With\\evidence\end{tabular} & \begin{tabular}[b]{@{}r@{}}Training steps\\(deletion methods)\end{tabular} & \begin{tabular}[b]{@{}l@{}}Evidence\\regions\end{tabular} \\
\midrule
DocVQA & val. / ANLS & 5,349 & 5,109 & 300 & OCR words \citep{tito2023hierarchicalmultimodaltransformersmultipage} \\
ChartQA & val. / relaxed acc. & 1,920 & 1,142 & 1{,}250 & Support boxes \citep{vogel2025refchartqagroundingvisualanswer} \\
InfoVQA & val. / ANLS & 2,801 & 1,916 & 300 & OCR words \citep{mathew2021infographicvqa} \\
TextVQA & val. / VQA acc. & 5,000 & 4,239 & 1{,}250 & OCR words \citep{singh2021textocrlargescaleendtoendreasoning} \\
VQAv2 & val. / VQA acc. & 25,600 & --- & 1{,}500 & --- \\
VizWiz & val. / VQA acc. & 4,319 & 1,131 & 160 & Grounding polygons \citep{chen2022groundinganswersvisualquestions} \\
GQA & val. / accuracy & 12,800 & 12,005 & 2{,}000 & Scene-graph boxes \citep{hudson2019gqanewdatasetrealworld} \\
TallyQA & test / count acc. & 38,589 & --- & 1{,}250 & --- \\
POPE & test / accuracy & 9,000 & 4,016 & VQAv2 & Object boxes \citep{lin2015microsoftcococommonobjects} \\
RefCOCO+ & val. / Acc@IoU$\geq$0.5 & 10,758 & 10,758 & 1{,}500 & Referent box \citep{yu2016modelingcontextreferringexpressions} \\
\bottomrule
\end{tabular}%
}
\end{table}

\begin{table}[htbp]
\centering
\caption{\textbf{Shared adaptation and gate recipe.} Apart from batch size, every listed setting is identical across the three backbones and all plotted arms.}
\label{tab:hyperparameters}
\small
\setlength{\tabcolsep}{4pt}
\renewcommand{\arraystretch}{1.12}
\begin{tabular}{@{}p{1.5in}p{3.85in}@{}}
\toprule
Setting & Value \\
\midrule
Adapter learning rate & $10^{-4}$ \\
Gate learning rate & $10^{-4}$ (separate parameter group) \\
Learning-rate schedule & Cosine for adapters and gate; no warmup \\
Weight decay & $10^{-4}$\\
Batch size & 256 (Gemma 3 4B: 128) \\
Gradient clipping & Maximum gradient norm 1 \\
LoRA rank / scaling / dropout & 32 / 64 / 0.05 \\
LoRA targets & Attention Q/K/V/O and all MLP projections \\
Vision encoder & Frozen (no LoRA) \\
Precision & BF16 backbone \\
Gate scale / deletion threshold & $c=-22$ / $\tau=-7$ \\
Controller P / I / D gains & $k_p=1$ / $k_i=10^{-5}$ / $k_d=0.5$ \\
Derivative lookback $\omega$ & 10 controller updates \\
Gate training mode & Soft deletion throughout training \\
Inference budget & Exactly $\lfloor\delta n_v\rfloor$ visual tokens deleted per example \\
\bottomrule
\end{tabular}
\end{table}

\FloatBarrier
\section{Additional Results and Analysis}
\label{app:additional-results}
\suppressfloats[t]%

\subsection{Matched-Rate Results on All Three Models}
\label{app:matched-all}
\label{app:gemma3-results}

\begin{figure}[tbp]
\centering
\includegraphics[width=\linewidth]{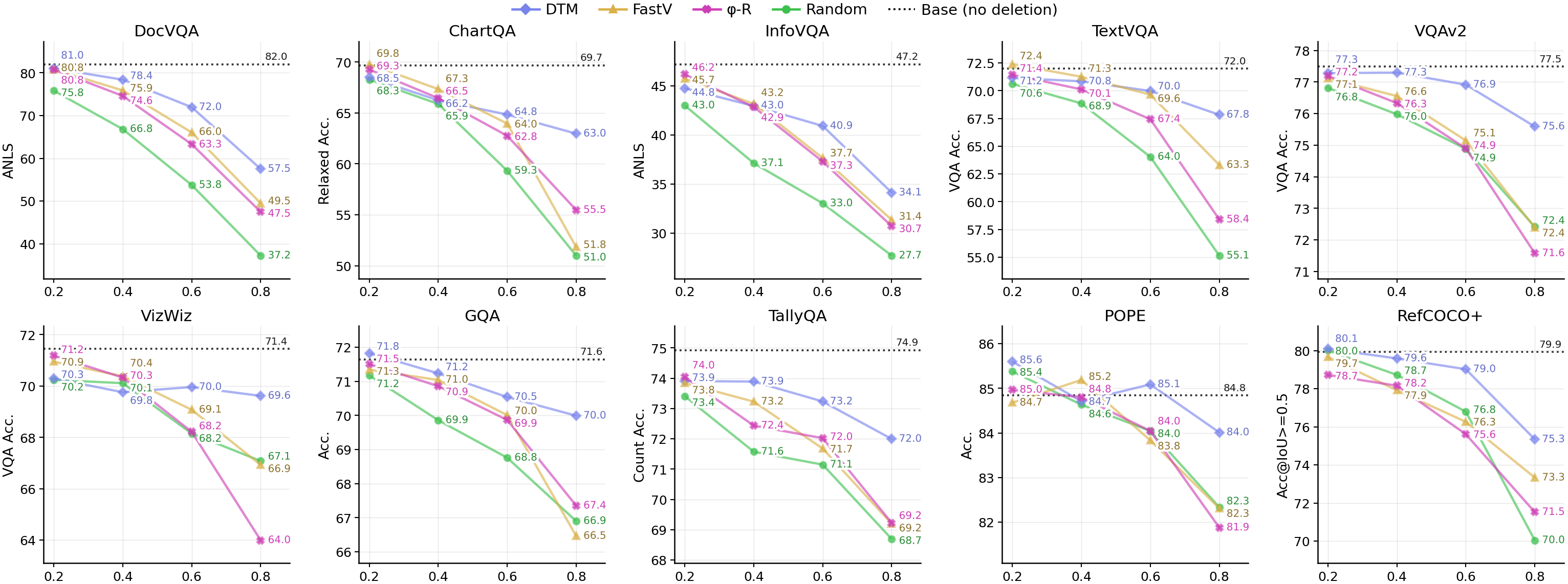}
\caption{\textbf{Matched-rate evaluation on T5Gemma 2 1B--1B (\vmrtfivegemmatwo), deletion after layer 5.} Each point uses a model trained and evaluated at the deletion rate $\delta$ on the horizontal axis. \dtm leads on all ten benchmarks at 60\% and 80\% deletion; at 20--40\%, the methods are close.}
\label{fig:t5-matched}
\end{figure}

\begin{figure}[tbp]
\centering
\includegraphics[width=\linewidth]{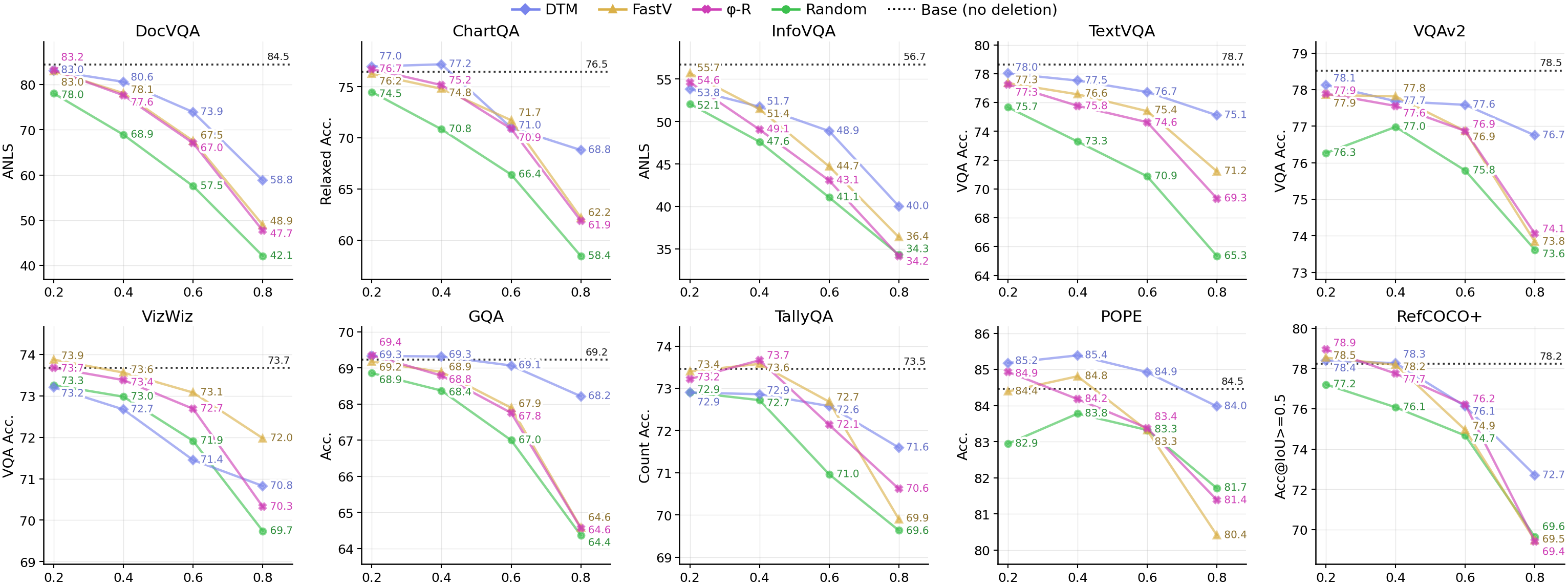}
\caption{\textbf{Matched-rate evaluation on Gemma 3 4B (\vmrgemmathree), deletion after five decoder layers during prompt prefill.} Each point uses a model trained and evaluated at the deletion rate $\delta$ on the horizontal axis. \dtm leads on nine benchmarks at 80\% and six at 60\%; at 20--40\%, the methods are close.}
\label{fig:gemma-matched}
\end{figure}

\begin{figure}[tbp]
\centering
\includegraphics[width=\linewidth]{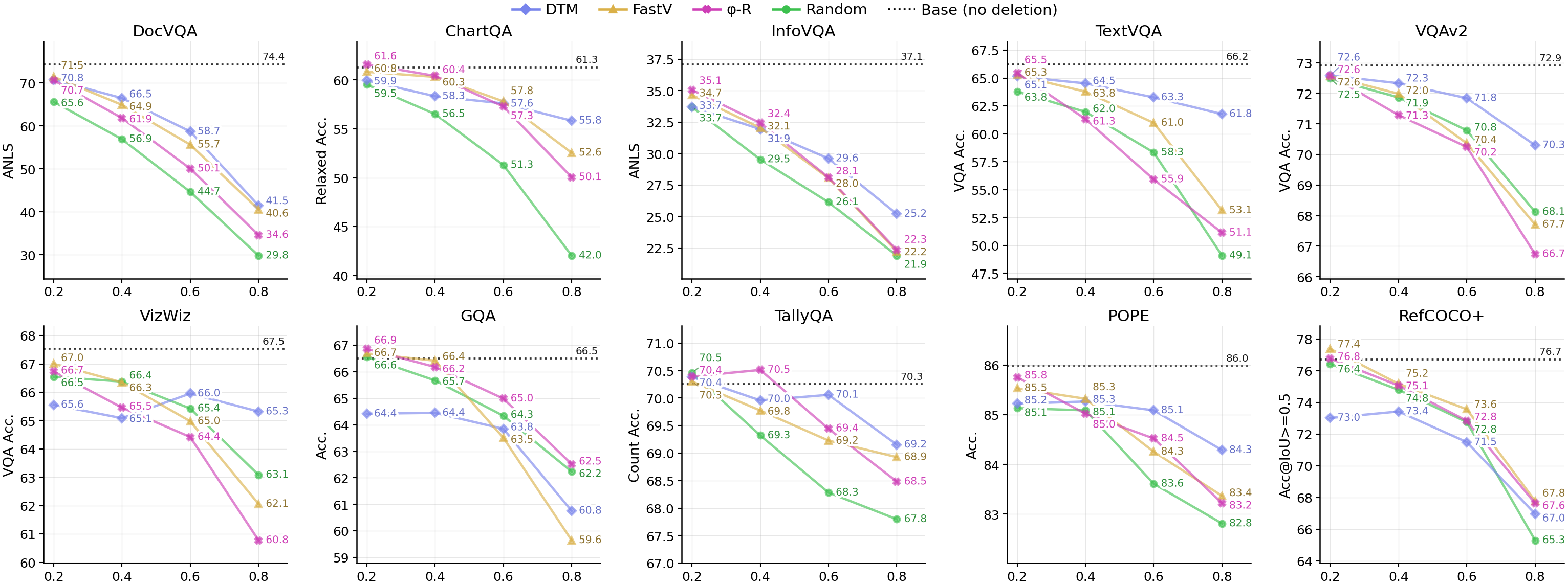}
\caption{\textbf{Matched-rate evaluation on T5Gemma 2 270M--270M, deletion after layer 5.} Each point uses a model trained and evaluated at the deletion rate $\delta$ on the horizontal axis. The advantage emerges at higher deletion rates on the smallest backbone: \dtm leads on 1, 3, 7, and 8 benchmarks at 20\%, 40\%, 60\%, and 80\%.}
\label{fig:270m-matched}
\label{fig:270m270m-matched-results}
\end{figure}

Figures~\ref{fig:t5-matched}, \ref{fig:gemma-matched}, and~\ref{fig:270m-matched} show every matched-rate result behind Table~\ref{tab:matched}, from 20\% to 80\% deletion, for T5Gemma 2 1B--1B, Gemma 3 4B, and T5Gemma 2 270M--270M. The three backbones share one pattern. At 20\% deletion, \dtm leads on 5, 4, and 1 of the ten benchmarks, respectively, and stays within 1.9 points of the strongest baseline on the two larger models; at 80\%, it leads on 10, 9, and 8, with mean margins of 3.8, 3.6, and 1.9 points. Section~\ref{sec:matched-results} discusses the two primary backbones, and Appendix~\ref{app:small-model} the smaller one.

\subsection{Fixed Checkpoints at Every Training Rate}
\label{app:fixed-all}

\begin{figure}[p]
\centering
\includegraphics[width=\linewidth]{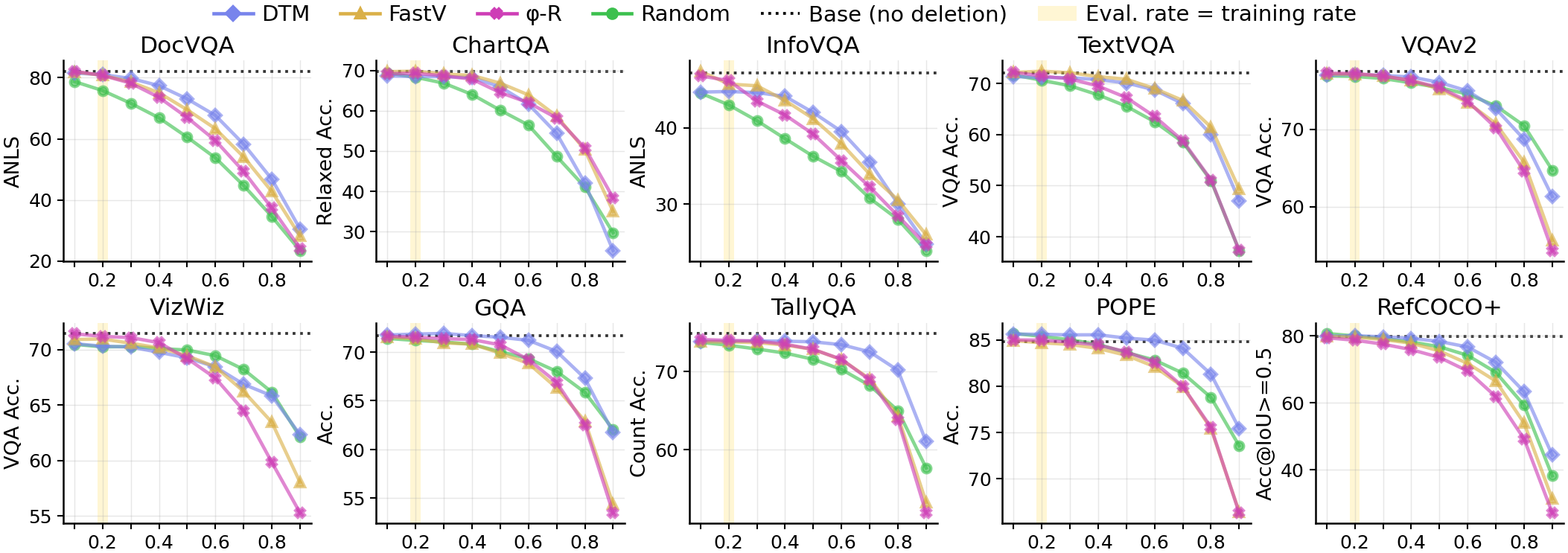}\\[-1pt]
{\small (a) Trained at 20\%}\\[6pt]
\includegraphics[width=\linewidth]{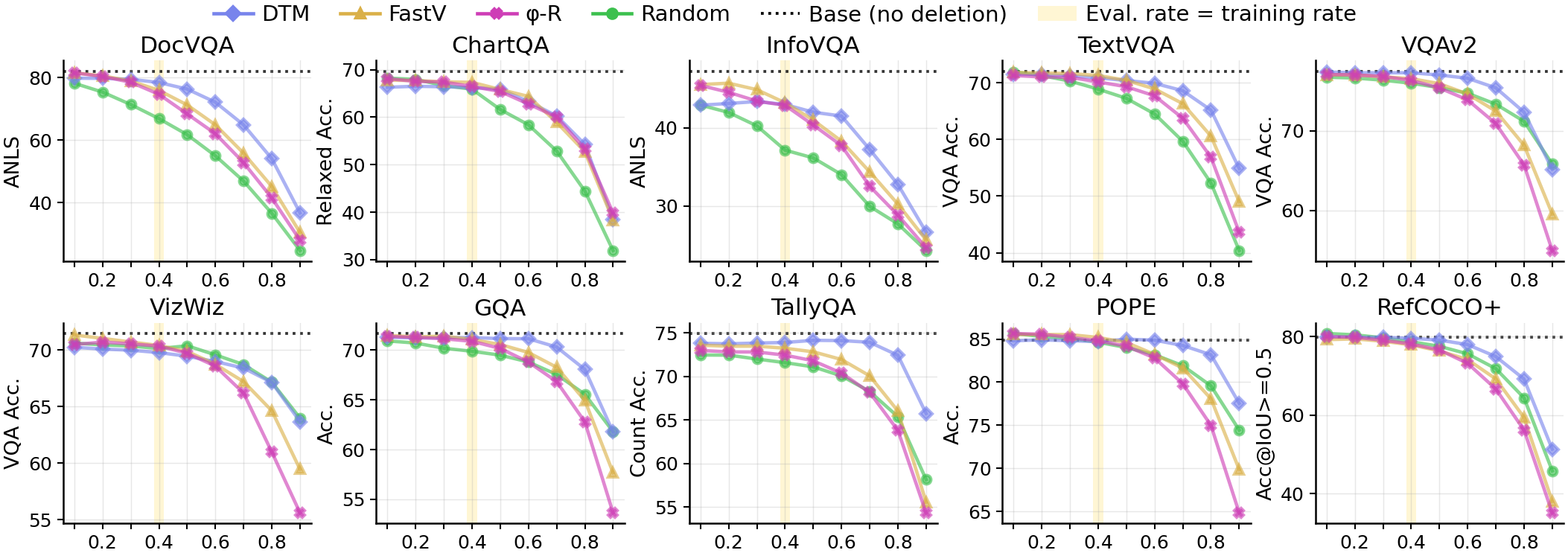}\\[-1pt]
{\small (b) Trained at 40\%}
\caption{\textbf{Fixed checkpoints on T5Gemma 2 1B--1B trained at 20\% (a) and 40\% (b) deletion.} Each curve evaluates one checkpoint at deletion rates from 10\% to 90\%; the training rate is highlighted. Blue diamonds: \dtm; gold triangles: \fastv; magenta crosses: \phir; green circles: \Random; dotted line: unpruned Base, fine-tuned the same way. Trained at 40\%, \dtm leads on nine benchmarks at 80\% inference deletion; checkpoints trained at only 20\% transfer less well, and \dtm leads on half.}
\label{fig:fixed-t5-2040}
\end{figure}

\begin{figure}[p]
\centering
\includegraphics[width=\linewidth]{figures/results/v5/train60_1b1b.png}\\[-1pt]
{\small (a) Trained at 60\%}\\[6pt]
\includegraphics[width=\linewidth]{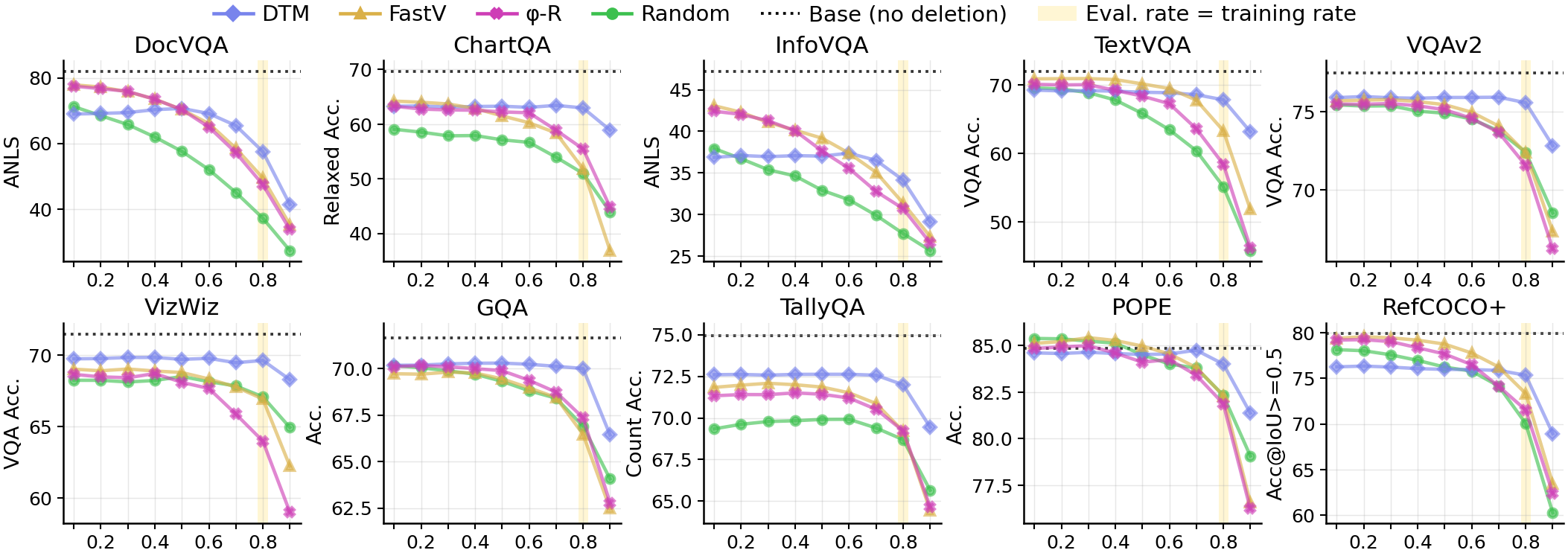}\\[-1pt]
{\small (b) Trained at 80\%}
\caption{\textbf{Fixed checkpoints on T5Gemma 2 1B--1B trained at 60\% (a) and 80\% (b) deletion.} Each curve evaluates one checkpoint at deletion rates from 10\% to 90\%; the training rate is highlighted. Blue diamonds: \dtm; gold triangles: \fastv; magenta crosses: \phir; green circles: \Random; dotted line: unpruned Base, fine-tuned the same way. Trained at 60\% or 80\%, \dtm leads on all ten benchmarks at 80\% and 90\% inference deletion.}
\label{fig:fixed-t5-6080}
\end{figure}

\begin{figure}[p]
\centering
\includegraphics[width=\linewidth]{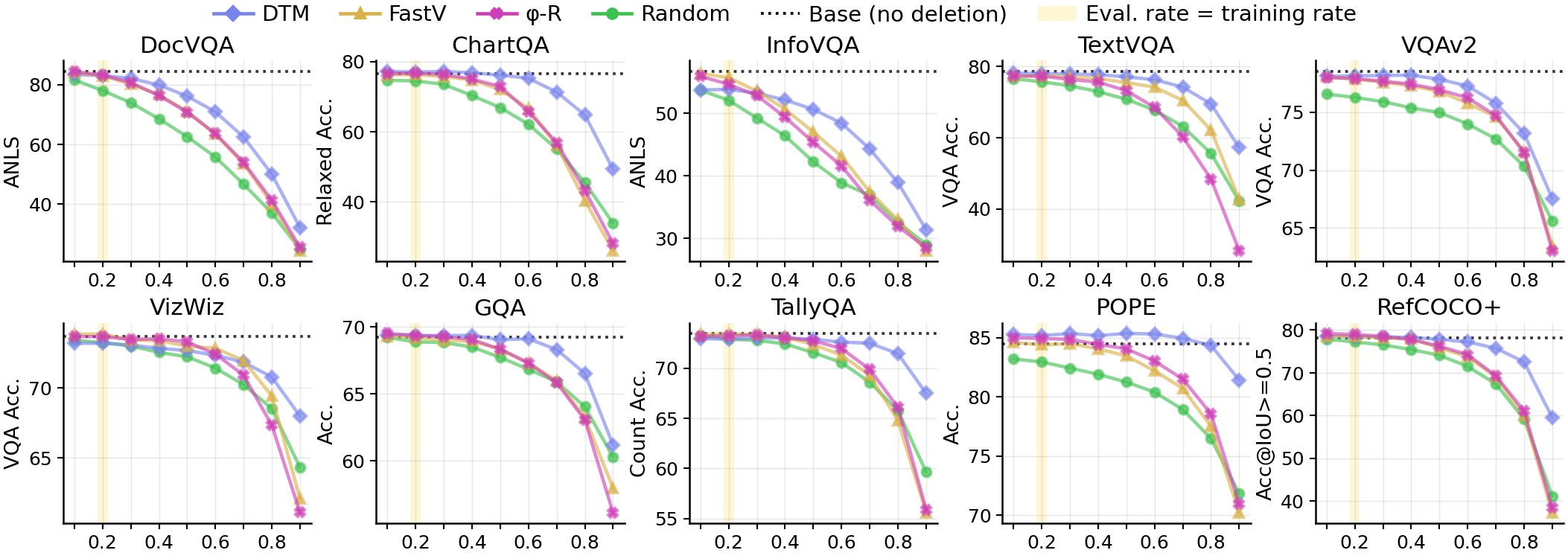}\\[-1pt]
{\small (a) Trained at 20\%}\\[6pt]
\includegraphics[width=\linewidth]{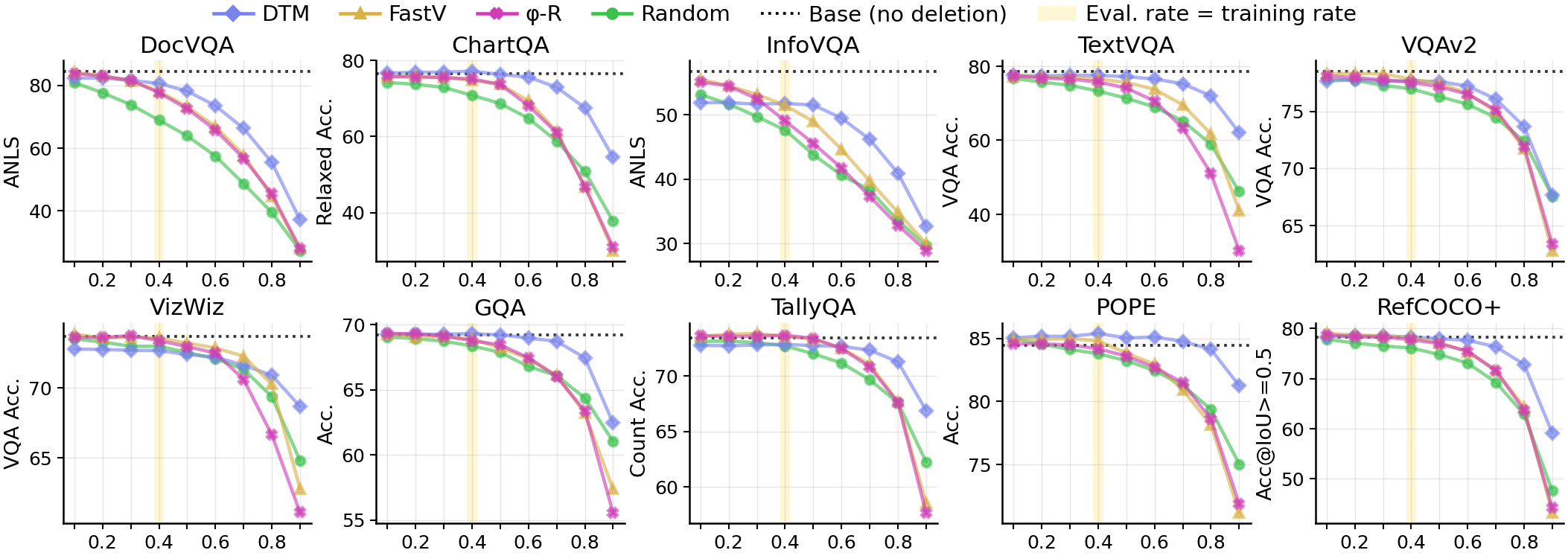}\\[-1pt]
{\small (b) Trained at 40\%}
\caption{\textbf{Fixed checkpoints on Gemma 3 4B trained at 20\% (a) and 40\% (b) deletion, deletion during prefill.} Each curve evaluates one checkpoint at deletion rates from 10\% to 90\%; the training rate is highlighted. Blue diamonds: \dtm; gold triangles: \fastv; magenta crosses: \phir; green circles: \Random; dotted line: unpruned Base, fine-tuned the same way. Even \dtm checkpoints trained at 20\% or 40\% lead on all ten benchmarks at 80\% and 90\% inference deletion.}
\label{fig:fixed-gemma-2040}
\end{figure}

\begin{figure}[p]
\centering
\includegraphics[width=\linewidth]{figures/results/v5/train60_gemma3.png}\\[-1pt]
{\small (a) Trained at 60\%}\\[6pt]
\includegraphics[width=\linewidth]{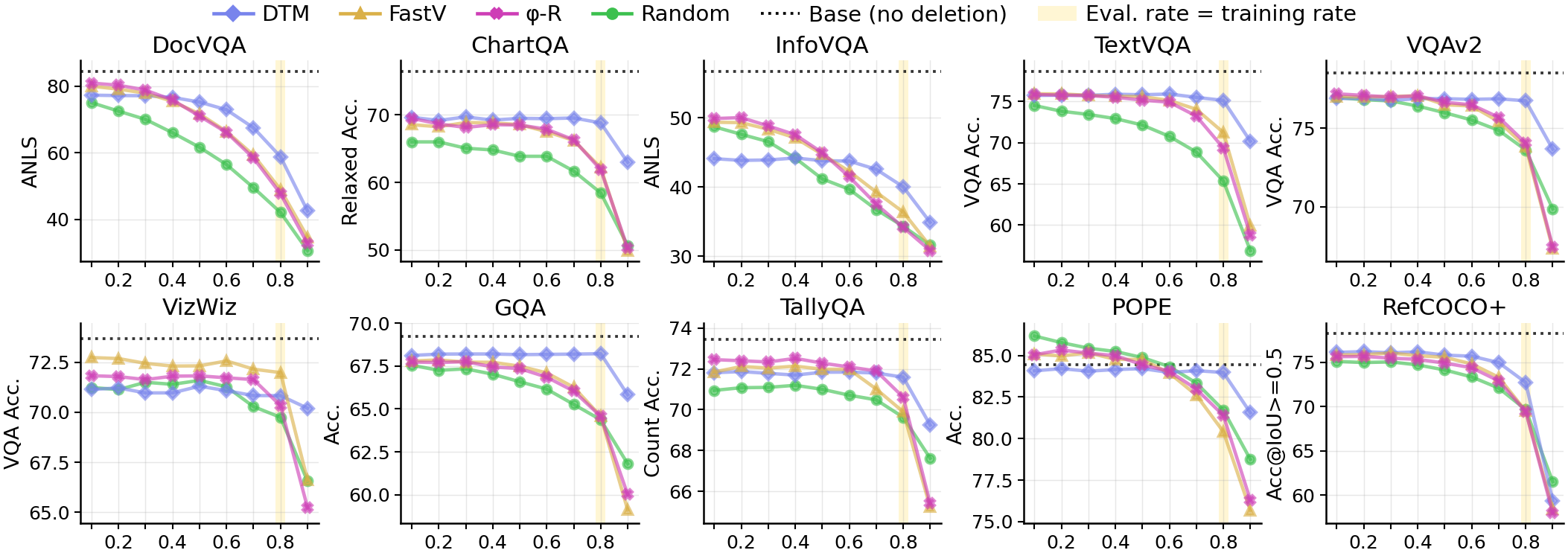}\\[-1pt]
{\small (b) Trained at 80\%}
\caption{\textbf{Fixed checkpoints on Gemma 3 4B trained at 60\% (a) and 80\% (b) deletion, deletion during prefill.} Each curve evaluates one checkpoint at deletion rates from 10\% to 90\%; the training rate is highlighted. Blue diamonds: \dtm; gold triangles: \fastv; magenta crosses: \phir; green circles: \Random; dotted line: unpruned Base, fine-tuned the same way. Trained at 60\%, \dtm leads on all ten benchmarks at 80\% and 90\% inference deletion; trained at 80\%, on nine.}
\label{fig:fixed-gemma-6080}
\end{figure}

\begin{figure}[p]
\centering
\includegraphics[width=\linewidth]{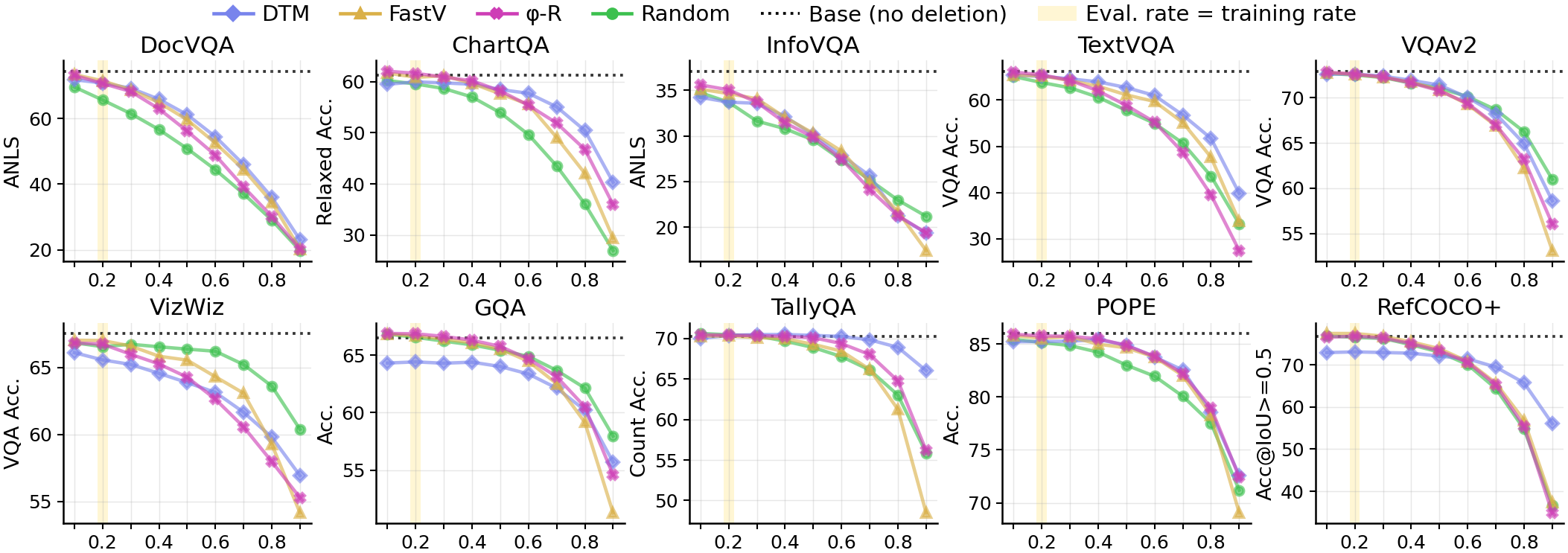}\\[-1pt]
{\small (a) Trained at 20\%}\\[6pt]
\includegraphics[width=\linewidth]{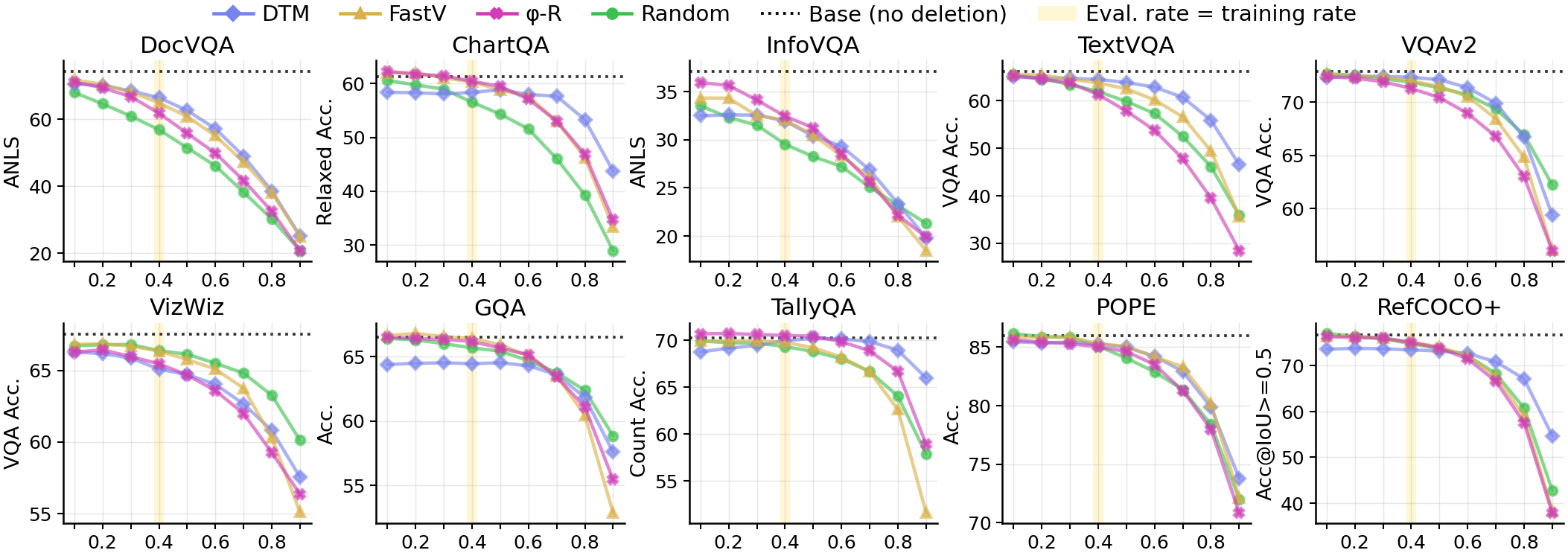}\\[-1pt]
{\small (b) Trained at 40\%}
\caption{\textbf{Fixed checkpoints on T5Gemma 2 270M--270M trained at 20\% (a) and 40\% (b) deletion.} Each curve evaluates one checkpoint at deletion rates from 10\% to 90\%; the training rate is highlighted. Blue diamonds: \dtm; gold triangles: \fastv; magenta crosses: \phir; green circles: \Random; dotted line: unpruned Base, fine-tuned the same way. The smallest backbone transfers less well from low training targets: \dtm leads on five or six benchmarks at 80\% and 90\% inference deletion.}
\label{fig:fixed-270m-2040}
\end{figure}

\begin{figure}[p]
\centering
\includegraphics[width=\linewidth]{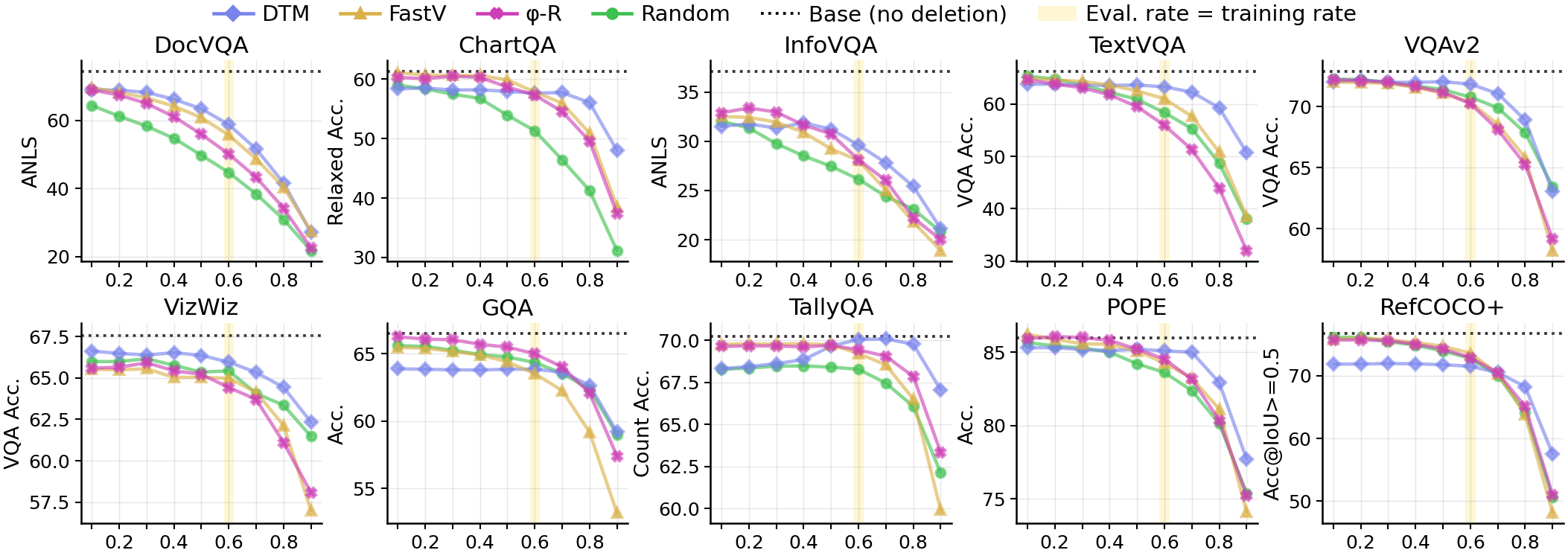}\\[-1pt]
{\small (a) Trained at 60\%}\\[6pt]
\includegraphics[width=\linewidth]{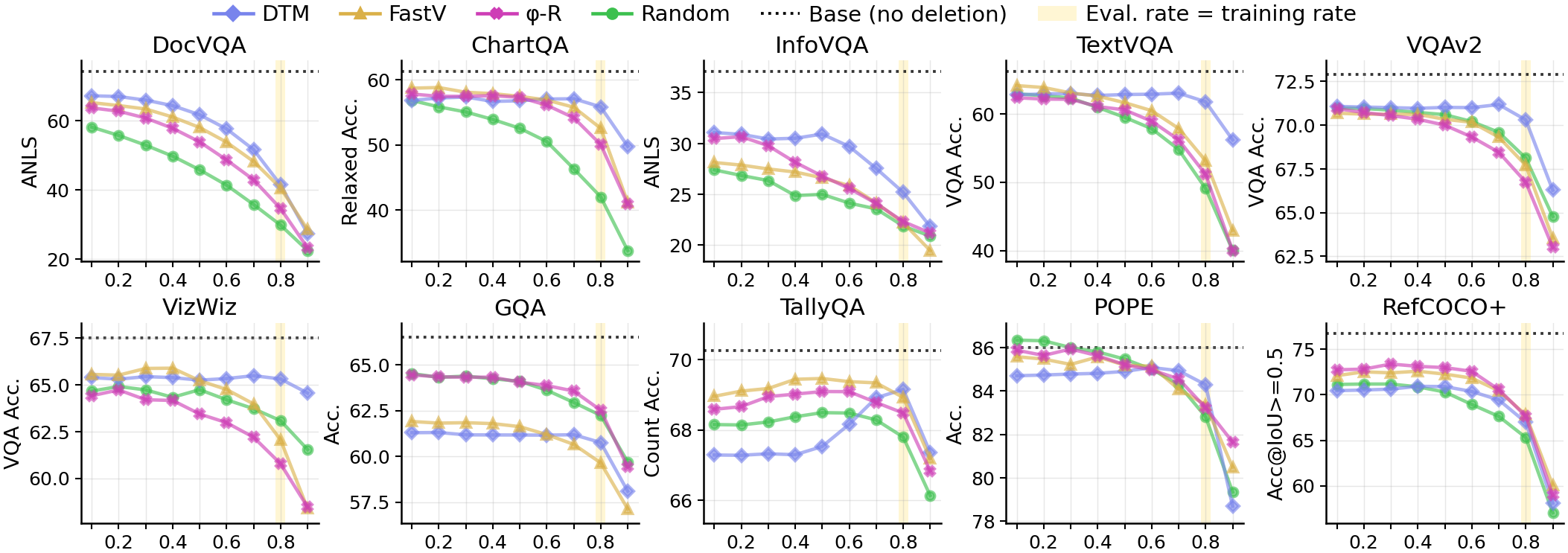}\\[-1pt]
{\small (b) Trained at 80\%}
\caption{\textbf{Fixed checkpoints on T5Gemma 2 270M--270M trained at 60\% (a) and 80\% (b) deletion.} Each curve evaluates one checkpoint at deletion rates from 10\% to 90\%; the training rate is highlighted. Blue diamonds: \dtm; gold triangles: \fastv; magenta crosses: \phir; green circles: \Random; dotted line: unpruned Base, fine-tuned the same way. Trained at 60\%, \dtm leads on all ten benchmarks at 80\% inference deletion (eight at 90\%); trained at 80\%, on eight and six.}
\label{fig:fixed-270m-6080}
\end{figure}

Figures~\ref{fig:fixed-t5-2040}--\ref{fig:fixed-270m-6080} repeat the fixed-checkpoint evaluation of Figures~\ref{fig:t5-train60} and~\ref{fig:gemma-train60} for checkpoints trained at 20\%, 40\%, 60\%, and 80\% deletion on all three backbones, comparing methods trained at the same target. Three patterns hold across training targets. First, \dtm's advantage appears when the inference budget is tight: at 80\% inference deletion, \dtm leads the strongest baseline on 5, 9, 10, and 10 of the T5Gemma 2 1B--1B benchmarks for training targets of 20\%, 40\%, 60\%, and 80\%, and on 10, 10, 10, and 9 of the Gemma 3 benchmarks. Second, under mild inference deletion (20\%), the methods are close, and \dtm leads on at most four benchmarks for any backbone and training target. Third, the decoder-only Gemma 3 transfers best to tighter budgets: its \dtm checkpoints trained at only 20\% still lead on all ten benchmarks at 80\% and 90\% deletion, by 6.9 and 8.1 points on average. Used at 80\% deletion, the 20\%-trained Gemma 3 \dtm checkpoint loses 2.5 points on average relative to training at 80\%, against 5.4--7.1 for the baselines; on T5Gemma 2 1B--1B, \dtm's loss (7.4) is similar to that of the attention-based selectors (5.6--7.4) (Figure~\ref{fig:rate-transfer}). The training target still shapes the trade-off (Appendices~\ref{app:budget-analysis} and~\ref{app:rate-transfer}).

\subsection{T5Gemma 2 270M--270M}
\label{app:small-model}

We apply the same L5 deletion procedure to the smaller T5Gemma 2 backbone (Figures~\ref{fig:270m-matched}, \ref{fig:fixed-270m-2040}, and~\ref{fig:fixed-270m-6080}). Across the ten benchmarks, \dtm's mean margin over the strongest baseline increases from 0.62 score points at matched 60\% deletion to 1.87 at 80\%. At 80\%, it scores 41.47 ANLS on DocVQA vs.\ \fastv's 40.58, and 55.83 relaxed accuracy on ChartQA vs.\ 52.55. The gains are task-dependent. TextVQA improves by 8.63 points over \fastv, while GQA trails \phir by 1.76 and RefCOCO+ trails \fastv by 0.82 (66.97 vs.\ 67.79). At 80\% inference deletion, a 60\%-trained checkpoint outperforms the strongest baseline on each of the ten tasks. This backbone also benefits most from deleting later (Appendix~\ref{app:depth-details}).

\subsection{Fixed-Checkpoint Budget Summaries}
\label{app:budget-analysis}

For a fixed checkpoint, we summarize scores $s_i$ at common inference deletion rates $r_i\in\{0.1,0.2,\ldots,0.9\}$ by normalized trapezoidal area:
\begin{equation}
 \overline s=\frac{1}{0.9-0.1}\sum_{i=1}^{8}
 \frac{s_i+s_{i+1}}{2}(r_{i+1}-r_i).
\end{equation}
This averages performance over the deletion rate axis. Table~\ref{tab:fixed-auc} reports each benchmark separately because the metrics differ: \dtm has the largest area on all ten T5Gemma 2 1B--1B benchmarks, nine Gemma 3 benchmarks, and eight T5Gemma 2 270M--270M benchmarks. Table~\ref{tab:training-target-auc} repeats the ten-benchmark mean for each training target: \dtm's mean area peaks at 60\% training on both T5Gemma 2 sizes and at 40\% on Gemma 3, and it is the highest of the four methods at every training target on all three backbones.

\begin{table}[htbp]
\centering
\caption{\textbf{Normalized area over inference deletion rates 10--90\%.} All checkpoints are trained at 60\% deletion and evaluated at nine deletion rates; the area is the trapezoidal integral divided by the 0.8-wide interval, so values remain on each task's 0--100 metric scale. The final row weights tasks equally. Bold marks the strongest method within each model. \dtm has the highest area on 10 of 10 T5Gemma 2 1B--1B benchmarks, 9 of 10 Gemma 3 4B benchmarks and 8 of 10 T5Gemma 2 270M--270M benchmarks, and the highest equal-weight mean on all three backbones.}
\label{tab:fixed-auc}
\label{tab:fixed-auc-gemma3}
\label{tab:fixed-auc-270m270m}
\small
\setlength{\tabcolsep}{2pt}
\renewcommand{\arraystretch}{1.12}
\resizebox{\linewidth}{!}{%
\begin{tabular}{@{}l@{\hspace{8pt}}*{4}{c}@{\hspace{9pt}}*{4}{c}@{\hspace{9pt}}*{4}{c}@{}}
\toprule
& \multicolumn{4}{c@{\hspace{9pt}}}{T5Gemma 2 1B--1B}
& \multicolumn{4}{c@{\hspace{9pt}}}{Gemma 3 4B}
& \multicolumn{4}{c@{}}{T5Gemma 2 270M--270M} \\
\cmidrule(l{2pt}r{9pt}){2-5}
\cmidrule(l{2pt}r{9pt}){6-9}
\cmidrule(l{2pt}){10-13}
Task
& \dtm & \fastv & \phir & \Random
& \dtm & \fastv & \phir & \Random
& \dtm & \fastv & \phir & \Random \\
\midrule
DocVQA
& \textbf{69.82} & 66.39 & 64.98 & 56.88
& \textbf{71.59} & 68.35 & 67.93 & 60.51
& \textbf{58.31} & 56.60 & 52.81 & 47.61 \\
ChartQA
& \textbf{63.76} & 61.37 & 61.49 & 58.39
& \textbf{69.65} & 67.19 & 67.52 & 65.70
& \textbf{57.15} & 57.01 & 56.20 & 51.30 \\
InfoVQA
& \textbf{39.78} & 38.37 & 38.44 & 34.44
& \textbf{47.33} & 45.85 & 44.81 & 42.71
& \textbf{29.41} & 28.12 & 28.93 & 27.13 \\
TextVQA
& \textbf{69.03} & 68.12 & 66.07 & 63.68
& \textbf{75.98} & 73.25 & 72.37 & 70.21
& \textbf{62.07} & 59.51 & 55.94 & 58.18 \\
VQAv2
& \textbf{76.12} & 74.20 & 73.81 & 74.54
& \textbf{76.80} & 75.94 & 76.04 & 75.59
& \textbf{70.95} & 69.55 & 69.54 & 70.47 \\
VizWiz
& \textbf{69.82} & 68.21 & 67.38 & 67.72
& 71.51 & \textbf{72.27} & 71.45 & 71.45
& \textbf{65.74} & 64.20 & 64.15 & 64.97 \\
GQA
& \textbf{70.02} & 69.20 & 69.18 & 68.77
& \textbf{68.50} & 67.22 & 67.08 & 66.97
& 63.36 & 63.01 & \textbf{64.52} & 64.12 \\
TallyQA
& \textbf{72.69} & 70.66 & 71.07 & 70.55
& \textbf{72.12} & 71.52 & 71.22 & 70.69
& \textbf{69.15} & 68.53 & 68.94 & 67.59 \\
POPE
& \textbf{84.58} & 83.32 & 82.85 & 83.59
& \textbf{84.63} & 82.64 & 82.92 & 83.08
& \textbf{84.41} & 83.85 & 83.96 & 83.32 \\
RefCOCO+
& \textbf{77.47} & 74.48 & 73.72 & 75.30
& \textbf{74.77} & 73.55 & 73.58 & 73.52
& 70.30 & 71.43 & \textbf{71.53} & 71.33 \\
\midrule
\begin{tabular}[c]{@{}l@{}}Mean benchmark\\score\end{tabular}
& \textbf{69.31} & 67.43 & 66.90 & 65.39
& \textbf{71.29} & 69.78 & 69.49 & 68.04
& \textbf{63.08} & 62.18 & 61.65 & 60.60 \\
\bottomrule
\end{tabular}%
}
\end{table}

\begin{table}[htbp]
\centering
\caption{\textbf{Mean normalized AUC over 10--90\% inference deletion, by training target.} Each mean weights the ten benchmarks equally on their 0--100 score scales. Bold marks the strongest training target for each method within each backbone. The final row counts tasks where the 60\%-trained method exceeds all three other methods in AUC. \dtm's mean area peaks at 60\% training on T5Gemma 2 1B--1B, 40\% on Gemma 3 4B and 60\% on T5Gemma 2 270M--270M, and \dtm has the highest mean area at every training target on all three backbones.}
\label{tab:training-target-auc}
\small
\setlength{\tabcolsep}{2pt}
\renewcommand{\arraystretch}{1.12}
\resizebox{\linewidth}{!}{%
\begin{tabular}{@{}l@{\hspace{8pt}}*{4}{c}@{\hspace{9pt}}*{4}{c}@{\hspace{9pt}}*{4}{c}@{}}
\toprule
& \multicolumn{4}{c@{\hspace{9pt}}}{T5Gemma 2 1B--1B}
& \multicolumn{4}{c@{\hspace{9pt}}}{Gemma 3 4B}
& \multicolumn{4}{c@{}}{T5Gemma 2 270M--270M} \\
\cmidrule(l{2pt}r{9pt}){2-5}
\cmidrule(l{2pt}r{9pt}){6-9}
\cmidrule(l{2pt}){10-13}
Training rate
& \dtm & \fastv & \phir & \Random
& \dtm & \fastv & \phir & \Random
& \dtm & \fastv & \phir & \Random \\
\midrule
20\%
& 67.67 & 66.50 & 65.25 & 64.93
& 71.58 & 68.40 & 67.99 & 66.47
& 62.01 & 61.17 & 60.86 & 60.07 \\
40\%
& 68.84 & 67.13 & 66.04 & \textbf{65.40}
& \textbf{71.92} & 69.32 & 68.60 & 67.61
& 62.69 & 61.95 & 61.08 & \textbf{60.65} \\
60\%
& \textbf{69.31} & \textbf{67.43} & \textbf{66.90} & 65.39
& 71.29 & \textbf{69.78} & \textbf{69.49} & \textbf{68.04}
& \textbf{63.08} & \textbf{62.18} & \textbf{61.65} & 60.60 \\
80\%
& 68.09 & 67.39 & 66.77 & 64.54
& 70.32 & 69.28 & 69.13 & 67.30
& 62.33 & 61.51 & 61.06 & 59.52 \\
\midrule
\begin{tabular}[c]{@{}l@{}}Tasks led\\at 60\% training\end{tabular}
& 10/10 & 0/10 & 0/10 & 0/10
& 9/10 & 1/10 & 0/10 & 0/10
& 8/10 & 0/10 & 2/10 & 0/10 \\
\bottomrule
\end{tabular}%
}
\end{table}

\begin{table}[htbp]
\centering
\caption{\textbf{Largest tested deletion rate retaining at least 90\% of Base.} All checkpoints are trained at 60\% deletion and evaluated at 10--90\% deletion in 10-point increments; Base is the separately fine-tuned unpruned model. None indicates that no tested rate meets this threshold. \dtm reaches the largest budget (ties included) on 9 of 10, 9 of 10, 10 of 10 benchmarks for T5Gemma 2 1B--1B, Gemma 3 4B and T5Gemma 2 270M--270M, respectively, but no \dtm checkpoint preserves 90\% of Base on InfoVQA.}
\label{tab:retention-budgets}
\label{tab:retention-budgets-gemma3}
\label{tab:retention-budgets-270m270m}
\small
\setlength{\tabcolsep}{2pt}
\renewcommand{\arraystretch}{1.12}
\resizebox{\linewidth}{!}{%
\begin{tabular}{@{}l@{\hspace{8pt}}*{4}{c}@{\hspace{9pt}}*{4}{c}@{\hspace{9pt}}*{4}{c}@{}}
\toprule
& \multicolumn{4}{c@{\hspace{9pt}}}{T5Gemma 2 1B--1B}
& \multicolumn{4}{c@{\hspace{9pt}}}{Gemma 3 4B}
& \multicolumn{4}{c@{}}{T5Gemma 2 270M--270M} \\
\cmidrule(l{2pt}r{9pt}){2-5}
\cmidrule(l{2pt}r{9pt}){6-9}
\cmidrule(l{2pt}){10-13}
Task
& \dtm & \fastv & \phir & \Random
& \dtm & \fastv & \phir & \Random
& \dtm & \fastv & \phir & \Random \\
\midrule
DocVQA
& 50\% & 40\% & 40\% & 10\%
& 50\% & 40\% & 40\% & 20\%
& 30\% & 20\% & 20\% & None \\
ChartQA
& 70\% & 60\% & 60\% & 30\%
& 70\% & 60\% & 60\% & 40\%
& 80\% & 70\% & 60\% & 40\% \\
InfoVQA
& None & 30\% & 30\% & None
& None & 30\% & 30\% & None
& None & None & None & None \\
TextVQA
& 80\% & 70\% & 60\% & 50\%
& 80\% & 70\% & 70\% & 60\%
& 70\% & 60\% & 40\% & 50\% \\
VQAv2
& 80\% & 80\% & 70\% & 80\%
& 80\% & 80\% & 80\% & 80\%
& 80\% & 80\% & 70\% & 80\% \\
VizWiz
& 90\% & 80\% & 70\% & 90\%
& 90\% & 80\% & 80\% & 80\%
& 90\% & 80\% & 80\% & 90\% \\
GQA
& 80\% & 80\% & 80\% & 80\%
& 90\% & 80\% & 80\% & 80\%
& 80\% & 70\% & 80\% & 80\% \\
TallyQA
& 80\% & 80\% & 80\% & 80\%
& 90\% & 80\% & 80\% & 80\%
& 90\% & 80\% & 90\% & 80\% \\
POPE
& 90\% & 80\% & 80\% & 80\%
& 90\% & 80\% & 80\% & 90\%
& 90\% & 80\% & 80\% & 80\% \\
RefCOCO+
& 80\% & 70\% & 60\% & 70\%
& 80\% & 70\% & 70\% & 70\%
& 70\% & 70\% & 70\% & 70\% \\
\bottomrule
\end{tabular}%
}
\end{table}

We also report the largest tested inference deletion rate whose score is at least $0.9$ times the separately trained unpruned Base model in Table~\ref{tab:retention-budgets}. ``None'' means that no tested rate meets the criterion, not that no possible configuration could meet it.

\FloatBarrier
\subsection{Checkpoint Reuse Across Deletion Budgets}
\label{app:rate-transfer}

\begin{figure}[!b]
\centering
\includegraphics[width=0.75\linewidth]{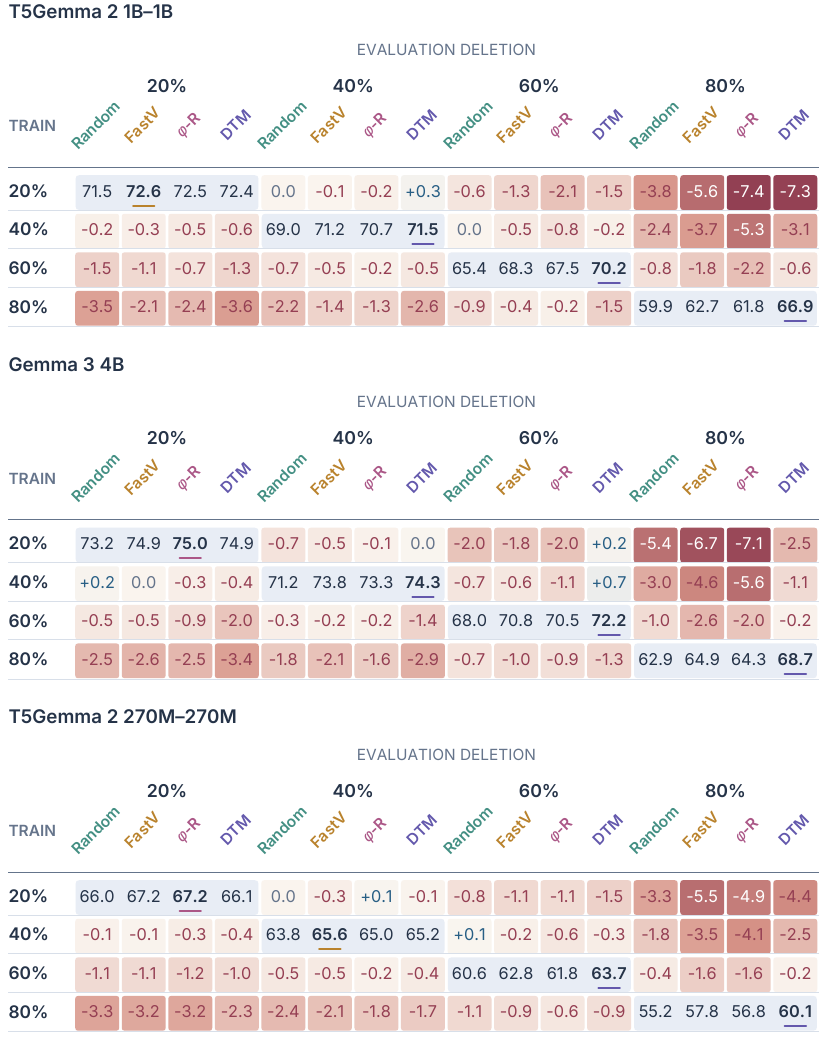}
\caption{\textbf{Checkpoint reuse across deletion budgets at L5.} Rows specify training targets; grouped columns specify inference deletion. Gray diagonal cells show matched scores, with the strongest method in bold and underlined. Off-diagonal cells show $S_t(e)-S_e(e)$; negative values favor training at the evaluation rate. All cells average the same ten benchmarks equally. Red shading deepens with the size of the penalty. A 60\%-trained \dtm checkpoint used at 80\% costs at most 0.6 points on average, and on Gemma 3 even a 20\%-trained \dtm checkpoint loses less at 80\% than any baseline.}
\label{fig:rate-transfer}
\end{figure}

For a given method and task, let $S_t(e)$ denote the score of a checkpoint trained at deletion target $t$ and evaluated at rate $e$. We measure the effect of checkpoint reuse as $S_t(e)-S_e(e)$, holding the inference budget fixed.

Figure~\ref{fig:rate-transfer} separates the loss associated with a tighter budget from the additional cost of checkpoint reuse. For \dtm, moving from matched 60\% to matched 80\% deletion lowers the mean score by 3.3, 3.5, and 3.6 points for T5Gemma 2 1B--1B, Gemma 3 4B, and T5Gemma 2 270M--270M, respectively. Reusing the 60\%-trained checkpoint at 80\% incurs only a further 0.6, 0.2, and 0.2 points. Matched training is not uniformly best: at 60\% inference deletion, Gemma 3's 40\%-trained \vdtm checkpoints average 0.7 points above their 60\%-trained counterparts. These means summarize heterogeneous task metrics and are descriptive comparisons of the selected checkpoints.

\subsection{Attention-Score Aggregation}
\label{app:query-aggregation}

\begin{figure}[tbp]
\centering
\includegraphics[width=0.8\linewidth]{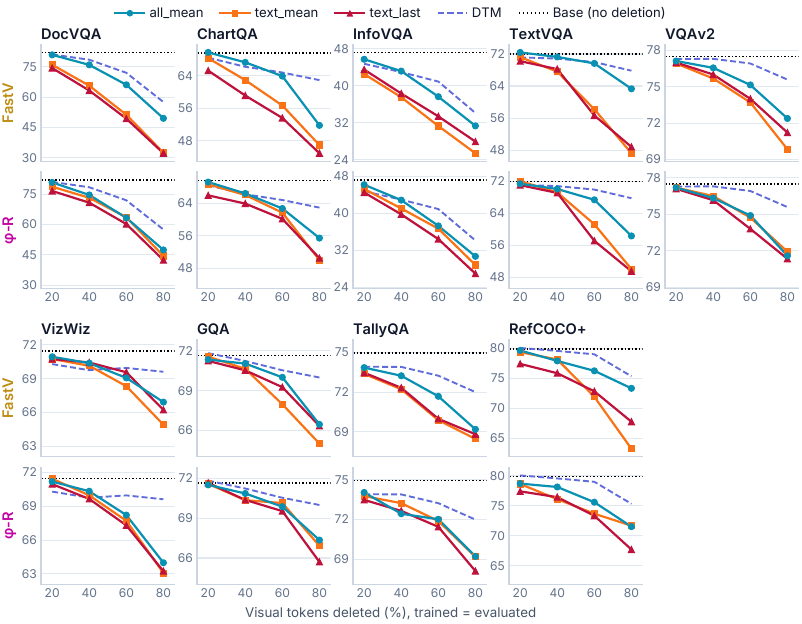}
\caption{\textbf{Query aggregation for \fastv and \phir on T5Gemma 2 1B--1B, matched deletion after layer 5.} Each selector ranks visual tokens by attention averaged over all valid queries (\texttt{all\_mean}, used in the main results), over text queries only (\texttt{text\_mean}), or from the final text query (\texttt{text\_last}); each variant is separately fine-tuned. Dashed: \dtm; dotted: unpruned Base. All-query scoring is the strongest \fastv variant almost everywhere, by up to 17 points on DocVQA.}
\label{fig:agg-1b1b}
\end{figure}

\begin{figure}[tbp]
\centering
\includegraphics[width=0.8\linewidth]{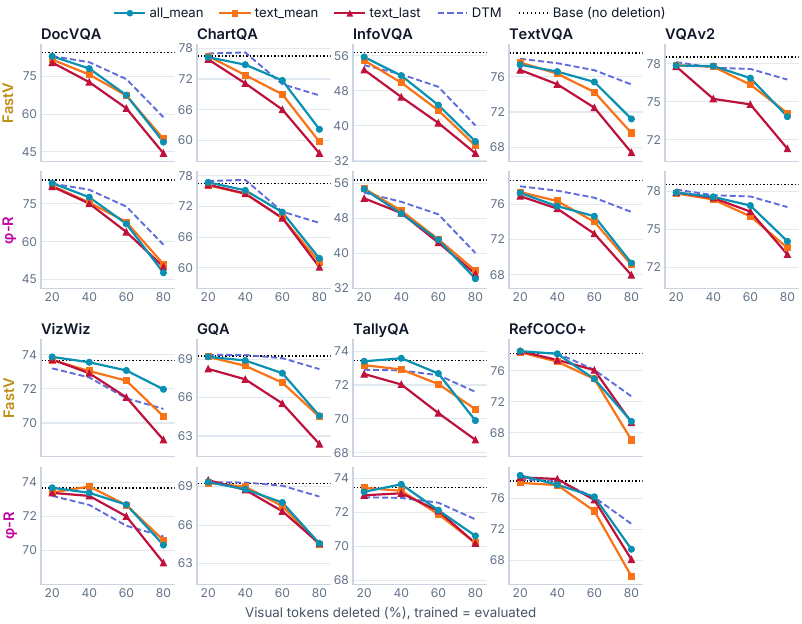}
\caption{\textbf{Query aggregation for \fastv and \phir on Gemma 3 4B, matched deletion after five decoder layers during prefill.} Variants and line styles as in Figure~\ref{fig:agg-1b1b}. \fastv again prefers all-query scoring (35 of 36 settings against \texttt{text\_last}), whereas \phir gains from text queries on DocVQA and InfoVQA at 80\%.}
\label{fig:agg-gemma3}
\end{figure}

\begin{figure}[tbp]
\centering
\includegraphics[width=0.8\linewidth]{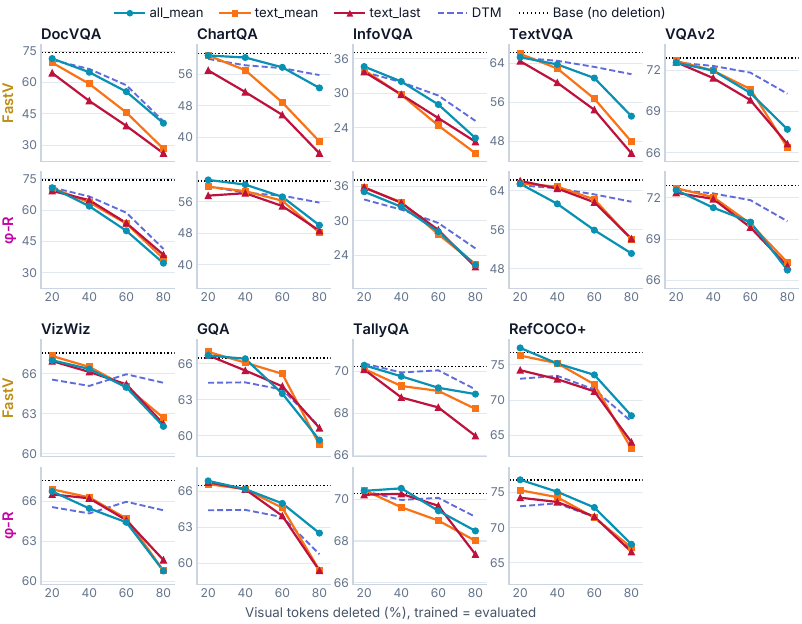}
\caption{\textbf{Query aggregation for \fastv and \phir on T5Gemma 2 270M--270M, matched deletion after layer 5.} Variants and line styles as in Figure~\ref{fig:agg-1b1b}. On the smallest backbone, text queries help \phir on DocVQA and TextVQA but not \fastv; neither reaches \dtm at 60--80\% deletion on these tasks.}
\label{fig:agg-270m}
\end{figure}

We compare \texttt{all\_mean} (all valid queries), \texttt{text\_mean} (text queries only), and \texttt{text\_last} (the final text query), averaging heads in each case. Each variant is separately fine-tuned. \texttt{text\_mean} matches the ranking criterion of LearnPruner's second stage \citep{takezoe2026learnprunerrethinkingattentionbasedtoken}; we evaluate it under our pruning placement and task fine-tuning, omitting LearnPruner's learned first stage, which operates before the language model. Figures~\ref{fig:agg-1b1b}--\ref{fig:agg-270m} compare the three variants for \fastv and \phir at matched deletion rates on all three backbones, using the nine benchmarks evaluated with text queries on every backbone (all but POPE).

\fastv favors \texttt{all\_mean} in most settings: over \texttt{text\_last} in 99 of 108 comparisons across the three backbones, and over \texttt{text\_mean} in 87 of 108. Its advantage is particularly large for T5Gemma 2: at 80\% deletion on DocVQA, the scores are 49.48, 32.54, and 32.24 for all, mean-text, and last-text queries, and on T5Gemma 2 270M--270M ChartQA, 52.55 against 35.89 with \texttt{text\_last}. However, all-query aggregation is not universally best, particularly for \phir, which it favors in 68 of 108 and 84 of 108 comparisons. On Gemma 3 DocVQA at 80\%, \phir reaches 51.11 with \texttt{text\_mean}, vs.\ 47.66 with \texttt{all\_mean}; \dtm remains higher at 58.79. On T5Gemma 2 270M--270M, text queries raise \phir by up to 6.3 points on TextVQA (62.24 vs.\ 55.91 at 60\%), still below \dtm's 63.27. At 80\% deletion, only one text-query variant on any backbone scores above \dtm (\phir with \texttt{text\_mean} on 270M--270M RefCOCO+, 67.19 vs.\ 66.97). Thus, the headline baselines use a strong common aggregation choice, while these additional variants expose where it leaves performance available, without closing the gap to \dtm under aggressive deletion.

\subsection{Choosing the Deletion Layer}
\label{sec:depth-analysis}

\begin{figure}[tbp]
\centering
\includegraphics[width=0.9\linewidth]{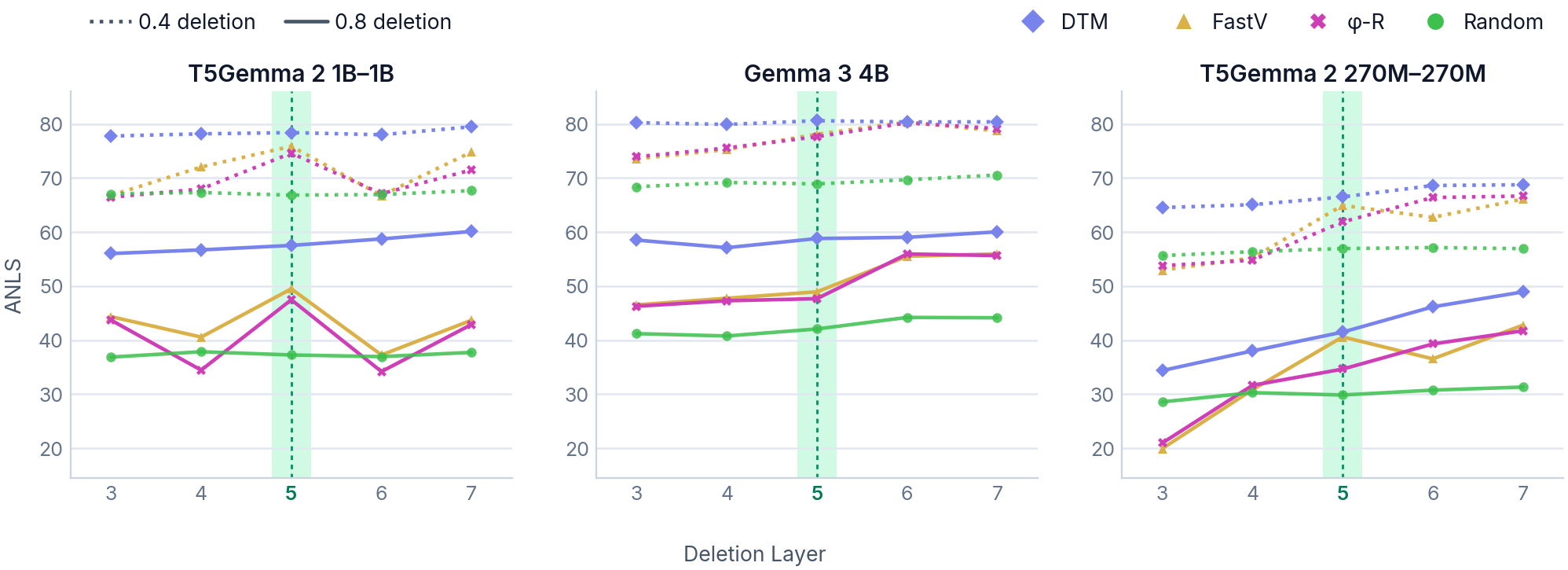}
\setlength{\abovecaptionskip}{7pt}
\caption{\textbf{Sensitivity to deletion depth on DocVQA.} Models are separately trained and evaluated at 40\% (dotted) or 80\% (solid) deletion with the gate after each layer. The horizontal axis gives the number of layers before deletion; shading marks L5, the setting used in our main experiments. On T5Gemma 2 1B--1B, the attention-based selectors drop when they read the first global-attention layer (L6), whereas \dtm does not.}
\label{fig:depth-analysis}
\end{figure}

\dtm learns token importance from contextual hidden states rather than attention weights, allowing it to operate across sliding-window and global attention without changing its scoring rule. Figure~\ref{fig:depth-analysis} demonstrates this robustness on T5Gemma 2: at 80\% deletion on the 1B--1B model, moving the gate from after layer 5 (L5), where attention-based selectors read a sliding-window layer, to after layer 6 (L6), where they read the first global-attention layer, reduces \fastv from 49.5 to 37.2 ANLS and \phir from 47.5 to 34.2, while \dtm improves from 57.5 to 58.7. This contrast is less apparent on Gemma 3, whose larger 1,024-token window, compared with T5Gemma 2's roughly 256-token reach in each direction, may narrow the difference in attention visibility between local and global layers for these inputs. Robustness across attention patterns lets us place the gate early, where more layers benefit from the shortened sequence; we use L5 as a shared quality--compute compromise, although later deletion can improve accuracy at additional cost: at 80\% deletion on T5Gemma 2 1B--1B, deleting after layer 7 instead of layer 5 raises \dtm's ANLS by 2.6 but saves 53.4\% rather than 58.9\% of FLOPs (Appendix~\ref{app:flops-by-layer}). Appendix~\ref{app:depth-details} extends this analysis to 20--80\% deletion.

\subsection{Deletion Depth Across Backbones and Budgets}
\label{app:depth-details}

\begin{figure}[htbp]
\centering
\includegraphics[width=0.92\linewidth]{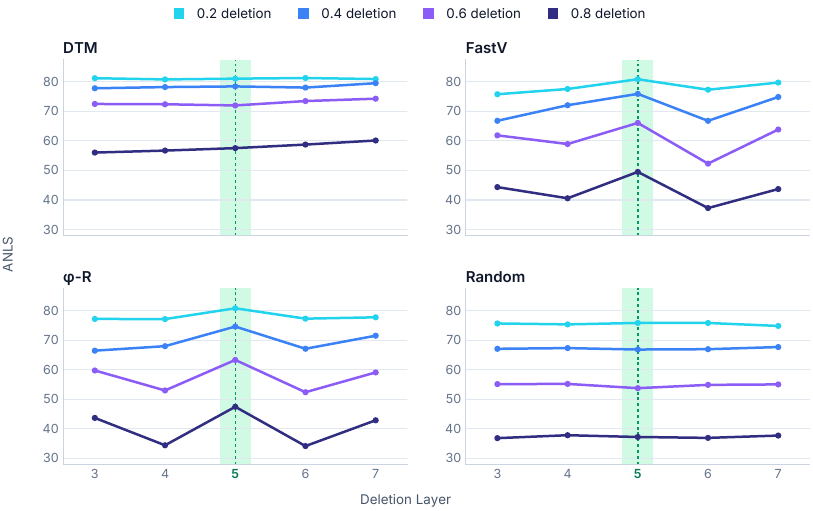}
\caption{\textbf{Deletion-depth sensitivity across budgets for T5Gemma 2 1B--1B on DocVQA.} Panels show individual methods; line colors indicate matched training and evaluation deletion rates (20\%, 40\%, 60\%, 80\%). The horizontal axis gives the number of layers before deletion; shading marks L5. \dtm is nearly flat across depth, whereas \fastv and \phir peak at L5 and fall at L6 at every rate.}
\label{fig:depth-t5-full}
\end{figure}

\begin{figure}[htbp]
\centering
\includegraphics[width=0.92\linewidth]{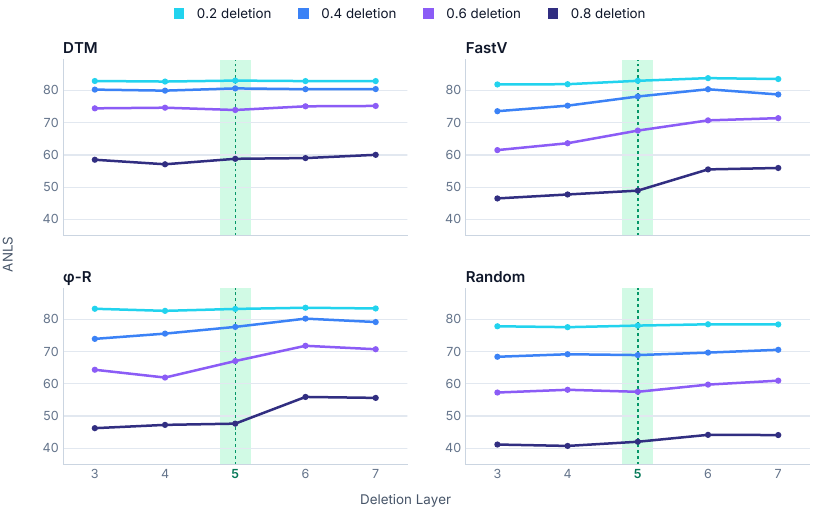}
\caption{\textbf{Deletion-depth sensitivity across budgets for Gemma 3 4B on DocVQA.} Panels show individual methods; line colors indicate matched deletion rates (20\%, 40\%, 60\%, 80\%); shading marks L5. The attention-based selectors improve with later deletion on this backbone, while \dtm stays nearly constant and is highest at 40--80\%.}
\label{fig:depth-gemma-full}
\end{figure}

\begin{figure}[htbp]
\centering
\includegraphics[width=0.92\linewidth]{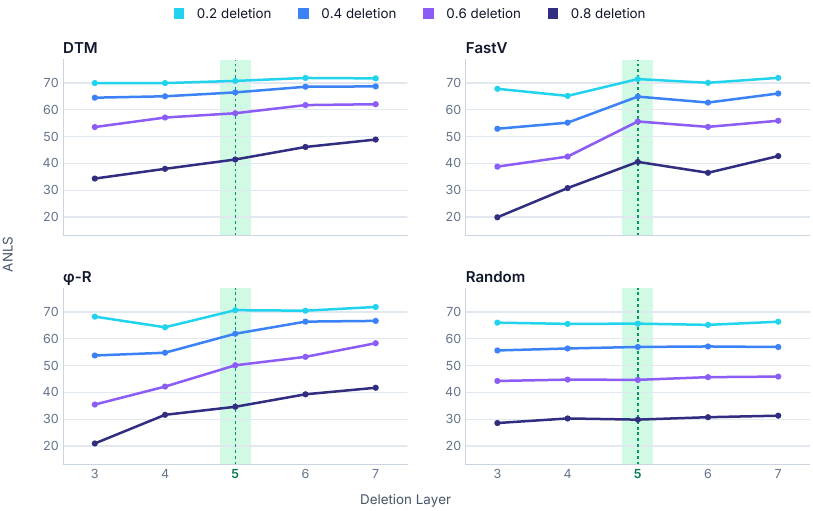}
\caption{\textbf{Deletion-depth sensitivity across budgets for T5Gemma 2 270M--270M on DocVQA.} Panels show individual methods; line colors indicate matched deletion rates (20\%, 40\%, 60\%, 80\%); shading marks L5. The smallest backbone benefits most from later deletion, especially at 80\%.}
\label{fig:depth-270m-full}
\end{figure}

Figures~\ref{fig:depth-t5-full}--\ref{fig:depth-270m-full} extend the depth analysis of Figure~\ref{fig:depth-analysis} to matched deletion rates of 20\%, 40\%, 60\%, and 80\%. Each point uses a separately fine-tuned checkpoint for its layer and deletion target, evaluated on the same 5,349 DocVQA validation examples.

\paragraph{Dependence on the deletion budget.}
Across L3--L7, \dtm achieves the highest ANLS at every layer on all three backbones at 40\%, 60\%, and 80\% deletion. Its advantage is less consistent at 20\%, where attention-based selection can match or exceed it. For example, at L6 on Gemma 3, \fastv scores 83.8 versus 82.9 for \dtm; at L7 on T5Gemma 2 270M--270M, the corresponding scores are 71.9 and 71.7. Some higher-deletion comparisons are also close: at 40\% deletion and L6 on Gemma 3, \dtm and \fastv differ by 0.02 ANLS. The overall ranking therefore includes settings where the methods perform nearly identically.

\paragraph{Model size and depth sensitivity.}
The smaller T5Gemma 2 backbone benefits increasingly from later deletion as the token budget tightens. Moving \dtm from L3 to L7 improves ANLS by 1.7, 4.2, 8.5, and 14.5 points at the four respective deletion rates. In contrast, \dtm on the two larger backbones varies little across depth at 20\% deletion, and its L3-to-L7 gains remain smaller even at 80\% (4.1 and 1.5 points). \Random deletion is comparatively insensitive to depth: its largest best-to-worst ranges across the tested rates are 1.4 ANLS on T5Gemma 2 1B--1B, 2.7 on T5Gemma 2 270M--270M, and 3.7 on Gemma 3. These results show that the value of additional processing before deletion depends on both the backbone and the token-selection method.

\paragraph{Which layer favors attention-based selection?}
On T5Gemma 2 1B--1B, both \fastv and \phir attain their highest ANLS at L5 at all four deletion rates. On Gemma 3, \phir instead peaks at L6 throughout, while \fastv peaks at L6 for 20\% and 40\% deletion and at L7 for 60\% and 80\%. On T5Gemma 2 270M--270M, both attention-based methods peak at L7, although \fastv drops from L5 to L6 before recovering. L6 reads the first global-attention layer, but the T5Gemma 2 1B--1B drop is not shared across backbones. These sweeps change the scoring layer and learned representations together, so they do not isolate a causal effect of attention type. Deleting later also saves fewer FLOPs (Table~\ref{tab:docvqa-layers-flops}).

\FloatBarrier
\section{What the Gate Keeps}
\label{app:gate-keeps}

\subsection{Spatial Evidence Retention at Other Deletion Rates}
\label{app:evidence-rates}

\begin{figure}[htbp]
\centering
\includegraphics[width=0.92\linewidth]{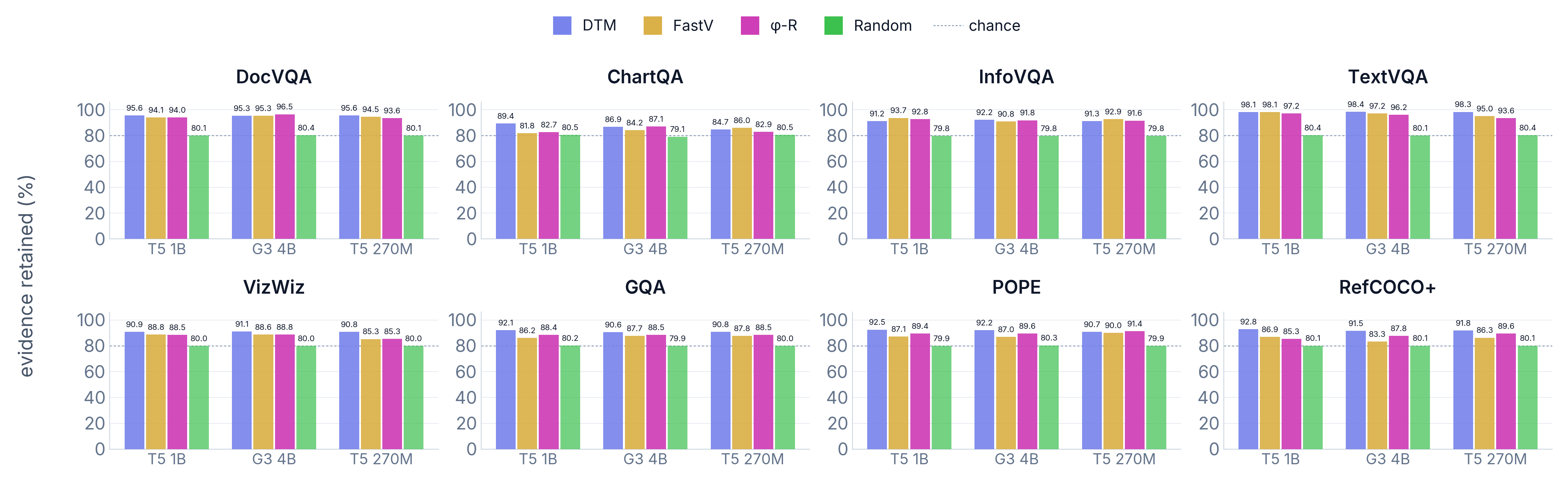}\\[2pt]
\includegraphics[width=0.92\linewidth]{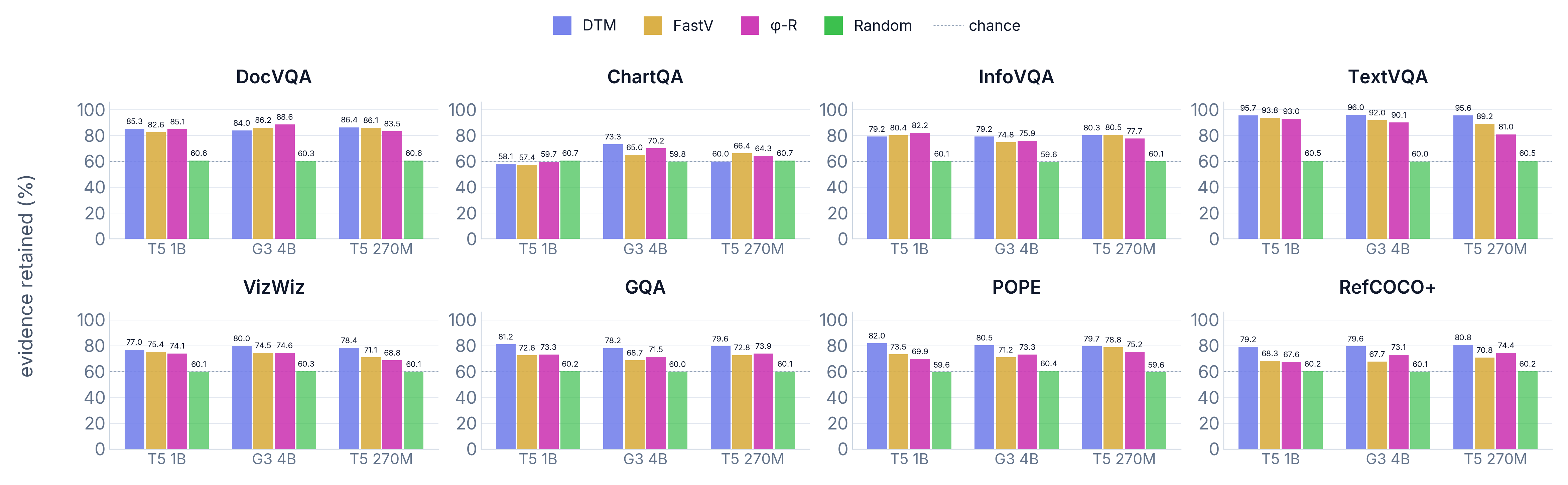}\\[2pt]
\includegraphics[width=0.92\linewidth]{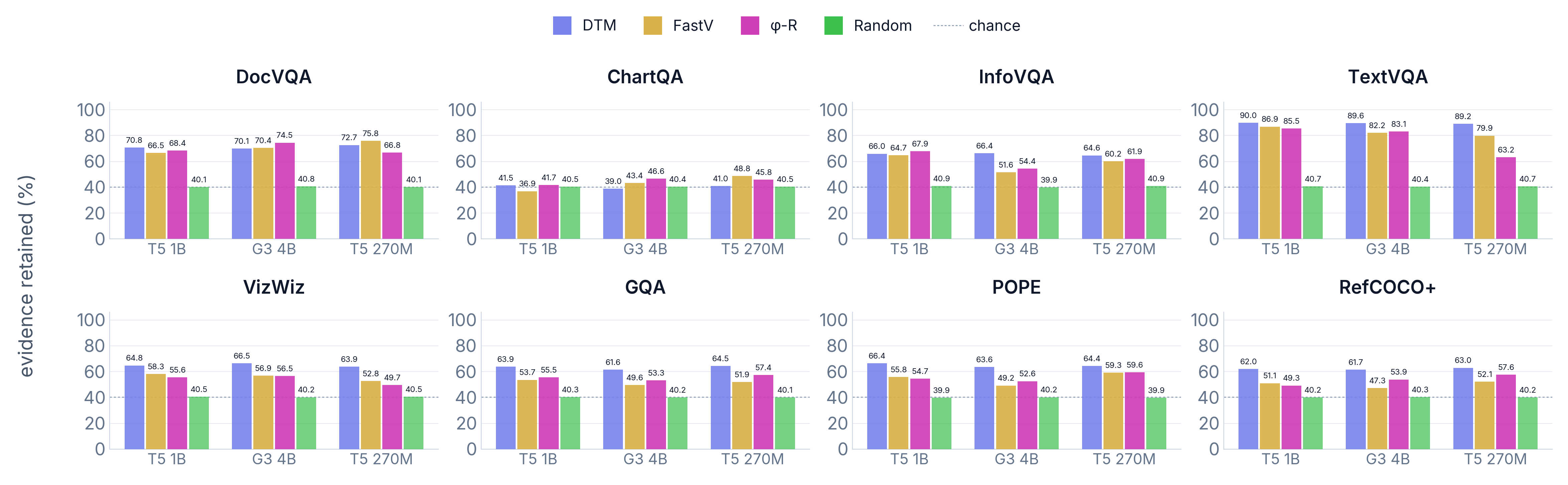}
\caption{\textbf{Spatial evidence retention at matched 20\% (top), 40\% (middle), and 60\% (bottom) visual token deletion after layer 5.} Bars show the mean percentage of evidence patches retained on the same eligible questions for each model--dataset pair; dashed lines mark chance. Blue: \dtm; gold: \fastv; magenta: \phir; green: \Random. \dtm retains the most evidence in 17, 19, and 18 of 24 comparisons at 20\%, 40\%, and 60\%, and its lead grows with the deletion rate.}
\label{fig:evidence-other-rates}
\end{figure}

Figure~\ref{fig:evidence-other-rates} repeats the analysis of Figure~\ref{fig:evidence-retained} at matched 20\%, 40\%, and 60\% deletion. Evidence regions come from each benchmark's spatial annotations where available (e.g., answer-bearing OCR words, object boxes, or referent boxes) and from grounding proxies otherwise; a patch counts as evidence if it overlaps such a region. Methods are compared on the same eligible questions within each model--dataset pair.

\subsection{Evidence, Correctness, and Selector Agreement}
\label{app:interp}

\begin{figure}[htbp]
\centering
\includegraphics[width=0.86\linewidth]{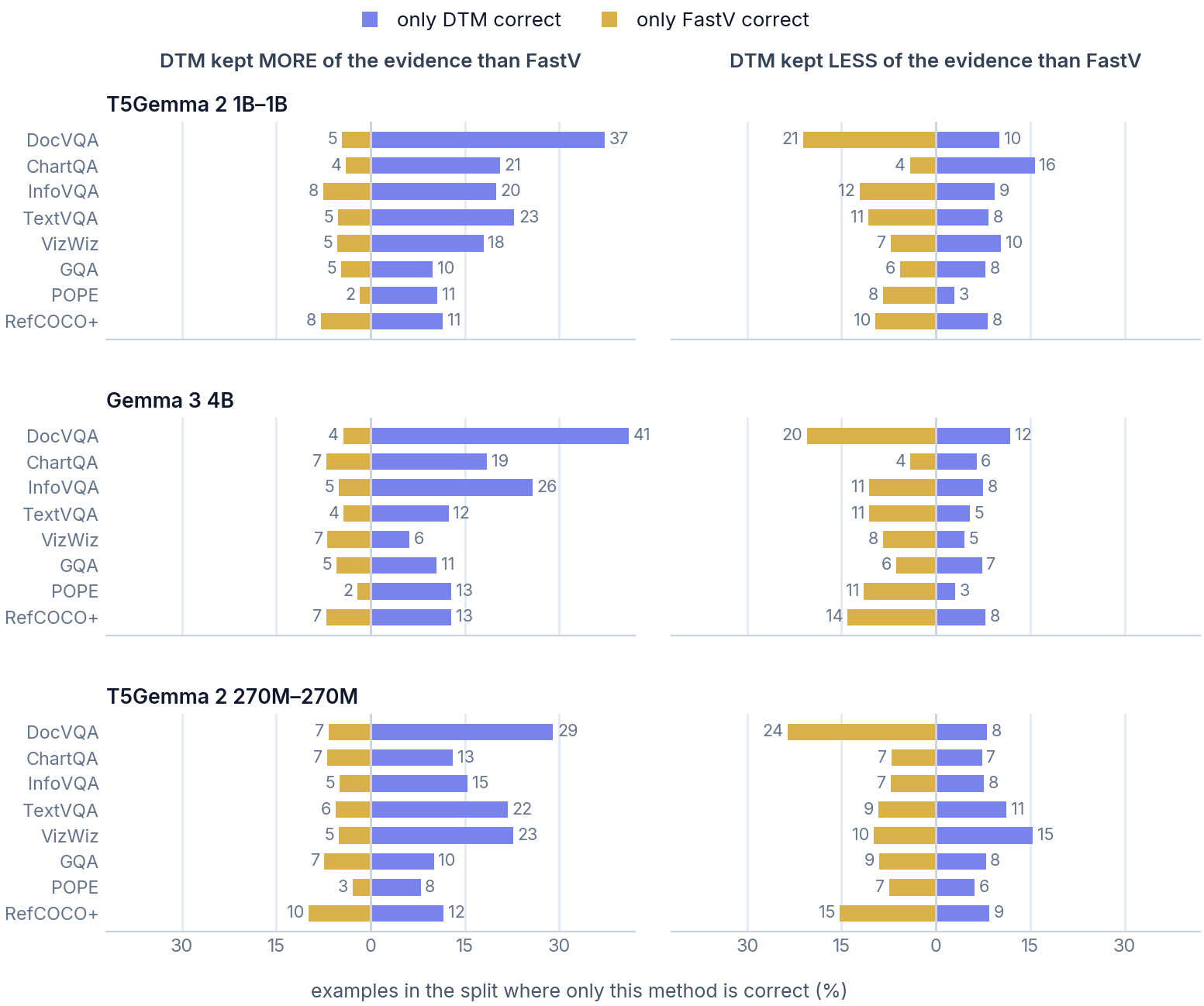}
\caption{\textbf{Evidence retention and which method answers correctly, at matched 80\% deletion.} For each model and benchmark, questions are split by whether \dtm kept more (left) or less (right) of the evidence patches than \fastv; bars give the percentage of questions in each split that only \dtm (blue) or only \fastv (gold) answers correctly. When \dtm keeps more evidence, it is the sole correct method more than twice as often as \fastv (13.5\% vs.\ 6.2\% of questions, pooled).}
\label{fig:evidence-paired}
\end{figure}

Figure~\ref{fig:evidence-paired} splits each model's 80\% evaluations by whether \dtm kept more or less of the evidence than \fastv and shows how often only one of the two answers correctly. Where \dtm keeps more evidence, it is the only correct method on 13.5\% of questions, against 6.2\% for \fastv alone (pooled over 72,884 questions), and it leads in 23 of 24 model--benchmark pairs; the gap is largest on DocVQA (37.3\% vs.\ 4.5\% on T5Gemma 2 1B--1B). Where it keeps less, the ordering reverses (8.0\% vs.\ 11.1\%). Retaining evidence is thus associated with which method answers correctly, although this comparison is correlational.

\begin{figure}[htbp]
\centering
\includegraphics[width=\linewidth]{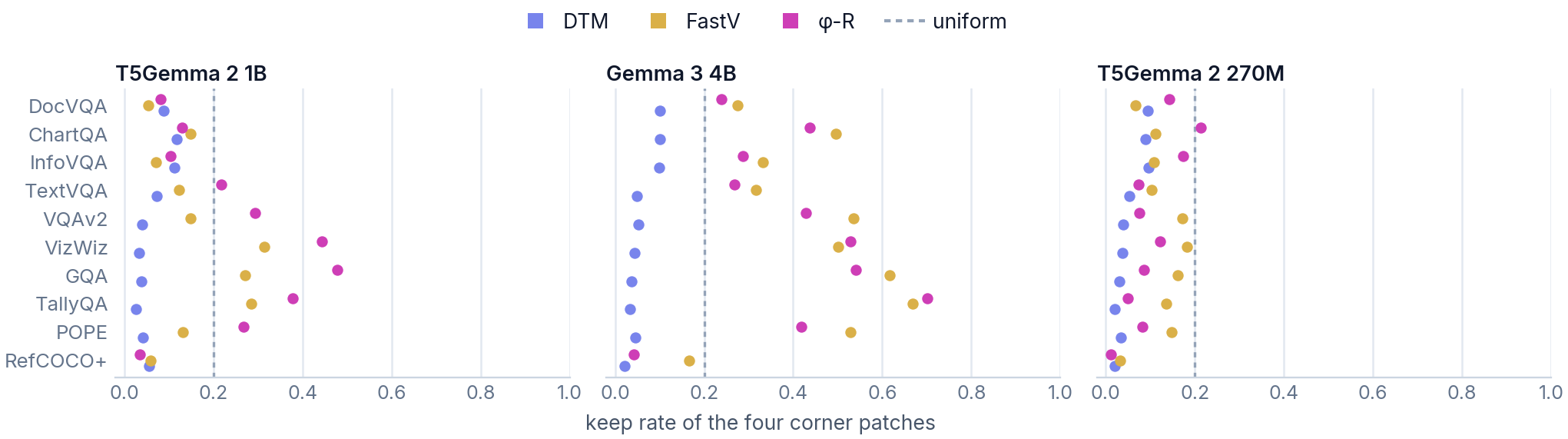}
\caption{\textbf{Keep rate of the four corner patches at matched 80\% deletion.} Blue: \dtm; gold: \fastv; magenta: \phir; the dashed line is the rate under uniform selection (0.20). On Gemma 3, the attention-based selectors keep the corners far more often than uniform selection would, whereas \dtm rarely does on any model.}
\label{fig:corners}
\end{figure}

Figure~\ref{fig:corners} shows one way the attention-based selectors spend budget that \dtm does not. On Gemma 3, \fastv and \phir keep the four corner patches of the image at median rates of 0.50 and 0.42, above the uniform rate of 0.20 on nine of ten benchmarks, whereas \dtm keeps them at 0.02--0.12 on every model and benchmark. The two attention-based selectors also agree with each other far more than either agrees with \dtm (median Cohen's $\kappa$ of 0.59, against 0.29 and 0.30), while \dtm's keep-sets agree across the three separately trained backbones (median $\kappa$ 0.63, against 0.34 for \fastv and 0.29 for \phir). Gates trained on different models thus converge on similar selections.

\subsection{Randomly Sampled Examples at 80\% Deletion}
\label{app:qualitative}

\begin{figure}[p]
\centering
\includegraphics[width=\linewidth,height=0.86\textheight,keepaspectratio]{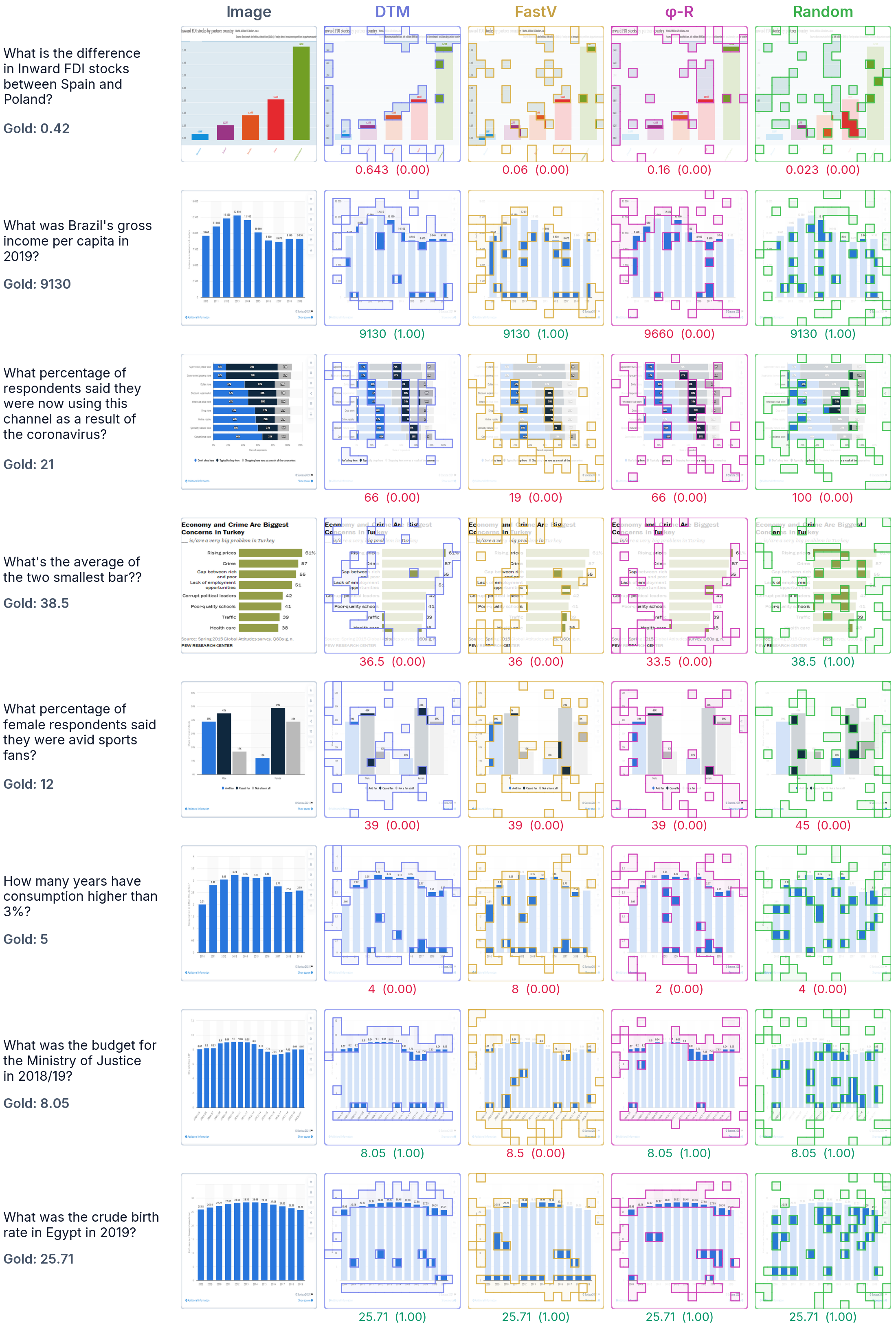}
\caption{\textbf{Randomly sampled ChartQA questions at matched 80\% deletion on T5Gemma 2 1B--1B (1 of 2).} Each row shows the input with its question and gold answer, then the 52 of 256 visual tokens kept by each method in full contrast (outlined), with deleted patches washed out. Below each panel: the prediction and its relaxed-accuracy score. Blue: \dtm; gold: \fastv; magenta: \phir; green: \Random. \dtm concentrates its budget on labels, tick values, and bar ends.}
\label{fig:qual-chartqa-1}
\end{figure}

\begin{figure}[p]
\centering
\includegraphics[width=\linewidth,height=0.86\textheight,keepaspectratio]{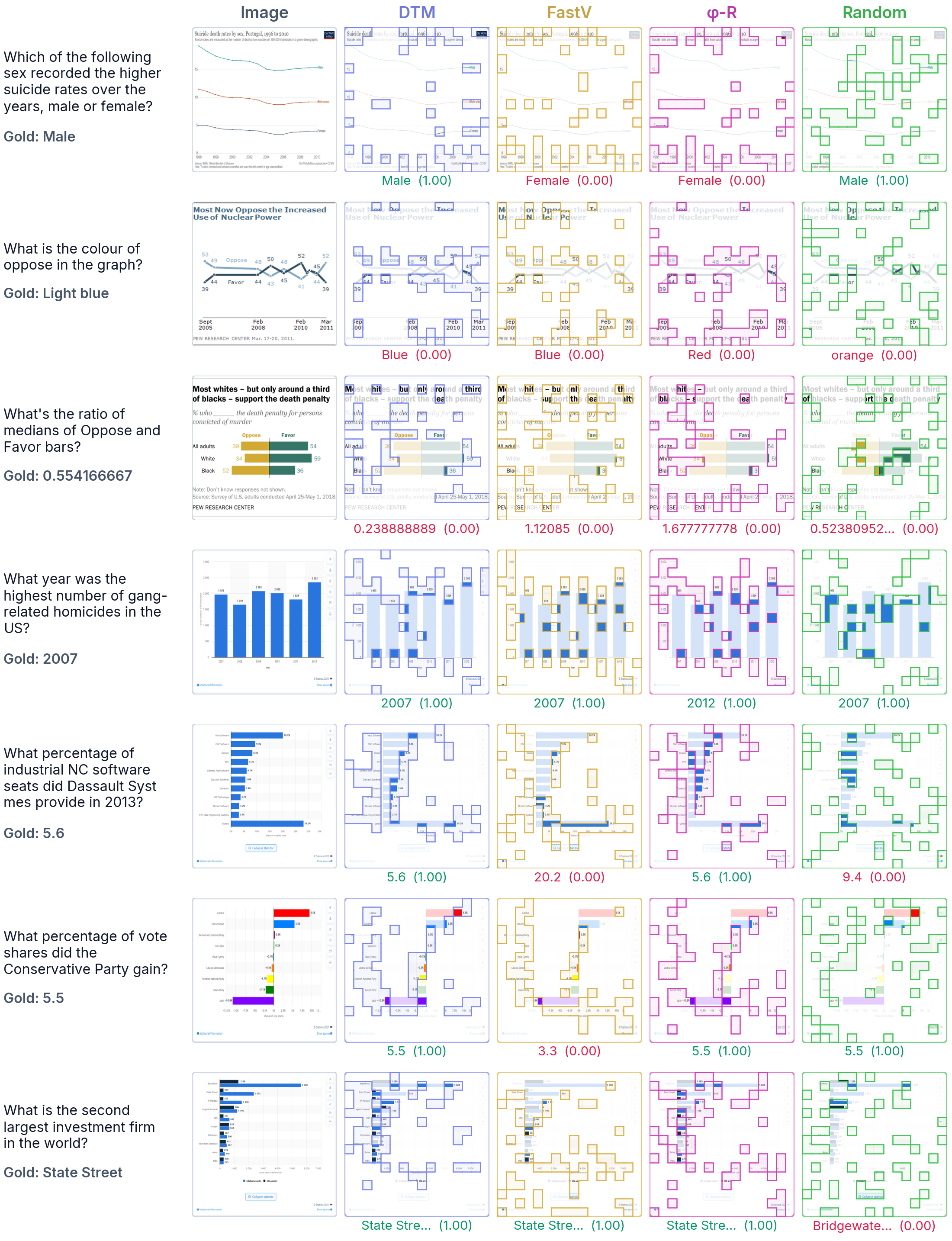}
\caption{\textbf{Randomly sampled ChartQA questions at matched 80\% deletion on T5Gemma 2 1B--1B (2 of 2).} Presented as in Figure~\ref{fig:qual-chartqa-1}: kept tokens in full contrast, predictions and relaxed-accuracy scores below each panel; blue \dtm, gold \fastv, magenta \phir, green \Random.}
\label{fig:qual-chartqa-2}
\end{figure}

\begin{figure}[p]
\centering
\includegraphics[width=\linewidth,height=0.86\textheight,keepaspectratio]{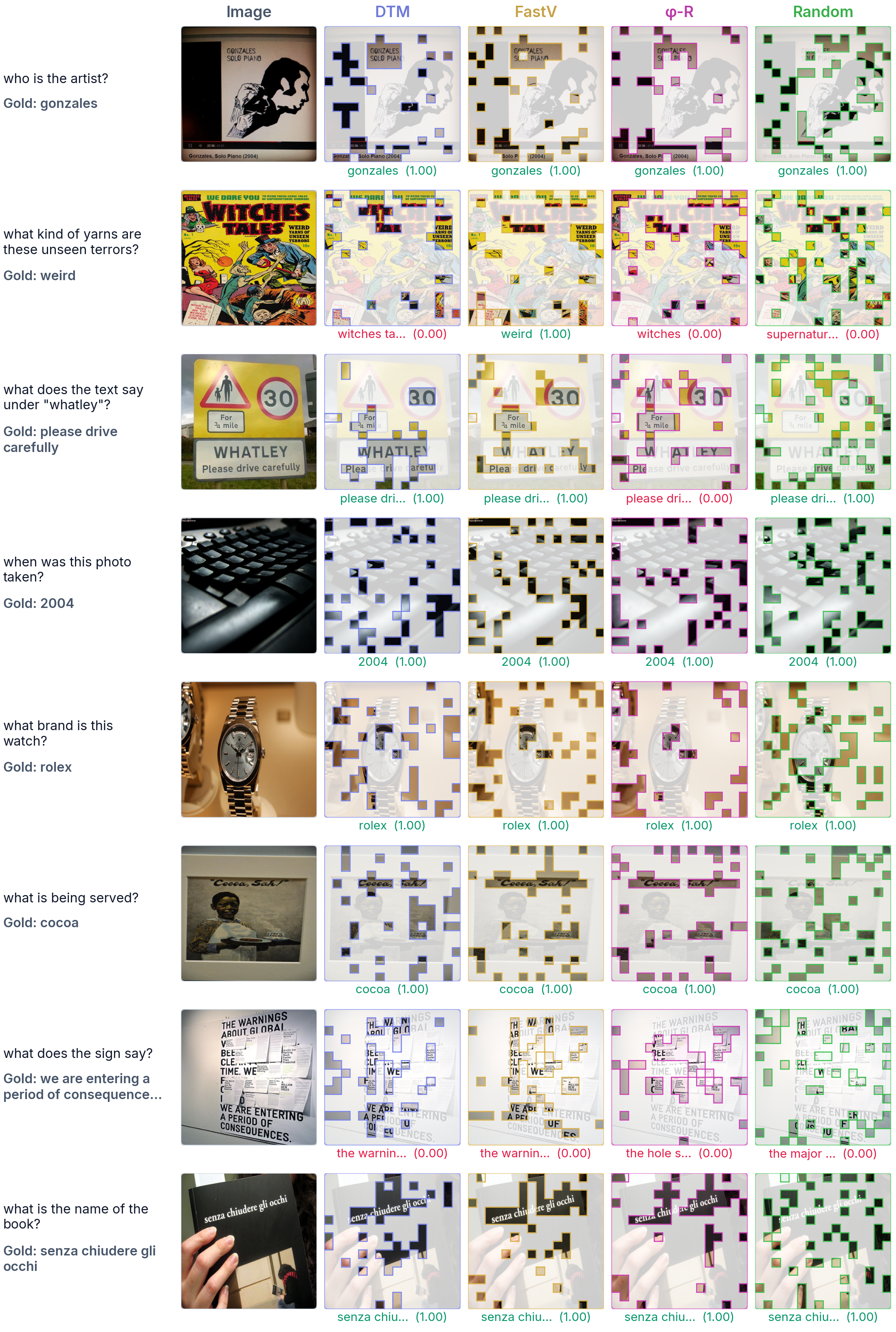}
\caption{\textbf{Randomly sampled TextVQA questions at matched 80\% deletion on T5Gemma 2 1B--1B (1 of 2).} Each row shows the input with its question and gold answer, then the 52 of 256 visual tokens kept by each method in full contrast (outlined), with deleted patches washed out; below each panel, the prediction (long answers truncated) and its VQA-accuracy score. Blue: \dtm; gold: \fastv; magenta: \phir; green: \Random. All three informed selectors keep much of the scene text; they differ in which words remain.}
\label{fig:qual-textvqa-1}
\end{figure}

\begin{figure}[p]
\centering
\includegraphics[width=\linewidth,height=0.86\textheight,keepaspectratio]{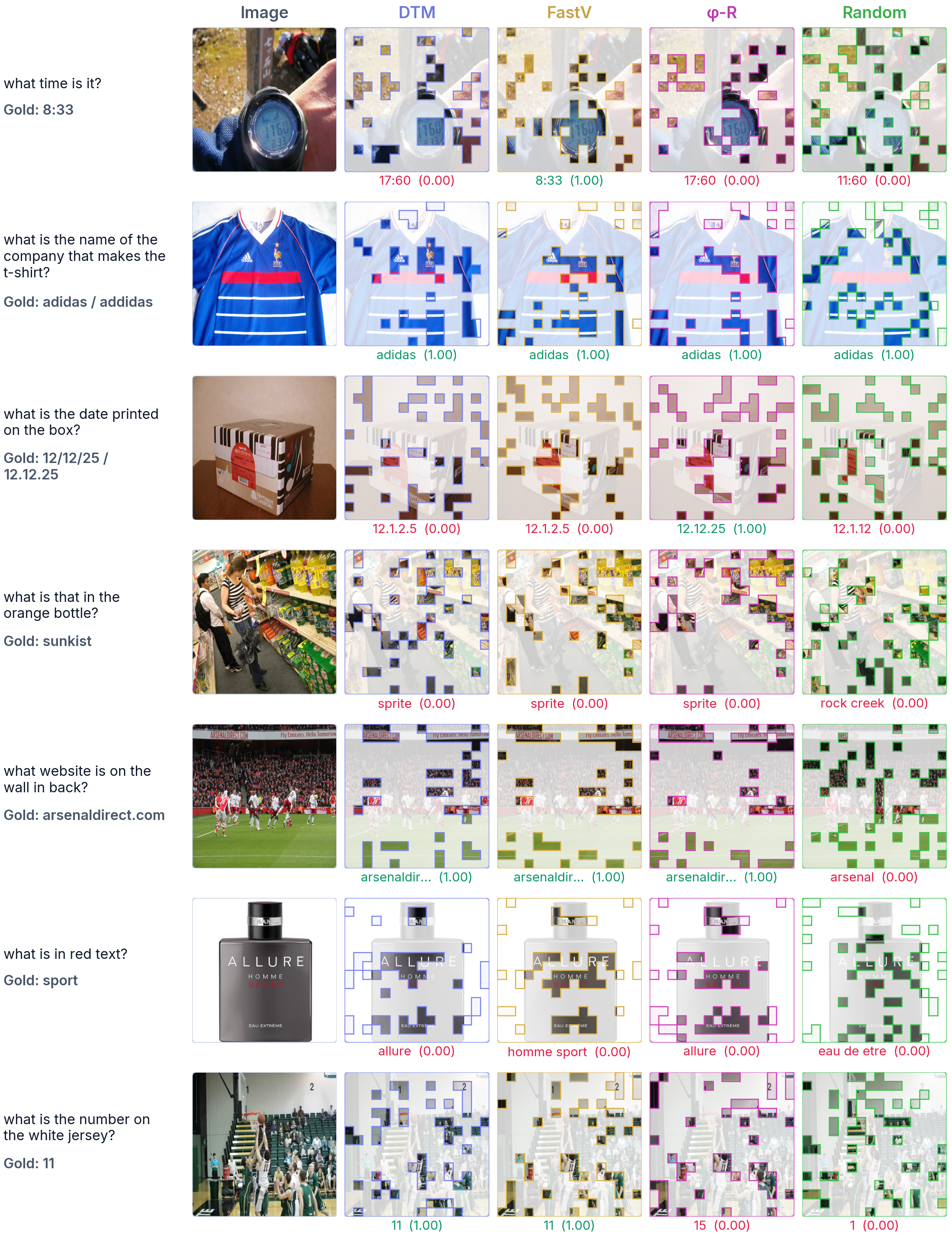}
\caption{\textbf{Randomly sampled TextVQA questions at matched 80\% deletion on T5Gemma 2 1B--1B (2 of 2).} Presented as in Figure~\ref{fig:qual-textvqa-1}; blue \dtm, gold \fastv, magenta \phir, green \Random.}
\label{fig:qual-textvqa-2}
\end{figure}

Figures~\ref{fig:qual-chartqa-1}--\ref{fig:qual-textvqa-2} show 15 ChartQA and 15 TextVQA validation questions drawn uniformly at random, without regard to correctness, from all questions scored by the four T5Gemma 2 1B--1B checkpoints trained and evaluated at 80\% deletion; each method keeps 52 of 256 visual tokens. On these samples, \dtm answers 8 of 15 ChartQA and 9 of 15 TextVQA questions correctly, compared with 4 and 11 for \fastv, 6 and 8 for \phir, and 7 and 7 for \Random. On charts, \dtm tends to keep titles, axis and tick labels, and the ends of bars and lines. Table~\ref{tab:content} measures this on the full ChartQA validation set: \dtm keeps more text-label and bar/line patches than both attention-based selectors on all three backbones, and more numeric tick labels on Gemma 3 and T5Gemma 2 270M--270M, slightly fewer than \fastv on T5Gemma 2 1B--1B (83.2\% vs.\ 84.1\%). This is not a general preference for text: on DocVQA, the attention-based selectors cover more of the OCR text on every backbone, and on TextVQA \fastv does so on T5Gemma 2 1B--1B. \dtm's advantage there lies in the answer-bearing words (Figure~\ref{fig:evidence-retained}).

\begin{table}[htbp]
\centering
\caption{\textbf{What each method keeps on ChartQA and OCR-text benchmarks at matched 80\% deletion.} Recall is the percentage of patches in each category that a method keeps; uniform selection keeps 20.3\%. ChartQA categories come from the dataset's element annotations (text labels include titles, axis, and legend text; bar and line marks are plotted elements). ``Blank'' is the share of the kept budget spent on blank patches, which make up 47\% of ChartQA patches. OCR text covers every word detected in the image. On charts, \dtm keeps more text labels and marks than both attention-based selectors on every backbone; on documents, the attention-based selectors cover more text.}
\label{tab:content}
\footnotesize
\setlength{\tabcolsep}{2.5pt}
\resizebox{\linewidth}{!}{%
\begin{tabular}{@{}ll*{4}{c}@{\hspace{10pt}}*{2}{c}@{}}
\toprule
 & & \multicolumn{4}{c}{ChartQA} & \multicolumn{2}{c}{OCR-text recall} \\
\cmidrule(lr){3-6}\cmidrule(l){7-8}
Model & Method & Text labels & Numeric ticks & Bars/lines & Blank (\% of kept) & TextVQA & DocVQA \\
\midrule
T5Gemma 2 1B--1B & \dtm & \textbf{66.8} & 83.2 & \textbf{24.3} & \textbf{20.2} & 58.0 & 31.4 \\
 & \fastv & 60.7 & \textbf{84.1} & 17.4 & 26.9 & \textbf{62.5} & \textbf{35.8} \\
 & \phir & 51.2 & 66.7 & 21.0 & 27.3 & 58.4 & 34.7 \\
\midrule
Gemma 3 4B & \dtm & \textbf{66.7} & \textbf{80.3} & \textbf{24.5} & 20.0 & \textbf{66.2} & 28.6 \\
 & \fastv & 49.9 & 49.9 & 20.6 & \textbf{17.7} & 54.5 & 34.0 \\
 & \phir & 47.1 & 37.3 & 23.4 & \textbf{17.7} & 56.4 & \textbf{34.1} \\
\midrule
T5Gemma 2 270M--270M & \dtm & \textbf{57.1} & \textbf{70.7} & \textbf{24.7} & 26.6 & \textbf{61.2} & 30.4 \\
 & \fastv & 52.9 & 53.7 & 21.4 & \textbf{24.0} & 51.0 & \textbf{41.2} \\
 & \phir & 38.8 & 39.8 & 22.9 & 28.3 & 45.7 & 33.0 \\
\bottomrule
\end{tabular}%
}
\end{table}

\FloatBarrier
\section{FLOPs Metrics}

\subsection{FLOPs Calculations}
\label{app:flops-details}

Tables~\ref{tab:dataset-flops}, \ref{tab:gemma3-dataset-flops}, and~\ref{tab:270m-dataset-flops} list, for each benchmark, the input and output text lengths and the FLOPs saved at each visual deletion rate after layer 5. Savings are nearly constant across the nine benchmarks with short prompts and lower on RefCOCO+, whose longer text prompts and outputs are never deleted. At every rate, Gemma 3 4B saves slightly more than T5Gemma 2 1B--1B, and T5Gemma 2 270M--270M saves the least; at 80\% deletion, the means are 61.4\%, 58.4\%, and 53.1\%. Table~\ref{tab:docvqa-layers-flops} (Appendix~\ref{app:flops-by-layer}) shows how the savings shrink as the gate moves later, and Appendix~\ref{app:visual_flops} defines the complementary cost attributable to visual tokens alone.

\begin{table}[H]
\centering
\caption{\textbf{Per-dataset input and output lengths and FLOPs savings for T5Gemma 2 1B--1B at L5.} FLOPs are calculated consistent with the description in Appendix~\ref{app:flops}. Input text counts are unpadded, nonvisual encoder-input tokens, including prompt and special markers but excluding the generated answer. Output text counts include generated stop tokens and are pooled across the nine deletion rates. The additional selection FLOPs of \dtm, \fastv, and \phir are negligible (Appendices~\ref{app:vdtm_flops_appendix} and~\ref{app:fastv_and_phir_flops_appendix}). SD is the population standard deviation across evaluated examples. Each saving is computed from summed theoretical forward costs at the recorded input and generated-output lengths; the vision encoder and image projector are excluded. The final row weights tasks equally. Savings are nearly constant across benchmarks except RefCOCO+, whose longer prompts and outputs are never deleted.}
\label{tab:dataset-flops}
\footnotesize
\setlength{\tabcolsep}{2pt}
\begin{tabular}{lc@{\hspace{8pt}}c@{\hspace{8pt}}*{9}{r}}
\toprule
Dataset & \shortstack{Input text tokens\\mean $\pm$ SD}
& \shortstack{Output text tokens\\mean $\pm$ SD}
& \multicolumn{9}{c}{FLOPs saved (\%) at visual deletion rate} \\
\cmidrule(lr){4-12}
& & & 10\% & 20\% & 30\% & 40\% & 50\% & 60\% & 70\% & 80\% & 90\% \\
\midrule
DocVQA & $18.9 \pm 4.8$ & $5.9 \pm 3.5$ & 7.3 & 14.9 & 22.1 & 29.6 & 37.1 & 44.3 & 51.7 & 58.9 & 66.3 \\
ChartQA & $23.6 \pm 5.6$ & $4.4 \pm 1.9$ & 7.2 & 14.7 & 21.9 & 29.4 & 36.8 & 43.9 & 51.2 & 58.3 & 65.6 \\
InfoVQA & $22.0 \pm 5.3$ & $4.3 \pm 2.2$ & 7.3 & 14.8 & 22.1 & 29.5 & 37.0 & 44.2 & 51.6 & 58.7 & 66.0 \\
TextVQA & $16.3 \pm 2.5$ & $3.7 \pm 2.2$ & 7.4 & 15.2 & 22.6 & 30.2 & 37.8 & 45.2 & 52.7 & 60.0 & 67.5 \\
VQAv2 & $15.3 \pm 2.0$ & $2.3 \pm 0.7$ & 7.5 & 15.3 & 22.8 & 30.5 & 38.3 & 45.6 & 53.3 & 60.6 & 68.2 \\
VizWiz & $16.9 \pm 7.0$ & $3.6 \pm 1.3$ & 7.4 & 15.1 & 22.5 & 30.2 & 37.8 & 45.1 & 52.7 & 59.9 & 67.4 \\
GQA & $17.9 \pm 3.7$ & $2.1 \pm 0.3$ & 7.5 & 15.2 & 22.6 & 30.3 & 37.9 & 45.3 & 52.9 & 60.1 & 67.7 \\
TallyQA & $15.8 \pm 2.1$ & $2.0 \pm 0.1$ & 7.5 & 15.3 & 22.8 & 30.5 & 38.2 & 45.6 & 53.3 & 60.6 & 68.2 \\
POPE & $16.2 \pm 0.4$ & $2.0 \pm 0.0$ & 7.5 & 15.3 & 22.8 & 30.5 & 38.2 & 45.6 & 53.2 & 60.5 & 68.1 \\
RefCOCO+ & $62.9 \pm 2.4$ & $29.0 \pm 0.8$ & 5.8 & 11.8 & 17.5 & 23.5 & 29.4 & 35.1 & 41.0 & 46.6 & 52.4 \\
\midrule
Mean across tasks & --- & --- & 7.3 & 14.8 & 22.0 & 29.4 & 36.9 & 44.0 & 51.4 & 58.4 & 65.8 \\
\bottomrule
\end{tabular}

\end{table}

\begin{table}[H]
\centering
\caption{\textbf{Per-dataset input and output lengths and FLOPs savings for Gemma 3 4B at L5.} FLOPs savings are presented in the same way as Table~\ref{tab:dataset-flops}. Gemma 3 saves slightly more than T5Gemma 2 1B--1B at every deletion rate.}
\label{tab:gemma3-dataset-flops}
\footnotesize
\setlength{\tabcolsep}{2pt}
\begin{tabular}{lc@{\hspace{8pt}}c@{\hspace{8pt}}*{9}{r}}
\toprule
Dataset & \shortstack{Input text tokens\\mean $\pm$ SD}
& \shortstack{Output text tokens\\mean $\pm$ SD}
& \multicolumn{9}{c}{FLOPs saved (\%) at visual deletion rate} \\
\cmidrule(lr){4-12}
& & & 10\% & 20\% & 30\% & 40\% & 50\% & 60\% & 70\% & 80\% & 90\% \\
\midrule
DocVQA   & $18.9 \pm 4.8$ & $5.8 \pm 3.5$ & 7.6 & 15.6 & 23.2 & 31.1 & 39.0 & 46.5 & 54.4 & 62.0 & 69.9 \\
ChartQA  & $23.6 \pm 5.6$ & $4.4 \pm 2.1$ & 7.6 & 15.4 & 23.0 & 30.8 & 38.6 & 46.1 & 53.8 & 61.3 & 69.1 \\
InfoVQA  & $22.0 \pm 5.3$ & $4.2 \pm 2.2$ & 7.6 & 15.5 & 23.1 & 31.0 & 38.8 & 46.4 & 54.2 & 61.7 & 69.5 \\
TextVQA  & $16.3 \pm 2.5$ & $3.7 \pm 2.5$ & 7.8 & 15.9 & 23.6 & 31.7 & 39.7 & 47.4 & 55.4 & 63.0 & 71.0 \\
VQAv2    & $15.3 \pm 2.0$ & $2.3 \pm 0.7$ & 7.9 & 16.0 & 23.9 & 32.0 & 40.1 & 47.9 & 55.9 & 63.7 & 71.7 \\
VizWiz   & $16.9 \pm 7.0$ & $3.5 \pm 1.3$ & 7.8 & 15.9 & 23.6 & 31.6 & 39.6 & 47.3 & 55.3 & 63.0 & 70.9 \\
GQA      & $17.9 \pm 3.7$ & $2.1 \pm 0.3$ & 7.8 & 15.9 & 23.7 & 31.7 & 39.8 & 47.5 & 55.5 & 63.1 & 71.1 \\
TallyQA  & $15.8 \pm 2.1$ & $2.0 \pm 0.1$ & 7.9 & 16.0 & 23.8 & 32.0 & 40.1 & 47.8 & 55.9 & 63.6 & 71.7 \\
POPE     & $16.2 \pm 0.4$ & $2.0 \pm 0.0$ & 7.8 & 16.0 & 23.8 & 31.9 & 40.0 & 47.8 & 55.8 & 63.6 & 71.6 \\
RefCOCO+ & $62.9 \pm 2.4$ & $29.0 \pm 0.6$ & 6.1 & 12.4 & 18.5 & 24.8 & 31.1 & 37.2 & 43.5 & 49.5 & 55.7 \\
\midrule
Mean across tasks & --- & ---
& 7.6 & 15.5 & 23.0 & 30.9 & 38.7 & 46.2 & 54.0 & 61.4 & 69.2 \\
\bottomrule
\end{tabular}

\end{table}

\begin{table}[H]
\centering
\caption{\textbf{Per-dataset input and output lengths and FLOPs savings for T5Gemma 2 270M--270M at L5.} FLOPs savings are presented in the same way as Table~\ref{tab:dataset-flops}. T5Gemma 2 270M--270M saves the least of the three backbones at every rate.}
\label{tab:270m-dataset-flops}
\footnotesize
\setlength{\tabcolsep}{2pt}
\begin{tabular}{lc@{\hspace{8pt}}c@{\hspace{8pt}}*{9}{r}}
\toprule
Dataset & \shortstack{Input text tokens\\mean $\pm$ SD}
& \shortstack{Output text tokens\\mean $\pm$ SD}
& \multicolumn{9}{c}{FLOPs saved (\%) at visual deletion rate} \\
\cmidrule(lr){4-12}
& & & 10\% & 20\% & 30\% & 40\% & 50\% & 60\% & 70\% & 80\% & 90\% \\
\midrule
DocVQA   & $18.9 \pm 4.8$ & $5.8 \pm 3.5$ & 6.9 & 13.9 & 20.6 & 27.4 & 34.2 & 40.6 & 47.2 & 53.4 & 59.9 \\
ChartQA  & $23.6 \pm 5.6$ & $4.4 \pm 2.0$ & 6.9 & 13.9 & 20.5 & 27.4 & 34.1 & 40.4 & 46.9 & 53.1 & 59.4 \\
InfoVQA  & $22.0 \pm 5.3$ & $4.2 \pm 2.2$ & 6.9 & 14.0 & 20.7 & 27.5 & 34.3 & 40.7 & 47.2 & 53.5 & 59.9 \\
TextVQA  & $16.3 \pm 2.5$ & $3.7 \pm 2.1$ & 7.1 & 14.3 & 21.1 & 28.2 & 35.1 & 41.6 & 48.3 & 54.7 & 61.2 \\
VQAv2    & $15.3 \pm 2.0$ & $2.3 \pm 0.7$ & 7.2 & 14.5 & 21.5 & 28.6 & 35.6 & 42.2 & 49.0 & 55.5 & 62.0 \\
VizWiz   & $16.9 \pm 7.0$ & $3.5 \pm 1.3$ & 7.1 & 14.3 & 21.1 & 28.1 & 35.1 & 41.6 & 48.3 & 54.6 & 61.1 \\
GQA      & $17.9 \pm 3.7$ & $2.1 \pm 0.3$ & 7.1 & 14.4 & 21.3 & 28.4 & 35.4 & 42.0 & 48.7 & 55.1 & 61.6 \\
TallyQA  & $15.8 \pm 2.1$ & $2.0 \pm 0.1$ & 7.2 & 14.5 & 21.5 & 28.6 & 35.6 & 42.3 & 49.1 & 55.5 & 62.1 \\
POPE     & $16.2 \pm 0.4$ & $2.0 \pm 0.0$ & 7.2 & 14.5 & 21.5 & 28.6 & 35.6 & 42.2 & 49.0 & 55.4 & 62.0 \\
RefCOCO+ & $62.9 \pm 2.4$ & $29.0 \pm 0.0$ & 5.2 & 10.6 & 15.7 & 20.9 & 26.0 & 30.9 & 35.8 & 40.5 & 45.4 \\
\midrule
Mean across tasks & --- & ---
& 6.9 & 13.9 & 20.5 & 27.4 & 34.1 & 40.4 & 47.0 & 53.1 & 59.5 \\
\bottomrule
\end{tabular}

\end{table}

\subsection{FLOPs Savings Across Deletion Layers}
\label{app:flops-by-layer}

\begin{table}[H]
\centering
\caption{\textbf{Per-layer FLOPs savings on DocVQA.} Each row shows the FLOPs saved (\%) when the gate follows the indicated layer, evaluated across visual deletion rates. FLOPs are calculated in the same way as Table~\ref{tab:dataset-flops}.}
\label{tab:docvqa-layers-flops}
\label{tab:docvqa-t5-layers}
\label{tab:docvqa-gemma3-layers}
\label{tab:docvqa-270m-layers}
\footnotesize
\setlength{\tabcolsep}{4pt}
\begin{tabular}{@{}l*{9}{r}@{}}
\toprule
Gate layer & \multicolumn{9}{c}{FLOPs saved (\%) at visual deletion rate} \\
\cmidrule(l){2-10}
& 10\% & 20\% & 30\% & 40\% & 50\% & 60\% & 70\% & 80\% & 90\% \\
\midrule
\multicolumn{10}{@{}l}{\textit{T5Gemma 2 1B--1B}} \\
3 & 8.0 & 16.2 & 24.1 & 32.3 & 40.5 & 48.4 & 56.5 & 64.3 & 72.4 \\
4 & 7.6 & 15.6 & 23.1 & 31.0 & 38.8 & 46.3 & 54.1 & 61.6 & 69.4 \\
5 & 7.3 & 14.9 & 22.1 & 29.6 & 37.1 & 44.3 & 51.7 & 58.9 & 66.3 \\
6 & 7.0 & 14.2 & 21.1 & 28.2 & 35.4 & 42.2 & 49.3 & 56.1 & 63.2 \\
7 & 6.6 & 13.5 & 20.1 & 26.9 & 33.7 & 40.2 & 46.9 & 53.4 & 60.1 \\
\midrule
\multicolumn{10}{@{}l}{\textit{Gemma 3 4B}} \\
3 & 8.2 & 16.7 & 24.8 & 33.2 & 41.7 & 49.7 & 58.1 & 66.2 & 74.6 \\
4 & 7.9 & 16.1 & 24.0 & 32.2 & 40.3 & 48.1 & 56.3 & 64.1 & 72.2 \\
5 & 7.6 & 15.6 & 23.2 & 31.1 & 39.0 & 46.5 & 54.4 & 62.0 & 69.9 \\
6 & 7.4 & 15.0 & 22.4 & 30.0 & 37.6 & 44.9 & 52.5 & 59.9 & 67.5 \\
7 & 7.1 & 14.5 & 21.6 & 28.9 & 36.3 & 43.3 & 50.7 & 57.7 & 65.0 \\
\midrule
\multicolumn{10}{@{}l}{\textit{T5Gemma 2 270M--270M}} \\
3 & 7.9 & 15.9 & 23.5 & 31.4 & 39.1 & 46.4 & 53.9 & 61.1 & 68.4 \\
4 & 7.4 & 14.9 & 22.1 & 29.4 & 36.6 & 43.5 & 50.6 & 57.3 & 64.1 \\
5 & 6.9 & 13.9 & 20.6 & 27.4 & 34.2 & 40.6 & 47.2 & 53.4 & 59.9 \\
6 & 6.4 & 12.9 & 19.1 & 25.5 & 31.7 & 37.7 & 43.8 & 49.6 & 55.6 \\
7 & 5.9 & 11.9 & 17.6 & 23.5 & 29.3 & 34.8 & 40.4 & 45.8 & 51.3 \\
\bottomrule
\end{tabular}
\end{table}

\FloatBarrier

\section{FLOPs Accounting}
\label{app:flops}

\subsection{Notation and counting convention}
\label{app:flops_overview}

We count two FLOPs per multiply--add. Counts include dominant
matrix multiplications in the language backbone and vocabulary projection,
but exclude SigLIP and the multimodal projector. Normalization, RoPE,
softmax, pointwise activations, residual additions, biases, averaging,
ranking, and token gathering are omitted. These are aggregate forward-FLOP
estimates, not wall-clock latency or detailed cached-decoding costs.

For T5Gemma~2, $N$ and $M$ are encoder and decoder lengths; $L_e,L_d$ are
their depths and $W_e,W_d$ their local-attention windows.
The hidden width, head width, GeGLU intermediate width, query-head count,
key/value-head count, and vocabulary size are denoted by
$d_{\mathrm{model}}$, $d_{\mathrm{head}}$, $d_{\mathrm{ff}}$,
$n_{\mathrm{heads}}$, $n_{\mathrm{kv\text{-}heads}}$, and
$n_{\mathrm{vocab}}$, respectively. Dimensions and layer types are read
from each stack's configuration; we do not assume
$d_{\mathrm{model}}=n_{\mathrm{heads}}d_{\mathrm{head}}$.
The main text's $n,m,d,d_h$ correspond to $N,M,d_{\mathrm{model}},
d_{\mathrm{head}}$ here.

\subsection{T5Gemma 2}
\label{app:flops_backbones}

An attention pair consists of one query position and one key position;
their projected vectors determine one attention score. The functions below
count these pairs \emph{per query head}, under the stated approximations,
for encoder and merged-attention decoder layer $i$, respectively:
\begin{align}
\mathcal P_{\mathrm{enc}}(N,i)
&=
\begin{cases}
N^2,&\text{global},\\
N\min(N,W_e),&\text{local},
\end{cases}
\label{eq:flops_pairs}\\
\mathcal P_{\mathrm{dec}}(N,M,i)
&=
\begin{cases}
MN+M^2,&\text{global},\\
MN+M\min(M,W_d),&\text{local}.
\end{cases}
\label{eq:flops_decoder_pairs}
\end{align}
Every decoder query attends to all encoder keys, even in local decoder
layers. The $M^2$ term deliberately uses a full-square approximation
rather than exact causal pairs. Local counts assume only window-permitted
pairs are computed, not a dense matrix followed by masking, and ignore
window-boundary effects.

Grouped-query attention shares key/value heads across query heads.
The query and output projections therefore scale with $n_{\mathrm{heads}}$,
whereas the key and value projections scale with
$n_{\mathrm{kv\text{-}heads}}\leq n_{\mathrm{heads}}$.
Each query head still computes its own attention scores and weighted
value sums, so both attention products scale with $n_{\mathrm{heads}}$,
not $n_{\mathrm{kv\text{-}heads}}$.

For each query head, $QK^\top$ computes scores from projected queries and
keys, and $AV$ aggregates projected value vectors using the normalized
attention weights $A$. Each product costs $2d_{\mathrm{head}}$ per
query--key pair. The GeGLU MLP has gate, up, and down projections.
The encoder-layer cost is therefore
\begin{equation}
\begin{aligned}
F_{\mathrm{enc\text{-}layer}}(N,i)
={}&\underbrace{4N d_{\mathrm{model}}d_{\mathrm{head}}
(n_{\mathrm{heads}}+n_{\mathrm{kv\text{-}heads}})}
_{Q,K,V\text{ and output projections}}\\
&+\underbrace{4\mathcal P_{\mathrm{enc}}(N,i)n_{\mathrm{heads}}d_{\mathrm{head}}}
_{QK^\top\text{ scores and }AV\text{ value aggregation}}
+\underbrace{6N d_{\mathrm{model}}d_{\mathrm{ff}}}
_{\text{GeGLU MLP (gate, up, down)}}.
\end{aligned}
\label{eq:flops_encoder}
\end{equation}
The decoder uses the same components, plus $K/V$ projections of the
encoder states within its merged attention:
\begin{equation}
\begin{aligned}
F_{\mathrm{dec\text{-}layer}}(N,M,i)
={}&\underbrace{4M d_{\mathrm{model}}d_{\mathrm{head}}
(n_{\mathrm{heads}}+n_{\mathrm{kv\text{-}heads}})}
_{\text{decoder-state }Q,K,V\text{ and output projections}}\\
&+\underbrace{4N d_{\mathrm{model}}d_{\mathrm{head}}n_{\mathrm{kv\text{-}heads}}}
_{\text{encoder-state }K,V\text{ projections}}\\
&+\underbrace{4\mathcal P_{\mathrm{dec}}(N,M,i)n_{\mathrm{heads}}d_{\mathrm{head}}}
_{QK^\top\text{ scores and }AV\text{ value aggregation}}
+\underbrace{6M d_{\mathrm{model}}d_{\mathrm{ff}}}
_{\text{GeGLU MLP (gate, up, down)}}.
\end{aligned}
\label{eq:flops_decoder}
\end{equation}
There is no separate cross-attention module. Summing layers gives
\begin{equation}
\begin{aligned}
F_{\mathrm{encoder}}(N)&=\sum_{i=1}^{L_e}F_{\mathrm{enc\text{-}layer}}(N,i),\\
F_{\mathrm{decoder}}(N,M)&=\sum_{i=1}^{L_d}F_{\mathrm{dec\text{-}layer}}(N,M,i).
\end{aligned}
\label{eq:flops_stacks}
\end{equation}
Adding the projection to vocabulary logits at the $M$ output positions,
\begin{equation}
\begin{aligned}
F_{\mathrm{T5Gemma}}(N,M)
={}&F_{\mathrm{encoder}}(N)+F_{\mathrm{decoder}}(N,M)\\
&+2M d_{\mathrm{model}}n_{\mathrm{vocab}}.
\end{aligned}
\label{eq:flops_t5_baseline}
\end{equation}

\subsection{Gemma 3}
\label{app:flops_gemma}

Gemma~3 has a single decoder-only stack. Let
$N_{\mathrm{prompt}}$ be prompt length, $M_{\mathrm{gen}}$ generated length,
$T=N_{\mathrm{prompt}}+M_{\mathrm{gen}}$, and $L_g,W_g$ its depth and window.
Its layer cost has the same form as the T5Gemma encoder, with Gemma's
own dimensions:
\begin{align}
\mathcal P_g(T,i)
&=
\begin{cases}
T^2,&\text{global},\\
T\min(T,W_g),&\text{local},
\end{cases}
\label{eq:flops_gemma_pairs}\\
F_{\mathrm{Gemma\text{-}layer}}(T,i)
&=\underbrace{4T d_{\mathrm{model}}d_{\mathrm{head}}
(n_{\mathrm{heads}}+n_{\mathrm{kv\text{-}heads}})}
_{Q,K,V\text{ and output projections}}
\nonumber\\
&\quad+\underbrace{4\mathcal P_g(T,i)n_{\mathrm{heads}}d_{\mathrm{head}}}
_{QK^\top\text{ scores and }AV\text{ value aggregation}}
+\underbrace{6T d_{\mathrm{model}}d_{\mathrm{ff}}}
_{\text{GeGLU MLP (gate, up, down)}},
\label{eq:flops_gemma_layer}\\
F_{\mathrm{Gemma3}}(N_{\mathrm{prompt}},M_{\mathrm{gen}})
&=\sum_{i=1}^{L_g}F_{\mathrm{Gemma\text{-}layer}}(T,i)
+2M_{\mathrm{gen}}d_{\mathrm{model}}n_{\mathrm{vocab}}.
\label{eq:flops_gemma_baseline}
\end{align}
There is no encoder or merged-attention term. Only output logits are
counted. The full-square $T^2$ convention is an aggregate proxy, not
permission for prompt tokens to attend to future generated tokens.
It uses $N_{\mathrm{prompt}}+M_{\mathrm{gen}}$ positions rather than the
$N_{\mathrm{prompt}}+M_{\mathrm{gen}}-1$ inputs of exact cached generation.
Local counts use compacted lengths, without original-position gaps or
image-specific attention masks.

\subsection{Token pruning methods}
\label{app:flops_pruning}

\subsubsection{General pruning backbone}

\dtm, \fastv, and \phir share the following backbone computation for fixed
original lengths, retained length, and pruning layer. Each method then
adds its own selection cost.

Let $k$ be the one-indexed layer \emph{after which} pruning
occurs: $1\leq k\leq L_e$ for T5Gemma or $1\leq k\leq L_g$ for Gemma.
Thus pruning \emph{after layer $k$} here places the gate \emph{before layer $k+1$} in the
main-text convention (Figure~\ref{fig:vmrt5gemma2}); our L5 setting is $k=5$ here. Write $\widetilde N$ for the retained encoder length.
For total-input deletion rate $\delta$, $\widetilde N=(1-\delta)N$,
where $0\leq\delta<1$.
The shared T5Gemma computation, before adding a selection criterion, is
\begin{equation}
\begin{aligned}
F_{\mathrm{pruned\text{-}backbone}}(N,\widetilde N,M,k)
={}&\sum_{i=1}^{k}F_{\mathrm{enc\text{-}layer}}(N,i)
+\sum_{i=k+1}^{L_e}F_{\mathrm{enc\text{-}layer}}(\widetilde N,i)\\
&+F_{\mathrm{decoder}}(\widetilde N,M)
+2M d_{\mathrm{model}}n_{\mathrm{vocab}}.
\end{aligned}
\label{eq:flops_t5_pruned}
\end{equation}
Gemma instead shortens the prompt during prefill from $N_\mathrm{prompt}$  to a retained length 
$\widetilde N_{\mathrm{prompt}}=(1-\delta)N_{\mathrm{prompt}}$;
generated tokens are never removed. Define $\widetilde{T}=\widetilde{N}_{\mathrm{prompt}}+M_{\mathrm{gen}}$, analogously to $T=N_{\mathrm{prompt}}+M_{\mathrm{gen}}$ for the unpruned sequence. With
\begin{equation}
\begin{aligned}
F_{\mathrm{Gemma\text{-}backbone}}
\left(
N_{\mathrm{prompt}},
\widetilde N_{\mathrm{prompt}},
M_{\mathrm{gen}},
k
\right)
={}&
\sum_{i=1}^{k}
F_{\mathrm{Gemma\text{-}layer}}(T,i)
\\
&+
\sum_{i=k+1}^{L_g}
F_{\mathrm{Gemma\text{-}layer}}(\widetilde T,i)
\\
&+
2M_{\mathrm{gen}}d_{\mathrm{model}}n_{\mathrm{vocab}}.
\end{aligned}
\label{eq:flops_gemma_pruned}
\end{equation}

Empty sums are zero: pruning after the final encoder layer can still
save T5Gemma decoder work, whereas pruning after Gemma's final layer saves
no backbone computation. Fractional retained lengths are allowed for
expected FLOPs.

In our visual-pruning experiments, only visual states are removed.
The retained encoder or prompt length is therefore the unchanged text-token
count plus the retained visual-token count; output lengths are unchanged.

\subsubsection{\texorpdfstring{\dtm}{DTM}}
\label{app:vdtm_flops_appendix}

\dtm adds one scalar projection per gated input state, not another
attention or feed-forward block. When gating all input tokens,
\begin{align}
F_{\mathrm{DTM}}
&=F_{\mathrm{pruned\text{-}backbone}}+2N d_{\mathrm{model}},
\label{eq:flops_dtm}\\
F_{\mathrm{Gemma3\text{-}DTM}}
&=F_{\mathrm{Gemma\text{-}backbone}}
+2N_{\mathrm{prompt}}d_{\mathrm{model}}.
\label{eq:flops_gemma_dtm}
\end{align}
The gate runs once, including at zero deletion. Its small cost is retained,
not rounded away. For visual-only DTM, the gate terms instead become
$2N_vd_{\mathrm{model}}$ for $N_v$ visual encoder tokens or
$2N_{\mathrm{prompt},v}d_{\mathrm{model}}$ for
$N_{\mathrm{prompt},v}$ visual prompt tokens. Text tokens are not gated;
no gate runs when there are no visual tokens.

\subsubsection{\texorpdfstring{\fastv and \phir}{FastV and Phi-R}}
\label{app:fastv_and_phir_flops_appendix}

Our \fastv and \phir baselines use \emph{all-mean} importance: attention
received by each eligible key is averaged over all query heads and all
available input queries in layer $k$'s attention pattern. Gemma uses
prompt queries only, never future generated queries.
We omit separate equations for other attention-aggregation variants, as
we find their differences in total FLOPs negligible in the evaluated settings.

\fastv reuses normal post-RoPE attention values and adds no dominant matrix
multiplication. \phir ($\phi_{-R}$) reuses projected pre-RoPE $Q/K$
(including Gemma's Q/K normalization), but adds one RoPE-free score
contraction. It adds neither new projections nor a second value-aggregation
product $AV$:
\begin{align}
F_{\mathrm{FastV}}&=F_{\mathrm{pruned\text{-}backbone}},\\
F_{\phi_{-R}}&=F_{\mathrm{pruned\text{-}backbone}}
+2\mathcal P_{\mathrm{enc}}(N,k)n_{\mathrm{heads}}d_{\mathrm{head}},\\
F_{\mathrm{Gemma3\text{-}FastV}}&=F_{\mathrm{Gemma\text{-}backbone}},\\
F_{\mathrm{Gemma3},\phi_{-R}}&=F_{\mathrm{Gemma\text{-}backbone}}
+2\mathcal P_g(N_{\mathrm{prompt}},k)n_{\mathrm{heads}}d_{\mathrm{head}}.
\label{eq:flops_selectors}
\end{align}
Thus \phir uses the original input's global or local pair count, not
the retained length or Gemma's combined length $T$. \fastv's zero
\emph{additional matmul} cost assumes reusable attention values; averaging
and selection are omitted, not literally free.
For visual-only pruning, only visual keys are eligible for retention, but
the all-mean criterion still uses all available input queries. \phir retains
the full-input score contraction; a visual-key-only optimization is not
assumed. Neither method runs a pruning criterion when no visual tokens exist.

\subsection{Marginal visual-token FLOPs}
\label{app:visual_flops}

Our gates target visual tokens, so we seek to quantify both the backbone
cost attributable to processing those tokens and how much of that cost the
gates remove. We therefore define \emph{marginal visual-token FLOPs} as the
difference between a method's cost with visual input and its cost on an
otherwise identical text-only input. Comparing this marginal cost with
the unpruned baseline then measures how much visual processing work the
gate saves, rather than expressing savings only relative to the entire model.

For either model family, let $N_t,N_v$ denote the numbers of text and
visual input tokens, and $M$ the output text length. For T5Gemma~2, the
input is the encoder sequence and $M$ is the decoder length. For Gemma~3,
the input is the prompt and $M=M_{\mathrm{gen}}$. For any method
$\mathcal M$ applied to either backbone, define
\begin{equation}
F_{\mathrm{visual}}^{\mathcal M}(N_t,N_v,M)
=\underbrace{F_{\mathcal M}(N_t,N_v,M)}_{\text{with visual tokens}}
-\underbrace{F_{\mathcal M}(N_t,0,M)}_{\text{text-only reference}}.
\label{eq:flops_visual}
\end{equation}
Here $F_{\mathcal M}$ is the whole-model cost for the chosen model family
and method, using the visual-only pruning rules above. Both evaluations
keep the model, text and output
lengths, pruning layer, and requested visual-retention setting fixed.
The text-only reference contains no visual tokens at any layer and skips
visual pruning and its criterion, so it equals the text-only baseline.

This is a \emph{marginal attribution}, not a separation into independent
text and visual operations. For example, the global encoder pair increment
is $(N_t+N_v)^2-N_t^2=N_v^2+2N_tN_v$: counting only visual--visual
pairs misses interactions with text.
T5Gemma attribution also includes the decoder's attention to visual states
and their $K/V$ projections. For Gemma, it includes interactions between
visual prompt states and text/generated states under the full-square proxy.
Vocabulary FLOPs cancel; SigLIP and the multimodal projector remain excluded.
The attribution includes visual-pruning overhead and still depends on
text/output length; it is not the full cost of image processing.

\end{document}